%% file: main.tex
\documentclass{article}

\PassOptionsToPackage{numbers,compress,sort&compress}{natbib}
\usepackage[eandd,final]{neurips_2026}

\usepackage[utf8]{inputenc}
\usepackage[T1]{fontenc}
\usepackage{hyperref}
\usepackage{etoc}
\usepackage{url}
\usepackage{booktabs}
\usepackage{amsfonts}
\usepackage{amsmath}
\usepackage{nicefrac}
\usepackage{microtype}
\usepackage{xcolor}
\usepackage{graphicx}
\usepackage{multirow}
\usepackage{enumitem}
\usepackage{colortbl}
\usepackage{pifont}
\newcommand{\cmark}{\ding{51}}
\newcommand{\xmark}{\ding{55}}
\usepackage{subcaption}

\newcommand{\listqa}{\textsc{ListQA}}
\newcommand{\simplelistqa}{\textsc{SimpleListQA}}
\newcommand{\complexlistqa}{\textsc{ComplexListQA}} 

\newcommand{\judgef}{LLM-Judge-F1}
\newcommand{\rougef}{ROUGE-F1}
\newcommand{\formatmetric}{Format-Correctness}
\newcommand{\orderedf}{LLM-Judge-F1-Ordered}
\newcommand{\relaxedrecall}{Relaxed-Recall}
\newcommand{\relaxedprecision}{Relaxed-Precision} 

\title{ListQA: A Benchmark for Evaluating List-Formatted Factual Knowledge Retrieval in Large Language Models}

\author{%
  Mayank Kulkarni$^{*}$\\
  Amazon AGI\\
  \And
  Rakesh Vaideeswaran$^{*}$\\
  Amazon AGI\\
}

\begin{document}

\maketitle
{\renewcommand{\thefootnote}{\ensuremath{*}}\footnotetext{Equal contribution.}}
\setcounter{footnote}{0}
\etocdepthtag.toc{mainpaper}
\input{sections/abstract}
\input{sections/intro_iteration}

\input{sections/dataset}
\input{sections/evaluation}

\input{sections/experiments}

\input{sections/conclusion}

\begin{ack}
We thank all our colleagues and mentors at Amazon AGI for their support and internal feedback in the building of this benchmark including Yang Liu, Jianhua Lu, Shankar Ananthakrishnan, Arindam Mandal, Kang Min Yoo, He Xie and Anna Rumshisky. 

\end{ack}

\bibliographystyle{unsrtnat}
\bibliography{references}

\appendix
\input{sections/appendix}

\newpage
\input{checklist}

\end{document}

%% file: sections/abstract.tex
\begin{abstract}
Many real-world questions demand not a single fact but an organized collection of them, yet existing factual knowledge benchmarks almost exclusively target single-answer retrieval. We introduce \listqa{}, a benchmark of 9,045 human-curated, cross-validated questions (audited error rate of 1.7\%) that require LLMs to recall multiple facts and compose them into structured lists of 3--10 elements. Questions range from flat lists (\textit{``Which NCAA teams went undefeated between 2000 and 2024?''}) to hierarchical lists with sub-attributes (\textit{``\ldots and what was their record and result?''}), spanning eight categories across 60+ countries. An evergreen design (explicit temporal constraints or historically immutable facts) keeps ground truth valid without periodic updates. For evaluation, we frame element matching as a linear sum assignment problem: optimal bipartite matching paired with LLM-based semantic grading jointly captures factual accuracy, hallucination, and format compliance. We benchmark 34 models (8 families, 42 configurations across standard and thinking-enabled inference) and find: the best standard-inference model scores below 37\% \judgef{}, multi-faceted questions are harder across the board, and providing the source Wikipedia page lifts scores dramatically, confirming that models can organize facts but struggle to retrieve them from parameters alone. Extended analyses cover base vs.\ instruction-tuned checkpoints and token efficiency. We release the dataset and evaluation code \footnote{Dataset and evaluation code can be found at \url{https://listqa.github.io}}.
\end{abstract}

%% file: sections/intro_iteration.tex
\section{Introduction}
\label{sec:introduction}

Large language models encode substantial factual knowledge in their parameters~\citep{petroni2019language, roberts2020knowledge}. When we evaluate this knowledge, we almost always ask for a single fact: \textit{Who directed Oppenheimer? What is the capital of Bhutan? When was penicillin discovered?} Yet many real information needs have not one correct answer but an entire set: a traveler asks which direct routes an airline launched since 2022, a researcher asks which countries ratified a treaty before a given year, a clinician asks which drugs were approved for a condition in a specific window along with their mechanisms of action. Answering such questions requires a model to recall multiple discrete entities that meet specified criteria, verify each against temporal and categorical constraints, and in the more demanding cases, enrich each with structured sub-attributes. In short, the user needs the \emph{complete} set of qualifying facts, without fabricated additions, organized with the requested attributes. Moreover, structure matters: when an answer contains multiple independent facts, organizing them as a list, which is the most natural format for enumeration in language, turns evaluation from a monolithic judgment into a per-element one, enabling users and automated metrics alike to audit each item independently, detect individual hallucinations, and act on specific entries without parsing an undifferentiated paragraph. These properties are not merely convenient but functionally necessary in domains such as clinical decision-making and legal research, where each element may carry independent downstream consequences. 

\begin{table}[t]
\caption{Comparison of \listqa{} with representative benchmarks across the following dimensions: \textbf{ME}: multi-element retrieval. \textbf{SO}: structured output. \textbf{EG}: evergreen design. \textbf{K+IF}: knowledge + instruction following. Shading indicates partial coverage.}
\vspace{4pt}
\label{tab:diff_to_related_work}
\centering
\scriptsize
\setlength{\tabcolsep}{2.5pt}
\renewcommand{\arraystretch}{1.1}
\aboverulesep=1pt
\belowrulesep=1pt
\begin{minipage}[t]{0.48\textwidth}
\begin{tabular*}{\textwidth}{@{}l@{\extracolsep{\fill}}cccc@{}}
\toprule
\textbf{Benchmark} & \textbf{ME} & \textbf{SO} & \textbf{EG} & \textbf{K+IF} \\
\midrule
\multicolumn{5}{@{}l}{\textit{Single-answer factual QA}} \\
TriviaQA~\citep{joshi2017triviaqa} & \xmark & \xmark & \xmark & \xmark \\
Natural\ Questions~\citep{kwiatkowski2019natural} & \xmark & \xmark & \xmark & \xmark \\
PopQA~\citep{mallen2023trust} & \xmark & \xmark & \xmark & \xmark \\
SimpleQA~\citep{wei2024simpleqa} & \xmark & \xmark & \cmark & \xmark \\
FACT-Bench~\citep{yao2024factbench} & \xmark & \xmark & \xmark & \xmark \\
\midrule
\multicolumn{5}{@{}l}{\textit{Multi-answer factual QA}} \\
QAMPARI~\citep{amouyal2023qampari} & \cmark & \xmark & \xmark & \xmark \\
RoMQA~\citep{zhong2023romqa} & \cmark & \xmark & \xmark & \xmark \\
\midrule
\multicolumn{5}{@{}l}{\textit{Multi-hop \& reasoning QA}} \\
HotpotQA~\citep{yang2018hotpotqa} & \xmark & \xmark & \xmark & \xmark \\
MuSiQue~\citep{trivedi2022musique} & \xmark & \xmark & \xmark & \xmark \\
GPQA~\citep{rein2024gpqa} & \xmark & \xmark & \xmark & \xmark \\
\bottomrule
\end{tabular*}
\end{minipage}%
\hfill
\begin{minipage}[t]{0.48\textwidth}
\begin{tabular*}{\textwidth}{@{}l@{\extracolsep{\fill}}cccc@{}}
\toprule
\textbf{Benchmark} & \textbf{ME} & \textbf{SO} & \textbf{EG} & \textbf{K+IF} \\
\midrule
\multicolumn{5}{@{}l}{\textit{Structured \& table QA}} \\
WikiTableQuestions~\citep{pasupat2015compositional} & \cellcolor{gray!15}\cmark & \xmark & \xmark & \xmark \\
HybridQA~\citep{chen2020hybridqa} & \xmark & \xmark & \xmark & \xmark \\
\midrule
\multicolumn{5}{@{}l}{\textit{Temporal \& dynamic}} \\
FreshQA~\citep{vu2024freshllms} & \xmark & \xmark & \cellcolor{gray!15}\cmark & \xmark \\
RecencyQA~\citep{piryani2026recencyqa} & \xmark & \xmark & \cellcolor{gray!15}\cmark & \xmark \\
\midrule
\multicolumn{5}{@{}l}{\textit{Long-form factuality}} \\
FActScore~\citep{min2023factscore} & \cellcolor{gray!15}\cmark & \xmark & \xmark & \xmark \\
FACTS Grounding~\citep{deepmind2024facts} & \cellcolor{gray!15}\cmark & \xmark & \xmark & \cmark \\
\midrule
\multicolumn{5}{@{}l}{\textit{Knowledge calibration}} \\
SelfAware~\citep{yin2023selfaware} & \xmark & \xmark & \xmark & \xmark \\
\midrule
\cellcolor{blue!8} \listqa{} (ours) & \cmark & \cmark & \cmark & \cmark \\
\bottomrule
\end{tabular*}
\end{minipage}
\vspace{-10pt}
\end{table}

Most factual benchmarks~\citep{kwiatkowski2019natural, joshi2017triviaqa, rajpurkar2016squad, wei2024simpleqa, mallen2023trust, yao2024factbench} currently are not designed to measure this: they ask for a single answer, so they cannot assess how completely a model retrieves all qualifying entities, how precisely it avoids hallucinating extras, or whether it can structure the result coherently. Multi-answer factual QA datasets~\citep{amouyal2023qampari, zhong2023romqa} retrieve multiple entities per question, but evaluate only set overlap, not structured output or format compliance. Furthermore, much of the factual knowledge that underpins list-formatted answers resides not in running text but in structured sources such as tables and infoboxes, where answering a single question may require reasoning across multiple columns, applying filtering constraints, and aggregating rows, a retrieval mode that is fundamentally harder to perform parametrically than locating a fact stated explicitly in a narrative paragraph. Addressing this gap requires navigating several further challenges:

\textbf{The intersection of factual retrieval and structured output remains uncharted.} Significant progress has been made on each capability individually. Multi-hop benchmarks~\citep{yang2018hotpotqa, trivedi2022musique, rein2024gpqa} and fact reasoning tasks~\citep{jiang2024structfact, wang2024factlogicqa, agarwal2025mlrbench} have advanced factual reasoning evaluation, though they typically evaluate a single final answer per query; table QA datasets~\citep{pasupat2015compositional, chen2020hybridqa} have enabled rich evaluation of comprehension over provided tables. Instruction-following benchmarks~\citep{zhou2023ifeval, qin2024infobench, jiang2024followbench, zhao2025wildifeval} have improved the measurement of models’ ability to satisfy user-specified constraints and formats. Yet real information needs often require models to simultaneously provide accurate factual information while adhering to user-specified constraints, and their interaction is non-trivial: formatting constraints can degrade factual accuracy~\citep{murthy2024knowledge_if, deepmind2024facts}, and populating a multi-element structure may increase the tendency to hallucinate. This interplay has not been systematically measured by a dedicated benchmark.

\textbf{Temporal drift is an unsolved problem for multi-element benchmarks.} \citet{jiang2025benchage} show that 24--64\% of time-sensitive samples in popular factual datasets already carry outdated gold answers, silently inflating model scores. All existing mitigations, periodic refresh~\citep{vu2024freshllms, kasai2023realtime, white2024livebench}, drift detection~\citep{fierro2024mulan, piryani2026recencyqa, dhingra2022temporal, wenzel2024temporal, chen2025evolveqa, yang2020factduration}, inherently stable questions~\citep{wei2024simpleqa}, assume a single answer per sample. Lists break this assumption: when ground truth contains multiple elements, temporal validity must hold across \emph{every} entry simultaneously, and the probability of silent corruption grows with each element added. To the best of our knowledge, no existing benchmark guarantees this property, leaving multi-element factual evaluation vulnerable to the same decay that has already compromised single-answer benchmarks.

\textbf{Standard metrics fail on set-based, multi-element outputs.} Consider a model that correctly recalls ``NE Patriots'' when the reference says ``New England Patriots'', lists elements in a different order, and includes one hallucinated entry. Exact match scores this as entirely wrong; ROUGE~\citep{lin2004rouge} computes overlap across the entire response, unable to distinguish which elements are correct and which are fabricated. Evaluating such outputs requires solving two coupled problems: (1) aligning each predicted element to its reference counterpart, a combinatorial assignment problem when element order is irrelevant and the predicted and reference sets may differ in size and, (2) judging semantic equivalence at the element level, where lexical variation (e.g., abbreviations, reorderings of sub-attributes) is the norm. Existing LLM-as-judge~\citep{zheng2023judging, deepmind2024facts} and atomic-claim verification~\citep{min2023factscore} approaches provide scalable semantic grading but assume a single answer or unconstrained generation, offering no mechanism for element-to-element alignment. Without this alignment step, a judge that receives the full predicted list at once tends to accept it holistically when most elements appear plausible, rather than detecting individual hallucinations or omissions.

To address these gaps, we introduce \listqa{}, which, as Tab.~\ref{tab:diff_to_related_work} summarizes, is to our knowledge the first benchmark to simultaneously require their conjunction: multi-element retrieval, structured output, and joint knowledge and instruction-following evaluation. Our contributions are as follows:
\begin{enumerate}[leftmargin=*,itemsep=2pt]
    \item We introduce \listqa, a benchmark of 9,045 human-curated questions that require retrieving multiple facts and composing them into structured lists. The dataset spans two subsets: \simplelistqa{} (single-faceted flat lists) and \complexlistqa{} (multi-faceted hierarchical lists, see Fig.~\ref{fig:list_qa_examples}) grounded in Wikipedia tables across eight topical categories and 60+ countries. An evergreen design ensures ground truth remains valid without periodic updates, with temporal emphasis on 2017--2024 to probe recall of recent, less-saturated knowledge.
    \item We propose an \judgef{} metric that \textbf{(1)} optimally matches each predicted list element to its closest reference element, and \textbf{(2)} uses an LLM to grade correctness, capturing factual accuracy and penalizing hallucination. We validate this metric against human judgments and complement it with four additional metrics covering coverage, precision, format compliance, and lexical overlap.
    \item We evaluate 34 models from 8 families across 42 configurations, spanning standard and thinking-enabled inference. The best standard-inference model does not exceed 37\% \judgef{} score; multi-faceted questions are harder across every model; and thinking universally improves factual accuracy but has architecture and model family-dependent effects on format compliance. 
    An oracle RAG experiment shows that supplying the source page yields large gains for every model, identifying parametric recall as the primary bottleneck. Extended analyses covering base vs.\  instruction-tuned checkpoints and token efficiency are provided in the appendix.
\end{enumerate}

\begin{figure}[t]
  \centering
  \begin{minipage}[t]{0.48\textwidth}
    \fboxsep=5pt
    \fbox{\parbox{\dimexpr\linewidth-2\fboxsep-2\fboxrule}{\footnotesize
      \textbf{Example: \simplelistqa{}}
      \smallskip\hrule\smallskip
      \textbf{Q:} Which songs by Bruno Mars were nominated for the MTV Video Music Award for Best Choreography between the years 2010 and 2024, where he was both the performer and the choreographer?
      \medskip

      \textbf{A:} The MTV Video Music Award for Best Choreography is a craft award given to the artist, the artist's manager, and choreographer of the music video.

The following songs by Bruno Mars were nominated for the MTV Video Music Award for Best Choreography between the years 2010 and 2024, where he was both the performer and the choreographer:
      \begin{enumerate}[nosep, leftmargin=1.5em, label=\arabic*.]
        \item The Lazy Song
        \item Treasure
        \item Finesse (Remix)
      \end{enumerate}
    }}
  \end{minipage}\hfill
  \begin{minipage}[t]{0.48\textwidth}
    \fboxsep=5pt
    \fbox{\parbox{\dimexpr\linewidth-2\fboxsep-2\fboxrule}{\footnotesize
      \textbf{Example: \complexlistqa{}}
      \smallskip\hrule\smallskip
      \textbf{Q:} In 2023 and 2024, for which awards was \textit{John Wick: Chapter 4} nominated in Best Stunts, who was the recipient and what was the result?
      \medskip

      \textbf{A:} The awards for which \textit{John Wick: Chapter 4} was nominated in Best Stunts are:
      \begin{enumerate}[nosep, leftmargin=1.5em, label=\arabic*.]
        \item Hollywood Critics Assoc.\ Midseason Awards
          \begin{itemize}[nosep, leftmargin=1em, label=$\bullet$]
            \item \textit{recipient:} John Wick: Chapter 4
            \item \textit{result:} Won
          \end{itemize}
        \item Las Vegas Film Critics Society
          \begin{itemize}[nosep, leftmargin=1em, label=$\bullet$]
            \item \textit{recipient:} John Wick: Chapter 4
            \item \textit{result:} Nominated
          \end{itemize}
        \item St.\ Louis Film Critics Assoc.
          \begin{itemize}[nosep, leftmargin=1em, label=$\bullet$]
            \item \textit{recipient:} Scott Rogers \& Stephen Dunlevy
            \item \textit{result:} Runner-up
          \end{itemize}
        \item Astra Film \& Creative Arts Awards
          \begin{itemize}[nosep, leftmargin=1em, label=$\bullet$]
            \item \textit{recipient:} John Wick: Chapter 4
            \item \textit{result:} Won
          \end{itemize}
      \end{enumerate}
    }}
  \end{minipage}
  \caption{Examples from the \listqa{} dataset.}
  \label{fig:list_qa_examples}
\vspace{-10pt}
\end{figure}

%% file: sections/dataset.tex
\section{Dataset}
\label{sec:dataset}

\subsection{Task Definition}
\label{sec:task_definition}

Given a factual question $q$ whose correct answer is a list of $n$ elements $\mathcal{L} = \{l_1, l_2, \ldots, l_n\}$ where $3 \leq n \leq 10$, the task requires a model to generate a structured list response $\hat{\mathcal{L}} = \{\hat{l}_1, \hat{l}_2, \ldots, \hat{l}_m\}$. Each element $l_i$ may be a simple entity (single-faceted) or a composite of a primary entity with associated sub-attributes (multi-faceted). The model is evaluated on the correctness, completeness, and structural formatting of $\hat{\mathcal{L}}$ with respect to $\mathcal{L}$.

\subsection{Data Collection}
\label{sec:data_collection}

\textbf{Source material.} All questions and answers are grounded in Wikipedia tables, chosen as the reference source for several reasons. First, Wikipedia serves as a common knowledge source across LLM training pipelines, providing a fair evaluation baseline. Second, tabular data naturally lends itself to questions with list-formatted answers, as multiple rows of a table can satisfy a given query criterion. Third, using a single, publicly available source eliminates concerns about proprietary data access and licensing disparities across models.

\subsubsection{Annotation Process.} 

Human annotators were instructed to: (i) identify Wikipedia tables covering popular entities and topics across eight categories: Entertainment (including Music), Sports, Politics, Movies, Lifestyle, Food, Travel, and Places of Interest; (ii) formulate questions whose answers require retrieving information from one or more columns of the table; and (iii) compose reference answers in list format with 3 to 10 elements. The lower bound of 3 avoids degenerate cases that reduce to single-answer QA, while the upper bound of 10 prevents excessively long lists that test exhaustive enumeration rather than targeted knowledge retrieval. Complete annotation guidelines are provided in App.~\ref{sec:appendix_guidelines}. 

\textbf{Question types.} The dataset comprises two question types, defined based on their complexity: \textit{(1) Single-faceted} questions require retrieving entities that satisfy one or more filtering conditions over the source table. The answer is a flat numbered list of entity names. \textit{(2) Multi-faceted} questions often require reasoning across multiple columns and additionally request structured sub-attributes (e.g., recipient, result) for each retrieved entity. The answer is a hierarchical list whose elements each contain multiple fields. These questions may further incorporate exclusion or negation constraints that require filtering across several dimensions simultaneously.

\textbf{Temporal constraints.} To ensure the dataset remains factually valid over time, annotators were instructed to include explicit time constraints in questions where applicable. Rather than asking \textit{``What are the Flying Star Cafe locations?''} (which may change), questions are framed as \textit{``Which Flying Star Cafe locations opened from 2000 to 2024?''}.

\textbf{Temporal distribution.} The dataset covers years from 2000 through 2024, with deliberate emphasis on 2017--2024 to probe recall of recent, less-saturated knowledge. The year 2024 is especially informative, as publicly available training data for it is substantially more restricted than for earlier years. All answers are validated as factually correct as of December 31, 2024.

\textbf{Objectivity constraints.} All questions must be close-ended and objective. Subjective questions (e.g., \textit{``top 10 vegan restaurants in NYC''}) are excluded, as disambiguating via a specific source would test website knowledge rather than factual recall and unfairly disadvantage models not trained on that source. This ensures unambiguous, source-independent evaluation.

\textbf{Ordering.} We note that not all lists are required to be ordered, for example \textit{“In 2022 what Artists were nominated for the MTV Video Music Award for Best Pop but did not win?”}. However, \textit{“Who were the first 8 presidents of the US Senate, what were their exact positions, and what were their terms?”} is an ordered list, so are chronological and sequential events; annotators are instructed to tag the requirement for ordering.

\subsection{Dataset Statistics}
\label{sec:dataset_statistics}

\begin{figure*}[t]
    \centering
    \begin{minipage}[b]{0.33\textwidth}
        \centering
        \includegraphics[width=\textwidth]{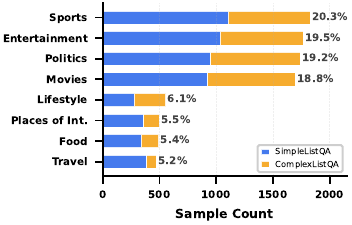}
    \end{minipage}%
    \hfill
    \begin{minipage}[b]{0.33\textwidth}
        \centering
        \includegraphics[width=\textwidth]{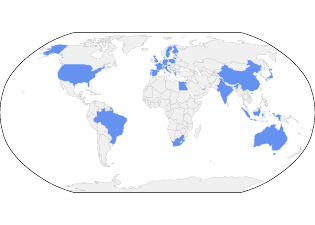}
    \end{minipage}%
    \hfill
    \begin{minipage}[b]{0.33\textwidth}
        \centering
        \includegraphics[width=\textwidth]{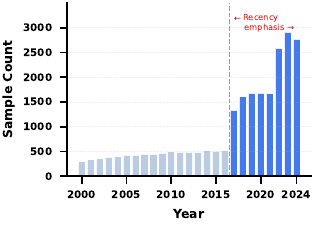}
    \end{minipage}

    \vspace{4pt}
    \footnotesize
    \setlength{\tabcolsep}{4pt}
    \renewcommand{\arraystretch}{0.95}
    \begin{tabular}{lclc}
      \toprule
      \textbf{Property} & \textbf{Value} & \textbf{Property} & \textbf{Value} \\
      \midrule
      Total samples & 9,045 & Elements per answer & 3--8 \\
      \simplelistqa{}/\complexlistqa{} & 59.5\% / 40.5\% & Mean elements & 4.67 \\
      Unique Wikipedia tables & 3,765 & Median elements & 4 \\
      Year-tagged / Timeless & 70.1\% / 29.9\% & Sub-elements per item (multi) & 1--10 \\
      Ordered / Unordered & 14.6\% / 85.4\% & Mean sub-elements & 8 \\
      \bottomrule
    \end{tabular}

    \caption{Overview of the \listqa{} dataset. \textbf{Top left:} Distribution across 8 topical categories, split by \simplelistqa{} and \complexlistqa{}. \textbf{Top center:} Geographic coverage spanning 60+ countries across 6 continents. \textbf{Top right:} Temporal distribution; bars after the dashed line (2017--2024) reflect the deliberate recency emphasis. \textbf{Bottom:} Key dataset statistics.}
    \label{fig:dataset_overview}
\vspace{-10pt}
\end{figure*}

The full dataset contains \textbf{9,045} question-answer pairs sourced from \textbf{3,765} unique Wikipedia tables. Fig.~\ref{fig:dataset_overview} provides a comprehensive overview. The four largest topical categories are Sports, Entertainment, Politics, and Movies; each contributes roughly one-fifth of the dataset, with ratios fairly maintained across both subsets. The dataset covers entities from over 60 countries across six continents, and its temporal distribution is deliberately skewed toward 2017--2024 to test recall of recent knowledge. The majority of Wikipedia tables contribute one or two questions, with a long tail extending to 28 questions per table. Each sample is released with rich metadata including the source Wikipedia URL, topical category, temporal range, question type, and ordering requirement, enabling independent verification of ground truth and fine-grained analysis along any dimension. Additional dataset statistics are provided in App.~\ref{sec:appendix_additional_data_stats}.

\subsection{Data Quality}
\label{sec:data_quality}

We employed a four-stage quality assurance process. (1) Two pilot rounds calibrated annotator understanding (pilot data excluded from final dataset). (2) Each question–answer pair was then authored by one out of 44 annotators and independently validated by a second annotator with no involvement in its creation, achieving 100\% coverage. (3) Validators flagged issues according to the taxonomy in Tab.~\ref{tab:validation_errors} (App.~\ref{sec:appendix_validation_errors}); annotations requiring significant rework were discarded (10\% discard rate), and the remainder were corrected for a final review pass. (4) All corrections were verified by at least one additional reviewer, with two reviewers covering 20\% of corrections. In total, every sample was assessed by at least three independent people.

\textbf{Reliability estimation.} Standard inter-annotator agreement metrics such as Cohen's $\kappa$ assume that multiple annotators independently label the same item, making them inapplicable to constructive annotation tasks where each annotator generates a unique question–answer pair rather than classifying a shared input. We therefore quantify annotation reliability through cross-validation outcomes, which serve as a functionally equivalent quality signal. Of all annotations submitted to cross-validation, 39\% were approved without any flags, 51\% required minor corrections (predominantly typographical or phrasing issues), and 10\% were discarded as requiring substantial rework. Among the error types identified in flagged annotations, factual accuracy errors constituted only 10\% of all flags, while the remainder were surface-level issues (typos, vague phrasing, formatting). To measure the residual error rate directly, we audited 481 questions stratified across all 8 categories and both question types, verifying each reference answer against its source Wikipedia table. We found errors in 8 questions (1.7\%; 95\% Wilson CI [0.8\%, 3.2\%]), mostly caused by ambiguous question framing rather than factual mistakes (App.~\ref{sec:appendix_validation_errors}).

%% file: sections/evaluation.tex
\section{Evaluation Framework}
\label{sec:evaluation}

We design five complementary metrics, each computed per sample and averaged over the dataset. Drawing on the notion of task demands~\citep{hu2024auxiliary}, they span demand levels, from format-independent \relaxedrecall{}/\relaxedprecision{} (reduced) to structure-dependent \formatmetric{} and \judgef{} (greater).

\subsection{Metrics}
\label{sec:metrics}

\subsubsection{List Element Parser}
\label{sec:list_ele_parser}
Our consistent formatting protocol enables a regex parser that extracts individual list elements from both reference answers and predicted answers, avoiding coarse-grained full-list comparison in a single evaluation pass. Details on the regex pattern are provided in App.~\ref{sec:appendix_list_element_parser}.

\subsubsection{Recall-focused Knowledge Metrics}
\label{sec:recall_metrics}
We design a metric to judge how accurately the model is able to retrieve and generate factual information decoupled from the formatting capabilities. This is done by first leveraging the parser from Sec.~\ref{sec:list_ele_parser} to extract each list element from the reference answer and then testing for the presence of each element independently in the unparsed model output as simple string matching. This judgment is made irrespective of whether the model prediction is formatted correctly in the list format. We compute both Precision and Recall here as \emph{Relaxed-Precision} and \emph{Relaxed-Recall}.

\subsubsection{List-Formatting Focused Metrics}

We use the parser from Sec.~\ref{sec:list_ele_parser} to determine whether the model’s prediction abides by the list formatting expected; we choose not to evaluate the actual content of the list elements as that is covered by the Recall-focused Knowledge metrics from Sec.~\ref{sec:recall_metrics}. The model is provided with a prompt to produce different styles of lists based on the constraints presented in the question and we expect the model to be able to reason on which list style to produce that aligns with the formatting styles of single and multi-faceted depending on the question. Subsequently, the model is evaluated on that type of list format for correctness: producing a flat list for a multi-faceted question or vice versa would be erroneous. Importantly, we require the model to produce at least 3 list elements for the output to be considered correct under this metric. This threshold mirrors the annotation constraint that all reference answers contain 3--10 elements (Sec.~\ref{sec:data_collection}), ensuring that models cannot achieve high format scores with trivially short outputs that sidestep the multi-element retrieval challenge. We term this the \textit{\formatmetric{}} metric.

\begin{figure}[t] %
    \centering
    \includegraphics[width=\textwidth]{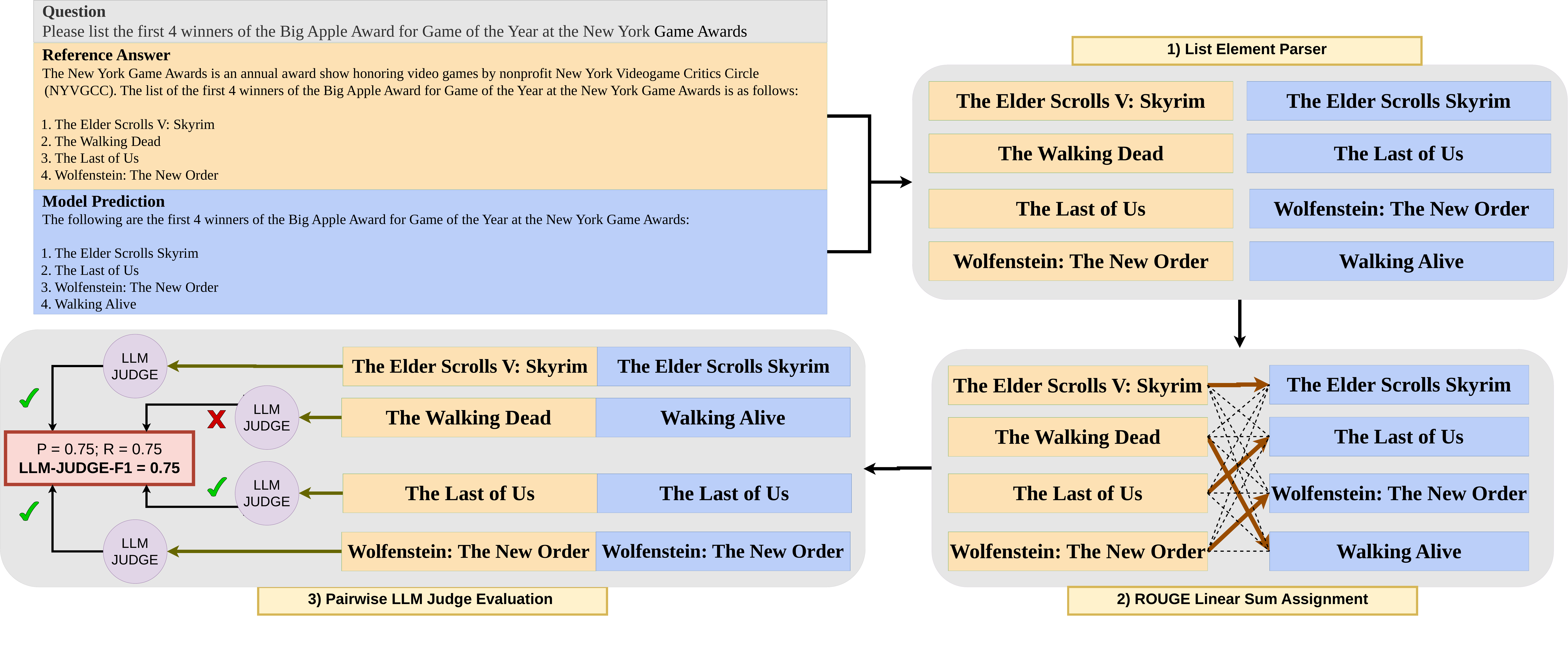} %
    \caption{Overview of the \judgef{} evaluation pipeline. Given a reference list and model prediction, \textbf{(1)} individual list elements are extracted via regex parsing, \textbf{(2)} pairwise ROUGE-1 scores form a cost matrix for optimal one-to-one element assignment via the Jonker-Volgenant algorithm, and \textbf{(3)} each matched pair is independently graded by an LLM judge as CORRECT or INCORRECT. Unmatched predicted elements are automatically graded INCORRECT. Precision and recall are computed over elements judged CORRECT.}
    \label{fig:llm_judge_metric} %
\vspace{-10pt}
\end{figure}

\subsubsection{Precision-focused Metric (Format + Knowledge)}
This metric jointly evaluates factual correctness and list formatting. Using the parser from Sec.~\ref{sec:list_ele_parser}, we independently extract individual list elements from both the reference answer and the model prediction, yielding two element lists. Rather than judging the predicted list holistically, which we find correlates poorly with human evaluation, we assess each predicted element against its reference counterpart using an LLM as a judge.

A naïve all-pairs comparison requires $\mathcal{O}(n^2)$ LLM judge calls, which is prohibitively expensive. We therefore reframe element matching as a \emph{linear sum assignment} problem: we construct a bipartite graph between reference and predicted elements, weight edges by pairwise ROUGE overlap, and solve for the optimal one-to-one assignment using the Jonker-Volgenant algorithm~\cite{Jonker1987ASA} as implemented in SciPy~\citep{ModifiedJonkerVolgenant2016, scipy_linearsum}. Although the assignment itself is $\mathcal{O}(n^3)$, it reduces LLM judge calls from $\mathcal{O}(n^2)$ to $\mathcal{O}(n)$, which is a substantial saving in practice.

Each matched pair is then independently scored by an LLM judge, GPT-OSS-120B, in all experiments reported in this paper (see Sec.~\ref{subsec:corr_human_perf} for selection rationale and robustness analysis across judge model sizes). We aggregate per-pair precision and recall into a final F1 score, which we term \emph{LLM-Judge-F1}. A visual representation of this workflow is shown in Fig. \ref{fig:llm_judge_metric}. As a lightweight alternative, we also report ROUGE F1 computed directly over the matched pairs, which serves as a cheaper but less discriminative proxy for early-stage evaluation. We call this \emph{ROUGE-F1}. A full scoring example is provided in App.~\ref{sec:appendix_scoring_example}.

\subsubsection{Ordering Metrics}

Given that we must honor the ordering of elements in the reference wherever applicable, we implement this metric by replacing the linear sum assignment algorithm with positional pairing: list elements are paired in the order the model predicted them and then graded by an LLM as a Judge. We call this metric \textit{LLM-Judge-F1-Ordered} and evaluate only a subset of \listqa{} samples with this metric, specifically those requiring implicit or explicit list sequencing.

\subsubsection{Correlation with Human Judgement}
\label{subsec:corr_human_perf}

\textbf{Judge model selection.} All experiments in this paper use GPT-OSS-120B as the default LLM judge. We chose GPT-OSS-120B for two reasons: (1) as an open-weight model, it ensures full reproducibility without dependence on proprietary API access, and (2) it is large enough to perform reliable semantic grading of factual equivalence across diverse topics. The judge prompt builds on the grading template from SimpleQA~\cite{wei2024simpleqa}, extended to handle list-structured, multi-element responses.

\textbf{Correlation with human judgment.} We validated \judgef{} against 3,313 human-graded list elements from 571 sample--model pairs (roughly 200 stratified questions, 3 models). \judgef{} correlates strongly with human F1 (Pearson $r = 0.965$; Spearman $\rho = 0.956$) (see App.~\ref{sec:appendix_human_eval}). Agreement is consistent across models, categories, and question types, and the metric is slightly conservative relative to human judgment. The ROUGE-based matching step aligns elements correctly 99.7\% of the time, and replacing it with embedding-based matching changes aggregate scores by less than 0.002 (App.~\ref{sec:appendix_matching_validation}).

\textbf{Robustness to judge model size.} To assess whether practitioners can substitute smaller, more affordable judges, we rescored all 9,045 samples from a single evaluated model using three additional reasoning-based open-weight judges: Qwen 3.5 27B, 9B, and 4B. All pairwise Spearman rank correlations exceed $\rho$ = 0.96, and mean absolute score differences are $\leq$ 0.024 across all judge pairs (App.~\ref{sec:appendix_judge_agreement}). Even the 4B judge produces near-identical sample rankings to the 120B default ($\rho$ = 0.972, MAD = 0.022), confirming that the metric is robust to judge scale and that lighter-weight alternatives are viable with minimal impact on conclusions.

\subsection{Evaluation Protocol}
\label{sec:eval_protocol}

\textbf{Instruction prompt.} Each question is preceded by a system instruction (prompt templates are provided in App.~\ref{sec:appendix_prompt}) that specifies the expected output format.

\textbf{Evaluation modes.} The benchmark supports two prompting modes: \textbf{zero-shot}, where only the instruction and question are provided, and \textbf{few-shot}, where 5 in-context examples are prepended. For few-shot evaluation, examples are drawn from the corresponding question type: single-faceted examples for \simplelistqa{}, multi-faceted for \complexlistqa{}, and a mix for the full dataset.

\textbf{Dataset variants.} Evaluation can be run on three splits: \texttt{full} (all 9,045 samples), \texttt{simple} (single-faceted only), and \texttt{complex} (multi-faceted only).

%% file: sections/experiments.tex
\section{Experiments}
\label{sec:experiments}

\subsection{Experimental Setup}
\label{sec:exp_setup}

We evaluate 34 models spanning 8 model families across 42 configurations, covering closed-source and open-source models and parameter counts from 3B to 675B. All models are evaluated using the zero-shot instruction prompt described in App.~\ref{subsec:appendix_prompt_zs} on the full \listqa{} dataset, with scores reported overall and separately for the \simplelistqa{} and \complexlistqa{} subsets. Unless otherwise noted, all evaluations use standard (non-thinking) inference. All results use GPT-OSS-120B as the LLM judge. Subsets of models are additionally evaluated with thinking enabled (Sec.~\ref{sec:analysis}) and with the source Wikipedia page supplied as context (oracle RAG; Sec.~\ref{sec:oracle_rag}).

\textbf{Metric selection.} We report two headline metrics: \textit{\judgef{}} and \textit{\formatmetric{}}. These two metrics capture the orthogonal capabilities that \listqa{} is specifically designed to disentangle: factual knowledge retrieval (\textit{\judgef{}}) and instruction-following compliance (\textit{\formatmetric{}}). LLM-Judge-F1 subsumes ROUGE-F1, which serves as a cheaper but less discriminative proxy for the same dimension, and captures semantic equivalences that \relaxedrecall{} and \relaxedprecision{} miss due to their reliance on string matching. Full results across all five metrics are provided in App.~\ref{sec:appendix_full_results}. Details on the evaluation infrastructure, including API providers and self-hosted model configurations, are provided in App.~\ref{sec:appendix_infrastructure}.

\subsection{Results \& Analysis}
\label{sec:main_results}
\label{sec:analysis}

Tab.~\ref{tab:main_results} presents results for all models evaluated under standard (non-thinking) inference, grouped by model family. We report \textit{\judgef{}} and \textit{\formatmetric{}} scores for \simplelistqa{} (S), \complexlistqa{} (C), and the full \listqa{} dataset (All).

\begin{table*}[t]
\caption{Results on \listqa{} under standard (non-thinking) inference. Models are grouped by family and sorted by overall \textit{LLM-Judge-F1} within each group. \textbf{LLM-Judge-F1}: LLM-as-judge semantic F1 measuring factual accuracy. \textbf{Format}: \formatmetric{} score. S = \simplelistqa{}, C = \complexlistqa{}. Per-sample distribution details are discussed in App.~\ref{subsec:appendix_persample_score_dist}.}
\label{tab:main_results}
\centering
\scriptsize
\setlength{\tabcolsep}{3pt}
\begin{minipage}[t]{0.48\textwidth}
\centering
\textbf{Closed-source models}\\[4pt]
\begin{tabular}{@{}lccccccc@{}}
\toprule
& \multicolumn{3}{c}{\textbf{LLM-Judge-F1} $\uparrow$} & \multicolumn{3}{c}{\textbf{Format} $\uparrow$} \\
\cmidrule(lr){2-4} \cmidrule(lr){5-7}
\textbf{Model} & \textbf{All} & \textbf{S} & \textbf{C} & \textbf{All} & \textbf{S} & \textbf{C} \\
\midrule
\multicolumn{7}{@{}l}{\textit{GPT}~\cite{gpt54}} \\
\midrule
GPT-5.4              & 36.5 & 44.5 & 24.9 & 87.2 & 85.0 & 90.6 \\
GPT-5.4 mini         & 27.0 & 35.2 & 15.1 & 64.1 & 84.8 & 33.6 \\
GPT-5.4 nano         & 14.5 & 20.4 & 5.7  & 58.1 & 77.7 & 29.2 \\
\midrule
\multicolumn{7}{@{}l}{\textit{Claude}~\cite{claude47, claude46opus, claude46sonnet, claude45opus, claude45sonnet,claude45haiku}} \\
\midrule
Claude Opus 4.7      & 35.0 & 42.1 & 24.6 & 75.0 & 75.4 & 74.5 \\
Claude Opus 4.6      & 34.6 & 43.0 & 22.4 & 78.3 & 73.9 & 84.8 \\
Claude Opus 4.5      & 33.9 & 42.0 & 22.2 & 70.8 & 63.0 & 82.1 \\
Claude Sonnet 4.6    & 28.0 & 35.5 & 16.9 & 60.2 & 45.3 & 82.2 \\
Claude Sonnet 4.5    & 25.3 & 30.2 & 18.1 & 49.2 & 41.1 & 61.0 \\
Claude Haiku 4.5     & 11.1 & 14.1 & 6.7  & 49.1 & 58.6 & 35.2 \\
\midrule
\multicolumn{7}{@{}l}{\textit{Gemini}~\cite{gemini25}} \\
\midrule
Gemini 2.5 Flash      & 33.7 & 41.9 & 21.6 & 76.4 & 72.5 & 82.2 \\
Gemini 2.5 Flash Lite & 20.4 & 27.2 & 10.4 & 63.9 & 83.6 & 35.1 \\
\midrule
\multicolumn{7}{@{}l}{\textit{Nova}~\cite{nova1, nova2}} \\
\midrule
Nova Premier         & 31.6 & 39.0 & 20.6 & 60.7 & 60.6 & 61.0 \\
Nova Pro             & 22.0 & 29.3 & 11.4 & 65.6 & 53.8 & 83.0 \\
Nova Micro           & 12.0 & 17.5 & 4.0  & 56.8 & 79.1 & 24.1 \\
\bottomrule
\end{tabular}
\end{minipage}%
\hfill
\begin{minipage}[t]{0.48\textwidth}
\centering
\textbf{Open-source models}\\[4pt]
\begin{tabular}{@{}lccccccc@{}}
\toprule
& \multicolumn{3}{c}{\textbf{LLM-Judge-F1} $\uparrow$} & \multicolumn{3}{c}{\textbf{Format} $\uparrow$} \\
\cmidrule(lr){2-4} \cmidrule(lr){5-7}
\textbf{Model} & \textbf{All} & \textbf{S} & \textbf{C} & \textbf{All} & \textbf{S} & \textbf{C} \\
\midrule
\multicolumn{7}{@{}l}{\textit{DeepSeek}~\cite{liu2024deepseek, deepseekv32}} \\
\midrule
DeepSeek V3.1        & 29.3 & 36.7 & 18.3 & 73.3 & 76.0 & 69.4 \\
DeepSeek V3.2        & 20.3 & 26.5 & 11.1 & 54.8 & 71.9 & 29.8 \\
\midrule
\multicolumn{7}{@{}l}{\textit{Qwen}~\cite{qwen3next, qwen35, qwen36}} \\
\midrule
Qwen 3.5 397B-A17B (FP8)        & 24.8 & 31.6 & 14.8 & 73.0 & 69.0 & 78.8 \\
Qwen 3.5 122B-A10B        & 18.9 & 25.3 & 9.7  & 82.2 & 84.1 & 79.5 \\
Qwen 3 Next 80B-A3B      & 17.4 & 23.6 & 8.3  & 39.1 & 7.8  & 85.1 \\
Qwen 3.5 27B         & 14.8 & 20.1 & 7.1  & 71.9 & 71.6 & 72.4 \\
Qwen 3.5 35B-A3B         & 13.4 & 18.5 & 5.9  & 51.7 & 37.0 & 73.2 \\
Qwen 3.5 9B          & 8.3  & 12.0 & 2.9  & 62.0 & 65.5 & 56.8 \\
Qwen 3.5 4B          & 5.1  & 7.8  & 1.3  & 39.2 & 43.9 & 32.3 \\
\midrule
\multicolumn{7}{@{}l}{\textit{Mistral}~\cite{mistral3, ministral3, magistral}} \\
\midrule
Mistral Large 3 675B-A41B & 25.7 & 33.2 & 14.8 & 58.9 & 77.9 & 31.0 \\
Magistral Small 24B      & 15.8 & 21.8 & 7.0  & 59.6 & 48.0 & 76.8 \\
Ministral 3 14B      & 7.2  & 10.4 & 2.6  & 46.9 & 76.2 & 3.8  \\
Ministral 3 8B       & 4.8  & 6.9  & 1.7  & 45.7 & 71.0 & 8.6  \\
Ministral 3 3B       & 3.3  & 5.1  & 0.5  & 46.2 & 76.3 & 2.1  \\
\midrule
\multicolumn{7}{@{}l}{\textit{Gemma}~\cite{gemma4}} \\
\midrule
Gemma 4 31B          & 16.0 & 22.0 & 7.3  & 54.3 & 45.8 & 66.7 \\
Gemma 4 26B-A4B      & 13.8 & 19.0 & 6.0  & 52.5 & 44.2 & 64.6 \\
Gemma 4 E4B          & 6.2  & 9.1  & 1.9  & 33.6 & 44.9 & 17.0 \\
\bottomrule
\end{tabular}
\end{minipage}
\vspace{-10pt}
\end{table*}

\begin{figure}[t]
    \centering
    \vspace{-5pt}
    \begin{minipage}[t]{0.48\textwidth}
        \centering
        \includegraphics[height=6cm, width=\textwidth, keepaspectratio]{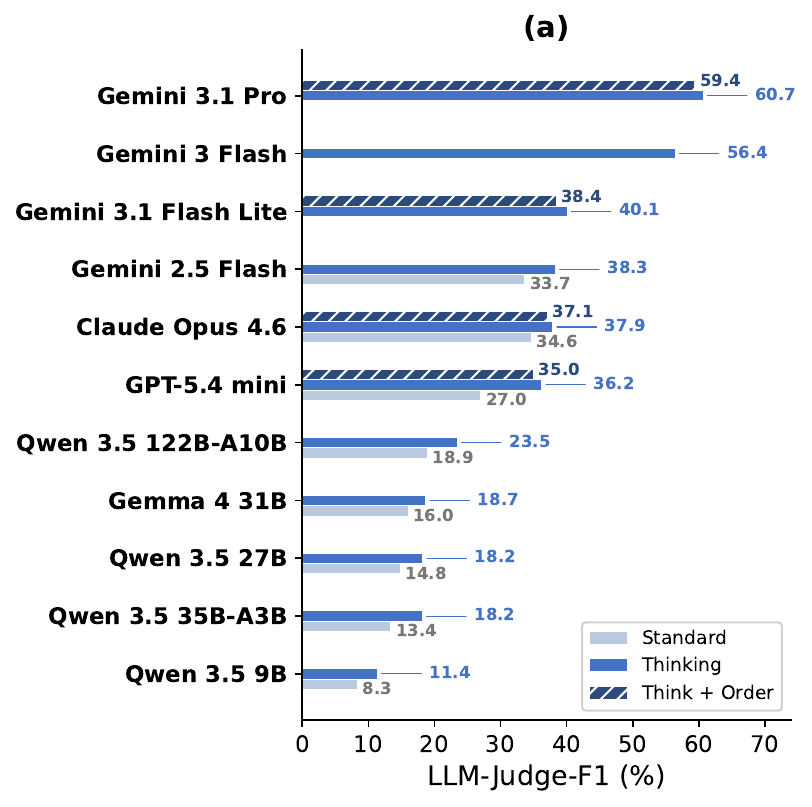}
    \end{minipage}%
    \hfill
    \begin{minipage}[t]{0.48\textwidth}
        \centering
        \includegraphics[height=6cm, width=\textwidth, keepaspectratio]{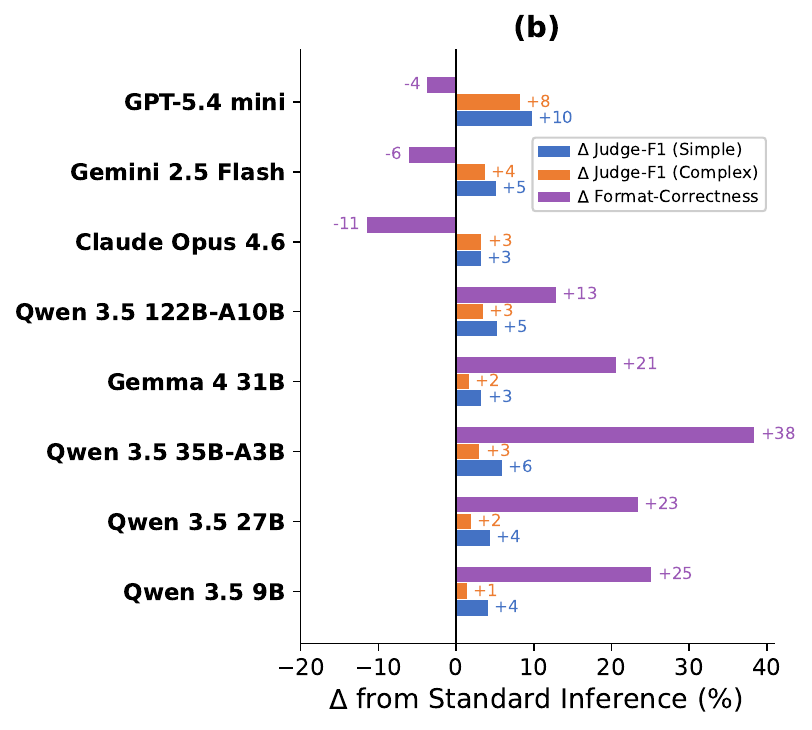}
    \end{minipage}
    \vspace{-4pt}
    \caption{(a) \textit{LLM-Judge-F1} scores across inference settings. Hatched bars show \textit{LLM-Judge-F1-Ordered}, a stricter metric penalizing incorrect sequencing for select samples (details in App.~\ref{sec:appendix_ordering}). (b) Improvement from thinking over standard inference. Thinking universally improves factual accuracy, but its effect on \formatmetric{} may be architecture and mode family-dependent. Exact scores and thinking levels are reported in App.~\ref{sec:appendix_thinking}.}
    \label{fig:thinking_results}
\vspace{-10pt}
\end{figure}

\textbf{The Knowledge Bottleneck.} Multi-element factual retrieval remains \textbf{largely unsolved: the best standard-inference model scores below 37\% \judgef{}}, with the top tier (GPT-5.4, Claude Opus variants, Gemini 2.5 Flash, Nova Premier) clustering in a narrow band that suggests a shared ceiling on parametric retrieval; the majority of models fall below 20\%. \textbf{Multi-faceted questions are consistently and substantially harder}: every model scores higher on \simplelistqa{} than \complexlistqa{}, with the gap widening for stronger models and the weakest collapsing to near-zero on Complex, confirming that compounding knowledge demands (reasoning across multiple dimensions, producing hierarchical sub-attributes, applying filtering constraints) amplify the bottleneck rather than merely adding formatting difficulty. Thinking-enabled inference (Fig.~\ref{fig:thinking_results}) reinforces this diagnosis: \textbf{thinking yields larger gains on \simplelistqa{} than \complexlistqa{}} (Fig.~\ref{fig:thinking_results}b), indicating it helps models organize knowledge they already possess but provides diminishing returns when parametric recall is the limiting factor. Gemini 3.1 Pro achieves the strongest overall result and is the only model to cross 50\% \judgef{} on \complexlistqa{} (App.~\ref{sec:appendix_thinking}), though only barely, underscoring multi-faceted retrieval's difficulty even for the strongest models. \textbf{Supplying the source page (oracle RAG) yields large gains for every model}, confirming parametric retrieval as the primary bottleneck (Sec.~\ref{sec:oracle_rag}). Additionally, \textbf{every model performs better on timeless than time-constrained questions} (App.~\ref{sec:appendix_year_range}), confirming that knowledge more heavily reinforced in training corpora is easier to retrieve parametrically. \textbf{Category-level difficulty} is model-invariant and intrinsic to training corpora distribution (App.~\ref{sec:appendix_category}). \textbf{A consistent geographic hierarchy} is observed (strongest on South American/North American content, weakest on African/Asian content), likely reflecting uneven reinforcement of regional knowledge during pretraining (App.~\ref{sec:appendix_continent}). Further, \textbf{list length} itself has only a \textbf{modest effect} (App.~\ref{sec:appendix_list_length}), and instruction tuning improves \judgef{} with thinking (App.~\ref{sec:appendix_base_models}).

\textbf{The Knowledge--Instruction Tension.} Orthogonal to the knowledge bottleneck, \textbf{factual retrieval and format compliance are partially decoupled}: several models achieve high \formatmetric{} while scoring low on \judgef{}, following structural instructions without the parametric knowledge to populate them, while others with moderate factual accuracy produce relatively poorly formatted output. This decoupling manifests as a \textbf{striking format asymmetry between question types}: many models score higher on \complexlistqa{} than \simplelistqa{} formatting, not because hierarchy is easier, but because they default to sub-bullet formatting unconditionally, inflating Complex scores while failing the flat-list requirement; the dominant error is producing the wrong list structure (App.~\ref{sec:appendix_format_errors}). Thinking sharpens this tension along model family lines: \textbf{closed-source models slightly lose format compliance while open-weight models gain it dramatically} (Fig.~\ref{fig:thinking_results}b), suggesting that extended reasoning can either refine or disrupt structural output depending on the model's baseline strategy. The stricter positional metric \orderedf{} further isolates sequencing as a distinct error source separable from both retrieval and formatting (App.~\ref{sec:appendix_ordering}), and generation verbosity does not correlate with factual accuracy (App.~\ref{sec:appendix_token_usage}). Qualitative examples illustrating these failure modes are provided in App.~\ref{sec:appendix_qualitative_examples}. The strongest models are those that jointly retrieve sufficient parametric knowledge \emph{and} adapt output structure to question complexity, \textbf{a frontier capability that no current model fully achieves}.

\subsection{Oracle RAG: Separating Knowledge from Composition}
\label{sec:oracle_rag}

Closed-book failures on \listqa{} conflate two error sources: not knowing the qualifying facts, and
failing to compose known facts into the requested structure. To separate them, we simulate perfect
retrieval by supplying the source Wikipedia page used during annotation as context, keeping the
list-formatting instructions and metrics unchanged (setup in App.~\ref{sec:appendix_oracle_rag}), for a subset of closed-source and open-weight models.

\begin{table}[t]
\caption{Oracle RAG vs.\ parametric-only \judgef{}. All = Overall, S = \simplelistqa{}, C =
\complexlistqa{}. \textbf{Context / Src.\ budget}: context window (tokens) and maximum number of source characters provided to open-weight models (App.~\ref{sec:appendix_oracle_rag}); closed-source models received the full page.}
\vspace{4pt}
\label{tab:oracle_rag}
\centering
\small
\setlength{\tabcolsep}{3pt}
\begin{tabular}{@{}lccccccccc@{}}
\toprule
& & & \multicolumn{3}{c}{\textbf{Parametric}} & \multicolumn{3}{c}{\textbf{Oracle RAG}} & \\
\cmidrule(lr){4-6} \cmidrule(lr){7-9}
\textbf{Model} & \textbf{Context} & \textbf{Src.\ budget} & \textbf{All} & \textbf{S} & \textbf{C} &
\textbf{All} & \textbf{S} & \textbf{C} & $\Delta$\textbf{All} \\
\midrule
\multicolumn{10}{@{}l}{\textit{Closed-source}} \\
Claude Opus 4.6       & \multicolumn{2}{c}{Full} & 34.6 & 43.0 & 22.4 & 84.7 & 86.0 & 82.8 & \textcolor{teal}{+50.1} \\
Claude Sonnet 4.6     & \multicolumn{2}{c}{Full} & 28.0 & 35.5 & 16.9 & 83.0 & 85.0 & 79.9 & \textcolor{teal}{+55.0} \\
GPT-5.4 mini          & \multicolumn{2}{c}{Full} & 27.0 & 35.2 & 15.1 & 75.8 & 79.8 & 70.0 & \textcolor{teal}{+48.8} \\
Gemini 2.5 Flash Lite & \multicolumn{2}{c}{Full} & 20.4 & 27.2 & 10.4 & 70.6 & 74.8 & 64.3 & \textcolor{teal}{+50.2} \\
\midrule
\multicolumn{10}{@{}l}{\textit{Open-weight}} \\
Gemma 4 31B         & 256K & 800K & 16.0 & 22.0 & 7.3 & 80.5 & 83.2 & 76.6 & \textcolor{teal}{+64.5} \\
Gemma 4 26B-A4B     & 256K & 800K & 13.8 & 19.0 & 6.0 & 73.3 & 76.8 & 68.1 & \textcolor{teal}{+59.5} \\
Magistral Small 24B & 128K & 400K & 15.8 & 21.8 & 7.0 & 63.5 & 67.8 & 57.3 & \textcolor{teal}{+47.7} \\
Ministral 3 8B      & 128K & 400K & 4.8  & 6.9  & 1.7 & 47.7 & 52.2 & 41.1 & \textcolor{teal}{+42.9} \\
Ministral 3 3B      & 128K & 400K & 3.3  & 5.1  & 0.5 & 37.6 & 43.7 & 28.7 & \textcolor{teal}{+34.3} \\
\bottomrule
\end{tabular}
\end{table}

Tab.~\ref{tab:oracle_rag} yields three takeaways. \textbf{Knowledge is the dominant bottleneck:} oracle context brings large gains for every model, far exceeding those from thinking (Fig.~\ref{fig:thinking_results}b), so inference-time reasoning cannot substitute for missing knowledge. \textbf{Closed-book scores reflect parametric knowledge, not reading ability:} model rankings reshuffle once the source page is provided; for example, Gemma 4 31B trails GPT-5.4 mini and Gemini 2.5 Flash Lite closed-book but overtakes both with context. \textbf{Composition is the residual bottleneck, especially at small scale:} oracle RAG does not saturate the benchmark, and with the source page in context the Simple--Complex gap becomes small for the strongest models but grows as models get smaller, even among models with identical context budgets. For frontier models, the multi-faceted penalty is thus largely a knowledge penalty; smaller models additionally struggle to reason across table dimensions and extract hierarchical sub-attributes even when the facts are in context.

%% file: sections/conclusion.tex
\section{Conclusion}
\label{sec:conclusion}

\listqa{} reframes factual evaluation as a multi-element problem: not whether a model can retrieve one fact, but whether it can recall, aggregate, and structurally organize many. Its evergreen design guarantees temporal validity at the list level, and its two-subset structure cleanly separates simple vs complex questions, and the provided metrics help separate knowledge retrieval from instruction following. The picture that emerges from 34 models across 42 configurations is stark. Parametric knowledge is the bottleneck: the best model (standard inference) scores below 37\%, oracle RAG adds 34--65 points, and reasoning provides diminishing returns where knowledge is absent. Format compliance is the second axis: models trade off retrieval accuracy against structural correctness rather than jointly optimizing both. No current model masters both dimensions simultaneously. Multi-element factual composition underlies summarization, report generation, and structured information retrieval, and \listqa{} demonstrates that current models have substantial room for improvement. We release the dataset and evaluation code.

\textbf{Limitations.} \listqa{} is English-only and US-centric (78\% of samples), does not reward abstention, incurs LLM-judge inference cost, and verifies reference completeness only against a single Wikipedia table; we discuss these limitations in detail in App.~\ref{sec:limitations}.

%% file: sections/appendix.tex
\clearpage
\etocdepthtag.toc{appendix}
\etocsettagdepth{mainpaper}{none}
\etocsettagdepth{appendix}{subsection}
\etocsettocstyle{\section*{Appendix Contents}}{}
\tableofcontents
\clearpage

\section{Limitations and Broader Impact}
\label{sec:limitations}

\paragraph{Limitations.} We identify the following limitations of the current work: \textit{\textbf{(1)}~English-only:} the benchmark is currently limited to English; multilingual extension is a natural next step. 
\textit{\textbf{(2)}~No abstention measurement:} our metrics penalize hallucinated elements, providing an indirect signal of overconfidence, but the benchmark does not yet explicitly reward models for recognizing the limits of their knowledge and declining to answer. 
\textit{\textbf{(3)}~Judge cost:} our primary metric relies on a 120B reasoning model as the judge, which adds non-trivial inference cost to each evaluation run; however, we report agreement scores with smaller judges in App.~\ref{sec:appendix_judge_agreement} to support practitioners who prefer a lighter-weight alternative. 
\textit{\textbf{(4)}~Judge noise:} as with any LLM-as-judge evaluation, the judge model may introduce noise, particularly when a model's prediction contains additional information beyond the reference that the judge cannot verify or refute without an external grounding source. While our metric correlates well with human evaluation overall, this inherent limitation of reference-based judging applies to \listqa{} as it does to other LLM-judged benchmarks. 
\textit{\textbf{(5)}~Geographic skew and language confound:} 78\% of samples concern US entities, so the overall score mainly reflects US-centric knowledge and may not generalize to regions with fewer samples (App.~\ref{sec:appendix_continent}). Moreover, all questions are posed in English, while facts about non-English-speaking regions are often documented mainly in local languages. The regional gaps we observe therefore combine regional knowledge with English-language coverage and cannot be attributed to either alone. Posing questions in local languages against localized sources would reduce this confound but not remove it: low-resource languages are themselves underrepresented in training, and localized probing adds a language-capability factor of its own. We therefore present the geographic results as a measured boundary that \listqa{} exposes rather than one it resolves.
\textit{\textbf{(6)}~Ground-truth completeness:} each reference answer is constructed from a single Wikipedia table, and validators checked for missing items against that table (Tab.~\ref{tab:validation_errors}). We did not independently audit completeness against external sources. If a valid element exists but is absent from the source table, for example because the table is incomplete or another source lists additional qualifying entities, a model that produces it would be penalized under \relaxedprecision{} and \judgef{}. The explicit temporal and categorical constraints in each question narrow the answer set and reduce this risk, and every sample includes its source URL so that such references can be audited and corrected in future releases.

\paragraph{Broader Impact.} \listqa{} encourages the development of models that are both factually comprehensive and precise, by explicitly measuring hallucination through precision and coverage through recall. The dataset is constructed entirely from publicly available Wikipedia content and contains no personally identifiable information, offensive content, or sensitive data; annotators were compensated in accordance with applicable labor standards. While improved factual retrieval could be misused to generate selectively curated information, we believe transparent evaluation of factual capabilities, including explicit hallucination measurement, is a net positive for responsible AI development.

\section{Annotation Guidelines}
\label{sec:appendix_guidelines}

We provide the complete annotation guidelines given to human annotators for the construction of \listqa{}.

\paragraph{Core Requirements}

\begin{enumerate}[leftmargin=*]
    \item \textbf{List format.} All reference answers must be a list of at least 3 and no more than 10 elements. This avoids single-element answers (which reduce to standard QA) and very large information dumps that are difficult for users to parse. Additionally, all single and multi-faceted list answers should follow rigid formatting guidelines to ensure consistency.

    \item \textbf{Close-ended and objective.} Questions must have unambiguous, objectively verifiable answers. Subjective questions (e.g., \textit{``What are the top 10 vegan restaurants in New York City''}) are excluded to prevent evaluation disputes arising from source-dependent rankings.

    \item \textbf{Wikipedia table reference.} Annotators must source questions from Wikipedia tables. Tables provide structured data that naturally supports list-formatted answers where multiple rows satisfy a query criterion.

    \item \textbf{Varying difficulty.} Questions should span a range of complexity:
    \begin{itemize}
        \item \textit{Single-faceted}: request a primary element from a single column.
        \item \textit{Multi-faceted}: add constraints across multiple columns and request secondary sub-elements, including filtering conditions (e.g., ``excluding publishers with `entertainment' in their name'').
    \end{itemize}

    \item \textbf{Temporal constraints.} Questions must include explicit time constraints where applicable to prevent reference answers from becoming outdated as the underlying data changes.

    \item \textbf{Temporal diversity.} Questions should be distributed across time periods from 2000 to 2024, including timeless questions whose answers are historically fixed. The distribution of these time periods should favor questions that are more recent but not be lopsided.

    \item \textbf{Answerable as of 2024.} All questions must be answerable as of December 31, 2024.
\end{enumerate}

\begin{table}[t]
  \caption{Validation round outcomes and error type distribution. The same sample may be associated with one or more error types}
  \label{tab:validation_errors}
  \centering
  \begin{tabular}{lrr}
    \toprule
    \multicolumn{3}{l}{\textit{Validation outcomes}} \\
    \midrule
    No issues found & 3,957 & 39\% \\
    Fixed & 5,088 & 51\% \\
    Discarded & 987 & 10\% \\
    \midrule
    \multicolumn{3}{l}{\textit{Error types identified}} \\
    \midrule
    Typo / grammar & 3,565 & 44\% \\
    Vague prompt & 2,502 & 31\% \\
    Sub-list issues & 1,020 & 13\% \\
    Data not accurate & 768 & 10\% \\
    Missing items & 532 & 7\% \\
    Year range error & 219 & 3\% \\
    Mis-matched order & 144 & 2\% \\
    \bottomrule
  \end{tabular}
\end{table}

\subsection{Validation Errors}
\label{sec:appendix_validation_errors}
During the validation process, annotators observed the following high level categories of errors as seen in Tab.~\ref{tab:validation_errors}. Categories or questions that required significant rewrite were discarded, constituting the 10\% discard rate and other annotations were fixed.

\paragraph{Independent error audit.} To measure the error rate of the released dataset directly, we audited a stratified random sample covering all 8 categories and both question types (5 random seeds of 100 questions each, yielding 481 unique questions after deduplication). Each reference answer was verified against its source Wikipedia table. We found errors in 8 of 481 questions (1.7\%; 95\% Wilson CI [0.8\%, 3.2\%]). These stemmed primarily from ambiguous question framing rather than factual mistakes.

\clearpage
\section{Additional Dataset Statistics}
\label{sec:appendix_additional_data_stats}

This section provides detailed breakdowns of the \listqa{} dataset composition, complementing the overview in Fig.~\ref{fig:dataset_overview}.

\paragraph{Geographic coverage.} Tab.~\ref{tab:geography_full} shows the full country-level distribution. The United States contributes the largest share with 7,049 samples (77.9\%), followed by 15 ``Global'' countries each contributing $\sim$120 samples across all major regions. Beyond these, the dataset includes countries across Europe, Asia, Africa, Australasia, and South America at smaller scales, ensuring representation from over 60 countries across 6 continents.

\begin{table}[ht]
\caption{Geographic distribution of samples across continents and countries (excl. USA}
\vspace{4pt}
\label{tab:geography_full}
\centering
\scriptsize
\setlength{\tabcolsep}{3pt}
\begin{tabular}{@{}lr|lr|lr|lr|lr|lr@{}}
\toprule
\multicolumn{2}{c|}{\textbf{Global (15)}} & \multicolumn{2}{c|}{\textbf{Europe}} & \multicolumn{2}{c|}{\textbf{Asia}} & \multicolumn{2}{c|}{\textbf{Africa}} & \multicolumn{2}{c|}{\textbf{Australasia}} & \multicolumn{2}{c}{\textbf{S. America}} \\
\midrule
Spain & 120 & Albania & 2 & Cambodia & 6 & Angola & 4 & New Zealand & 15 & General & 21 \\
Italy & 120 & Lithuania & 1 & Sri Lanka & 1 & Kenya & 5 & Papua N.G. & 3 & Chile & 9 \\
Germany & 119 & Andorra & 1 & Maldives & 4 & Seychelles & 2 & Cook Islands & 1 & Colombia & 14 \\
Portugal & 120 & Austria & 3 & Thailand & 12 & Madagascar & 3 & General & 1 & Bolivia & 3 \\
France & 120 & Greece & 1 & Uzbekistan & 2 & Algeria & 7 & & & Argentina & 6 \\
India & 120 & Spain & 1 & S. Korea & 10 & Ethiopia & 5 & & & Venezuela & 4 \\
UK & 120 & Switzerland & 1 & Bangladesh & 2 & Mauritius & 3 & & & Uruguay & 3 \\
Japan & 120 & Estonia & 1 & Vietnam & 2 & Nigeria & 3 & & & & \\
China & 119 & General & 3 & Philippines & 2 & Morocco & 8 & & & & \\
UAE & 120 & Denmark & 2 & Singapore & 4 & Tanzania & 3 & & & & \\
Indonesia & 119 & Sweden & 2 & China & 1 & General & 1 & & & & \\
Australia & 119 & Finland & 1 & Iran & 4 & Uganda & 1 & & & & \\
S. Africa & 120 & Poland & 1 & Afghanistan & 1 & & & & & & \\
Brazil & 120 & & & Armenia & 3 & & & & & & \\
Egypt & 120 & & & Nepal & 1 & & & & & & \\
\bottomrule
\end{tabular}
\end{table}

\begin{figure}[ht]
    \centering
    \begin{minipage}[b]{0.48\textwidth}
        \centering
        \includegraphics[width=\textwidth]{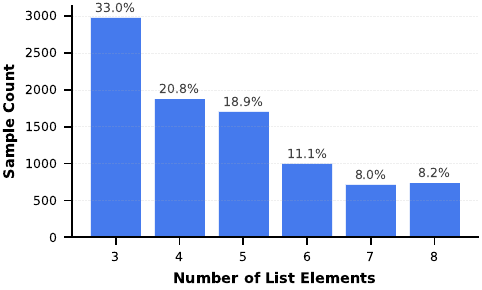}
    \end{minipage}%
    \hfill
    \begin{minipage}[b]{0.48\textwidth}
        \centering
        \includegraphics[width=\textwidth]{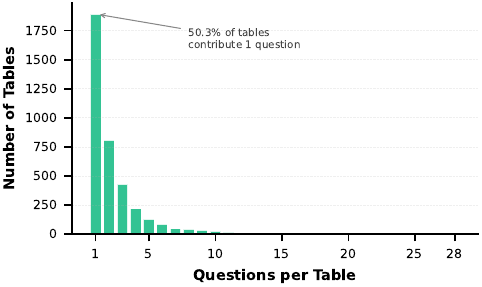}
    \end{minipage}
    \caption{\textbf{Left:} Distribution of list elements per answer. Most answers contain 3 elements; the distribution is right-skewed. \textbf{Right:} Questions per unique Wikipedia table. Over half of tables contribute a single question, with a long tail extending to 28.}
    \label{fig:additional_stats}
\end{figure}

\paragraph{List structure.} Fig.~\ref{fig:additional_stats} (left) shows the distribution of list elements per answer. Most common answer length is 3 elements, with a right-skewed distribution tapering toward the maximum of 8. The mean is 4.67 and the median is 4, indicating that the benchmark tests retrieval of a moderate number of facts rather than exhaustive enumeration.

\paragraph{Source diversity.} Fig.~\ref{fig:additional_stats} (right) shows the number of questions sourced from each unique Wikipedia table. Over half of tables contribute a single question, reducing the risk of topical clustering. The long tail (up to 28 questions per table) reflects the richness of certain source tables, particularly in Sports and Entertainment.

\paragraph{Category balance.} As shown in Fig.~\ref{fig:dataset_overview}, the category distribution is fairly maintained across both \simplelistqa{} and \complexlistqa{}, ensuring that neither subset is biased toward particular topics.

\clearpage
\section{List Element Parser}
\label{sec:appendix_list_element_parser}

\subsection{Overview}

Evaluating list-formatted outputs from large language models (LLMs) requires the reliable extraction of individual list elements from both model-generated predictions and gold-standard reference answers. This is a non-trivial parsing problem because LLM outputs exhibit a two-level hierarchical structure: \emph{top-level numbered items} that may optionally contain \emph{nested sub-elements} (key--value facets). We describe a regex-based list element parser that serves as the foundational component upon which all downstream \listqa{} metrics, \relaxedprecision{}, \relaxedrecall{}, \rougef{} element matching, and \judgef{} scoring, are built.

\subsection{List Format Specification}

In line with the list format guidelines provided to the annotators, the evaluation prompt instructs the model to produce lists in one of two canonical formats. Annotators' adherence to the guidelines helps us reliably parse list elements from the ground truth reference.

For \textbf{single-faceted} (simple) questions, the output format is as follows:

\begin{quote}
\texttt{1. List Element 1}\\
\texttt{2. List Element 2}\\
\texttt{3. List Element 3}
\end{quote}

\noindent For \textbf{multi-faceted} (complex) questions requiring structured sub-attributes, the output format is as follows:

\begin{quote}
\texttt{1. List Element 1}\\
\texttt{* Category 1: Sub-Element 1}\\
\texttt{* Category 2: Sub-Element 2}\\

\texttt{2. List Element 2}\\
\texttt{* Category 1: Sub-Element 1}\\
\texttt{* Category 2: Sub-Element 2}
    
\end{quote}

\noindent The parser must handle both formats uniformly, extracting top-level elements along with any nested sub-element structure. All answers are bounded between 3 and 10 main elements.

\subsection{Two-Stage Parsing Architecture}

Parsing is decomposed into two stages, each governed by its own regular expression.

\subsubsection{Stage 1: Top-Level List Element Extraction}

The first stage extracts all top-level numbered list items from the input string using:

\begin{equation}
\texttt{R}_1 = \texttt{\textbackslash n\textbackslash d+\textbackslash.\textbackslash s(.+?)(?=\textbackslash n\textbackslash d+\textbackslash.|\textbackslash Z)}
\label{eq:regex-top-level}
\end{equation}

\noindent applied with the \texttt{re.DOTALL} flag. Each match captures the full content of a list element (including any multi-line body and nested sub-elements) as a single contiguous block of text. Results are lowercased and stripped of leading/trailing whitespace.

\subsubsection{Stage 2: Multi-Faceted Sub-Element Extraction}

The second stage is applied to each individual element extracted by Stage~1. It identifies bullet-prefixed key--value sub-elements within the body of a list item:

\begin{equation}
\texttt{R}_2 = \texttt{\^{}\textbackslash *~(.*?):~(.*?)\$}
\label{eq:regex-sub-element}
\end{equation}

\noindent applied with the \texttt{re.MULTILINE} flag. The main element (the text before the first \texttt{*} bullet) is extracted via string splitting, while sub-elements are parsed into structured $\{\text{category} \rightarrow \text{value}\}$ dictionaries.

\subsection{Detailed Regex Analysis}

\subsubsection{Top-Level Pattern \texorpdfstring{$\texttt{R}_1$}{R1}}

Tab.~\ref{tab:regex-top-level} provides a component-by-component breakdown of the top-level extraction regex.

\begin{table}[ht]
\centering
\caption{Component analysis of the top-level list element regex $\texttt{R}_1$.}
\label{tab:regex-top-level}
\begin{tabular}{@{}ll@{}}
\toprule
\textbf{Component} & \textbf{Purpose} \\
\midrule
\texttt{\textbackslash n} & Anchors to a newline boundary, preventing false matches on \\
 & numbered references appearing mid-sentence. \\
\texttt{\textbackslash d+} & Matches one or more digits, accommodating lists of arbitrary \\
 & length (1--9, 10--99, 100+) without imposing an upper bound. \\
\texttt{\textbackslash .} & Matches the literal period delimiter, distinguishing numbered \\
 & list items from other digit-newline patterns. \\
\texttt{\textbackslash s} & Matches mandatory whitespace between the period and item text. \\
\texttt{(.+?)} & Non-greedy capture group. With \texttt{re.DOTALL}, the \texttt{.} \\
 & metacharacter matches newlines, enabling capture of multi-line \\
 & elements spanning nested sub-element lines. \\
\texttt{(?=\textbackslash n\textbackslash d+\textbackslash .|\textbackslash Z)} & Zero-width positive lookahead terminating capture at the start \\
 & of the next numbered item or end of string (\texttt{\textbackslash Z}). \\
\bottomrule
\end{tabular}
\end{table}

\subsubsection{Sub-Element Pattern \texorpdfstring{$\texttt{R}_2$}{R2}}

Tab.~\ref{tab:regex-sub-element} provides the corresponding breakdown for the sub-element extraction regex.

\begin{table}[ht]
\centering
\caption{Component analysis of the sub-element extraction regex $\texttt{R}_2$.}
\label{tab:regex-sub-element}
\begin{tabular}{@{}ll@{}}
\toprule
\textbf{Component} & \textbf{Purpose} \\
\midrule
\texttt{\^{}} & With \texttt{re.MULTILINE}, anchors to line start, matching only \\
 & bullets that begin a line. \\
\texttt{\textbackslash *~} & Matches the literal asterisk bullet marker followed by a space. \\
\texttt{(.*?)} (1st) & Non-greedy capture of the category/key name, stopping at the \\
 & first colon-space delimiter. \\
\texttt{:~} & Literal colon-space separator between category and value. \\
\texttt{(.*?)} (2nd) & Non-greedy capture of the value, terminating at end of line. \\
\texttt{\$} & With \texttt{re.MULTILINE}, anchors to line end, bounding value capture. \\
\bottomrule
\end{tabular}
\end{table}

\clearpage
\section{Scoring Example}
\label{sec:appendix_scoring_example}

We illustrate how each metric is computed using two real examples from the benchmark, one each from single-faceted and multi-faceted.

\paragraph{Single-faceted example.}

\textit{Question:} Which films won the Silwerskerm Award for Best Feature Film between 2017 and 2022?

\smallskip
\noindent\fbox{\begin{minipage}[t]{0.46\textwidth}
\small
\textbf{Reference (4 elements):}\\[2pt]
\texttt{1. Vaselinetjie}\\
\texttt{2. Kanarie}\\
\texttt{3. Poppie Nongena}\\
\texttt{4. Gaia}
\end{minipage}%
\hfill\vrule\hspace{8pt}
\begin{minipage}[t]{0.46\textwidth}
\small
\textbf{Model prediction (4 elements):}\\[2pt]
\texttt{1. Wonderlus}\\
\texttt{2. Gaia}\\
\texttt{3. Poppie Nongena}\\
\texttt{4. Vaselinetjie}
\end{minipage}}

\vspace{4pt}
The model retrieves 3 of 4 correct names but hallucinates ``Wonderlus'' (which was nominated in 2017 but did not win) and misses ``Kanarie''.  The order differs from the reference, but order is irrelevant for this question.

\paragraph{Multi-faceted example.}

\textit{Question:} Which years between 2017 and 2024 were works performed by Dua Lipa nominated for the MTV Video Music Award for Best Choreography, and what was the name of the work?

\smallskip
\noindent\fbox{\begin{minipage}[t]{0.46\textwidth}
\small
\textbf{Reference (4 elements):}\\[2pt]
\texttt{1. 2018}\\
\texttt{~~* Name of the work: ``IDGAF''}\\[2pt]
\texttt{2. 2020}\\
\texttt{~~* Name of the work: ``Physical''}\\[2pt]
\texttt{3. 2023}\\
\texttt{~~* Name of the work: ``Dance the Night''}\\[2pt]
\texttt{4. 2024}\\
\texttt{~~* Name of the work: ``Houdini''}
\end{minipage}%
\hfill\vrule\hspace{8pt}
\begin{minipage}[t]{0.46\textwidth}
\small
\textbf{Model prediction (2 elements):}\\[2pt]
\texttt{1. 2018}\\
\texttt{~~* Work: ``IDGAF''}\\[2pt]
\texttt{2. 2024}\\
\texttt{~~* Work: ``Houdini''}
\end{minipage}}

\vspace{4pt}
The model produces only 2 of 4 elements, missing 2020 (``Physical'') and 2023 (``Dance the Night''). Both predicted elements are factually correct with no hallucination, but the output falls below the 3-element minimum.

Additionally, only content parsed from the structured list for a given element is considered; information that appears outside the expected list structure (e.g., in free-text following the list) is excluded from evaluation and deemed incorrect. Consider the following example for the question, \emph{What are the NCAA Division I men's basketball teams that were undefeated in the tournaments between 2000 and 2024, as well as the year, the record, and the result?}

\smallskip
\noindent\fbox{\begin{minipage}[t]{0.46\textwidth}
\small
\textbf{Reference Answer:}\\[2pt]
\texttt{1. Wichita State}\\
\texttt{~~* Year: 2014, Record: 34-0}\\[2pt]
\texttt{~~* Result: Lost R32 to Kentucky}\\[2pt]
\texttt{2. Kentucky}\\
\texttt{~~* Year: 2015, Record: 34-0}\\[2pt]
\texttt{~~* Result: Lost SF to Wisconsin}\\[2pt]
\texttt{3. Gonzaga}\\
\texttt{~~* Year: 2021, Record: 26-0}\\[2pt]
\texttt{~~* Result: Lost final to Baylor}\\[2pt]
\end{minipage}%
\hfill\vrule\hspace{8pt}
\begin{minipage}[t]{0.46\textwidth}
\small
\textbf{Model prediction:}\\[2pt]
\texttt{1. Wichita State}\\
\texttt{~~* Year: 2014, Record: 34-0}\\[2pt]
\texttt{2. Kentucky}\\
\texttt{~~* Year: 2015, Record: 34-0}\\[2pt]
\texttt{3. Gonzaga}\\
\texttt{~~* Year: 2021, Record: 26-0}\\[2pt]
\texttt{``Wichita lost to Kentucky in R32, Kentucky lost to Wisconsin in the semis, Gonzaga lost the final to Baylor.''}
\vspace{4pt}
\end{minipage}}

\smallskip

As the results appear in trailing free text rather than as list sub-elements, all three \texttt{Result} fields are marked \textsc{incorrect} as factual knowledge must be delivered in the prescribed format to receive credit.

\subsection{\relaxedrecall{}}

\relaxedrecall{} measures the fraction of reference elements found \emph{anywhere} in the model's full output text via substring matching, without requiring the output to be correctly formatted. Before comparison, both texts are normalized: the reference undergoes whitespace normalization (trailing spaces before newlines are removed) and lowercasing; the prediction text is lowercased. For multi-faceted questions, the main element and each sub-element \emph{value} are checked independently against the full prediction text.

\begin{minipage}[t]{0.48\textwidth}
\centering
\small
\textbf{Single-faceted}\\[4pt]
\begin{tabular}{@{}lc@{}}
\toprule
\textbf{Reference item} & \textbf{Found?} \\
\midrule
vaselinetjie & \cmark \\
kanarie & \xmark \\
poppie nongena & \cmark \\
gaia & \cmark \\
\bottomrule
\end{tabular}\\[4pt]
\textbf{Recall = 3/4 = 0.75}
\end{minipage}%
\hfill\vrule\hfill
\begin{minipage}[t]{0.48\textwidth}
\centering
\small
\textbf{Multi-faceted}\\[4pt]
\begin{tabular}{@{}lc@{}}
\toprule
\textbf{Reference item} & \textbf{Found?} \\
\midrule
\textit{main:} 2018 & \cmark \\
~~\textit{sub-value:} ``idgaf'' & \cmark \\
\textit{main:} 2020 & \xmark \\
~~\textit{sub-value:} ``physical'' & \xmark \\
\textit{main:} 2023 & \xmark \\
~~\textit{sub-value:} ``dance the night'' & \xmark \\
\textit{main:} 2024 & \cmark \\
~~\textit{sub-value:} ``houdini'' & \cmark \\
\bottomrule
\end{tabular}\\[4pt]
\textbf{Recall = 4/8 = 0.50}
\end{minipage}

\vspace{6pt}
In the single-faceted case, substring matching works well for entity names; 3 out of 4 are found directly. In the multi-faceted case, the scorer checks each sub-element's \emph{value} (e.g., \texttt{"idgaf"}) rather than the full key-value string, so matching succeeds for the two predicted elements. However, the two missing elements (2020/``Physical'' and 2023/``Dance the Night'') are entirely absent from the prediction, yielding 4 hits out of 8 total items.

\subsection{\relaxedprecision{}}

\relaxedprecision{} measures the fraction of \emph{predicted} list elements that match a reference element, penalizing hallucinated items. The same normalization is applied as for \relaxedrecall{}. For multi-faceted predictions, the main element and each sub-element value are checked independently against the reference text, mirroring the recall computation. If a direct substring match fails, a fallback check tests whether any flattened reference element is contained within the predicted item.

\begin{minipage}[t]{0.48\textwidth}
\centering
\small
\textbf{Single-faceted}\\[4pt]
\begin{tabular}{@{}lc@{}}
\toprule
\textbf{Predicted item} & \textbf{In reference?} \\
\midrule
wonderlus & \xmark~(hallucinated) \\
gaia & \cmark \\
poppie nongena & \cmark \\
vaselinetjie & \cmark \\
\bottomrule
\end{tabular}\\[4pt]
\textbf{Precision = 3/4 = 0.75}
\end{minipage}%
\hfill\vrule\hfill
\begin{minipage}[t]{0.48\textwidth}
\centering
\small
\textbf{Multi-faceted}\\[4pt]
\begin{tabular}{@{}lc@{}}
\toprule
\textbf{Predicted item} & \textbf{In reference?} \\
\midrule
\textit{main:} 2018 & \cmark \\
~~\textit{sub-value:} ``idgaf'' & \cmark \\
\textit{main:} 2024 & \cmark \\
~~\textit{sub-value:} ``houdini'' & \cmark \\
\bottomrule
\end{tabular}\\[4pt]
\textbf{Precision = 4/4 = 1.00}
\end{minipage}

\vspace{6pt}
The single-faceted example shows precision catching the ``Wonderlus'' hallucination. The multi-faceted example achieves perfect precision (both predicted elements are correct) but this masks the fact that 2 of 4 reference elements are entirely missing. Precision and recall are complementary: high precision with low recall indicates a conservative model that produces few but accurate elements.

\subsection{\formatmetric{}}

\formatmetric{} evaluates structural compliance without judging factual content. The prediction is parsed into list elements using a regex pattern. If fewer than 3 elements are parsed, the score is automatically 0. Otherwise, for each parsed element, the scorer checks whether sub-elements are present: single-faceted questions expect no sub-bullets, multi-faceted questions expect sub-bullets. The score is the fraction of correctly structured elements.

\begin{minipage}[t]{0.48\textwidth}
\centering
\small
\textbf{Single-faceted}\\[4pt]
\begin{tabular}{@{}lc@{}}
\toprule
\textbf{Predicted element} & \textbf{Correct?} \\
\midrule
wonderlus & \cmark~(flat, no subs) \\
gaia & \cmark~(flat, no subs) \\
poppie nongena & \cmark~(flat, no subs) \\
vaselinetjie & \cmark~(flat, no subs) \\
\bottomrule
\end{tabular}\\[4pt]
4 elements $\geq$ 3 minimum \cmark\\
\textbf{Format = 4/4 = 1.00}
\end{minipage}%
\hfill\vrule\hfill
\begin{minipage}[t]{0.48\textwidth}
\centering
\small
\textbf{Multi-faceted}\\[4pt]
\begin{tabular}{@{}lc@{}}
\toprule
\textbf{Predicted element} & \textbf{Correct?} \\
\midrule
2018 / * Work: ``IDGAF'' & \cmark~(has subs) \\
2024 / * Work: ``Houdini'' & \cmark~(has subs) \\
\bottomrule
\end{tabular}\\[4pt]
2 elements $<$ 3 minimum \xmark\\
\textbf{Format = 0.00}
\end{minipage}

\vspace{6pt}
The single-faceted prediction uses the correct flat format and meets the 3-element minimum, as instructed in the prompt (see App.~\ref{sec:appendix_prompt}). The multi-faceted prediction uses the correct sub-bullet structure for both elements, but produces only 2 elements (below the required minimum of 3), so it receives a format score of 0 regardless of structural correctness. This design choice ensures that models cannot achieve high \formatmetric{} by producing trivially short lists.

\subsection{\rougef{}}

ROUGE F1 handles element reordering and partial lexical overlap via a two-step process. First, each list element is flattened into a single string by concatenating the main element and all sub-element values (e.g., \texttt{2018 "idgaf"}). A ROUGE-1 F-measure is computed between every predicted--reference pair, forming a cost matrix. The Jonker-Volgenant algorithm finds the optimal one-to-one assignment that maximizes total ROUGE-1 score. A pair is considered matched if its ROUGE-1 score $\geq$ 0.5. Precision and recall are computed over matched pairs.

\begin{minipage}[t]{0.48\textwidth}
\centering
\small
\textbf{Single-faceted}\\[4pt]
\begin{tabular}{@{}llc@{}}
\toprule
\textbf{Predicted} & \textbf{Matched ref.} & \textbf{R-1} \\
\midrule
wonderlus & \textit{(none $\geq$ 0.5)} & 0.00 \\
gaia & gaia & 1.00 \cmark \\
poppie nongena & poppie nongena & 1.00 \cmark \\
vaselinetjie & vaselinetjie & 1.00 \cmark \\
\bottomrule
\end{tabular}\\[4pt]
$P = 3/4,\; R = 3/4$\\
\textbf{\rougef{} = 0.75}
\end{minipage}%
\hfill\vrule\hfill
\begin{minipage}[t]{0.48\textwidth}
\centering
\small
\textbf{Multi-faceted}\\[4pt]
\begin{tabular}{@{}llc@{}}
\toprule
\textbf{Predicted} & \textbf{Matched ref.} & \textbf{R-1} \\
\midrule
2018 ``idgaf'' & 2018 ``idgaf'' & 1.00 \cmark \\
2024 ``houdini'' & 2024 ``houdini'' & 1.00 \cmark \\
\bottomrule
\end{tabular}\\[4pt]
$P = 2/2,\; R = 2/4$\\
\textbf{\rougef{} = 0.67}
\end{minipage}

\vspace{6pt}
For the single-faceted case, exact entity name matches yield ROUGE-1 = 1.0, and ``wonderlus'' has 0.0 overlap with all reference elements. For the multi-faceted case, both predicted elements match their corresponding reference elements perfectly after flattening. Precision is 1.0 (both predictions are correct) but recall is only 0.5 (2 of 4 reference elements are matched), yielding F1 = 0.67.

\subsection{\judgef{}}

The \judgef{} follows the same Jonker-Volgenant assignment process as \rougef{} but with two differences: (1) the ROUGE filter threshold is set to 0.0 instead of 0.5, allowing even low-overlap pairs to be candidates for matching, and (2) each matched pair is sent to an LLM grader that returns a CORRECT or INCORRECT verdict based on semantic equivalence (see App.~\ref{subsec:appendix_judge_llm_prompt} for the full judge prompt). Predicted elements that remain unmatched after assignment are automatically graded INCORRECT. Precision and recall are computed over elements judged CORRECT.

\begin{minipage}[t]{0.48\textwidth}
\centering
\small
\textbf{Single-faceted}\\[4pt]
\begin{tabular}{@{}llc@{}}
\toprule
\textbf{Predicted} & \textbf{Matched ref.} & \textbf{Verdict} \\
\midrule
wonderlus & \textit{unmatched} & \xmark \\
gaia & gaia & \cmark \\
poppie nongena & poppie nongena & \cmark \\
vaselinetjie & vaselinetjie & \cmark \\
\bottomrule
\end{tabular}\\[4pt]
$P = 3/4,\; R = 3/4$\\
\textbf{\judgef{} = 0.75}
\end{minipage}%
\hfill\vrule\hfill
\begin{minipage}[t]{0.48\textwidth}
\centering
\small
\textbf{Multi-faceted}\\[4pt]
\begin{tabular}{@{}llc@{}}
\toprule
\textbf{Predicted} & \textbf{Matched ref.} & \textbf{Verdict} \\
\midrule
2018 / ``IDGAF'' & 2018 / ``IDGAF'' & \cmark \\
2024 / ``Houdini'' & 2024 / ``Houdini'' & \cmark \\
\bottomrule
\end{tabular}\\[4pt]
$P = 2/2,\; R = 2/4$\\
\textbf{\judgef{} = 0.67}
\end{minipage}

\vspace{6pt}
In these examples, the \judgef{} agrees with \rougef{} because the matches are either exact or clearly unrelated. The metrics diverge in cases with heavy paraphrasing (where the judge recognizes semantic equivalence that ROUGE misses, e.g., ``championship game'' vs.\ ``title game'') or when token overlap is high but facts are wrong (where ROUGE gives credit but the judge does not, e.g., a predicted year of 2002 vs.\ a gold year of 2000 would share most surrounding tokens but be factually incorrect).

\subsection{\orderedf{}}

For the 14.6\% of questions requiring a specific sequence, we additionally compute an ordered variant of \judgef{} that penalizes correct elements in the wrong position. Consider the following ordered question:

\smallskip
\noindent\textit{Question:} Who were the first 4 winners of the Big Apple Award for Game of the Year?

\smallskip
\noindent\fbox{\begin{minipage}[t]{0.46\textwidth}
\small
\textbf{Reference (4 elements):}\\[2pt]
\texttt{1. The Elder Scrolls V: Skyrim}\\
\texttt{2. The Walking Dead}\\
\texttt{3. The Last of Us}\\
\texttt{4. Wolfenstein: The New Order}
\end{minipage}%
\hfill\vrule\hspace{8pt}
\begin{minipage}[t]{0.46\textwidth}
\small
\textbf{Model prediction (4 elements):}\\[2pt]
\texttt{1. The Walking Dead}\\
\texttt{2. The Elder Scrolls V: Skyrim}\\
\texttt{3. The Last of Us}\\
\texttt{4. Wolfenstein: The New Order}
\end{minipage}}

\smallskip
The model retrieves all four correct elements, but swaps the first two. Under the standard (order-agnostic) \judgef{}, all four are matched as CORRECT, yielding $P = R = 1.0$ and \textbf{\judgef{} = 1.0}. Under the ordered variant, elements are matched positionally: position 1 (The Walking Dead $\neq$ The Elder Scrolls V: Skyrim) and position 2 (The Elder Scrolls V: Skyrim $\neq$ The Walking Dead) are marked INCORRECT, while positions 3 and 4 match. This yields $P = R = 0.5$ and \textbf{\orderedf{} = 0.5}. The gap between the two scores isolates ordering as a source of error distinct from factual retrieval.

\clearpage
\section{Instruction Prompt Templates}
\label{sec:appendix_prompt}

This section describes the prompt template used as part of different evaluation settings, as well as the LLM Judge prompt template.

\subsection{Instruction Prompt Template - Zero shot}
\label{subsec:appendix_prompt_zs}

The following prompt template is used during evaluation in the zero-shot setting:

\begin{quote}
\small
\texttt{Answer the question below by producing a list-formatted output, lists can be formatted one of two ways:\\\\
For simple questions with few constraints requesting only main elements and not requesting sub-elements, use the following format:}\\\\ \texttt{1. List Element 1 \\ 2. List Element 2}\\\\ \texttt{For complex questions asking for multiple factoids in addition to the primary question, use the following format:}\\\\ \texttt{1. List Element 1 \\ $\ast$ Category 1: Sub-Element 1 \\ $\ast$ Category 2: Sub-Element 2}\\\\ \texttt{2. List Element 2 \\ $\ast$ Category 1: Sub-Element 1 \\ $\ast$ Category 2: Sub-Element 2}\\\\ \texttt{All answers are bounded by 3 to 10 main elements, do not generate more than 10 main elements in the list. Do not add any information that is not asked for in the question.}\\

\texttt{<QUESTION>}
\end{quote}

\subsection{Instruction Prompt Template - Zero shot (Order Aware)}
\label{subsec:appendix_prompt_zs_order}

The following template is used during order-aware evaluation (App.~\ref{sec:appendix_ordering}) in the zero-shot setting:

\begin{quote}
\small
\texttt{Answer the question below by producing a list-formatted output, lists can be formatted one of two ways:\\\\
For simple questions with few constraints requesting only main elements and not requesting sub-elements, use the following format:}\\\\ \texttt{1. List Element 1 \\ 2. List Element 2}\\\\ \texttt{For complex questions asking for multiple factoids in addition to the primary question, use the following format:}\\\\ \texttt{1. List Element 1 \\ $\ast$ Category 1: Sub-Element 1 \\ $\ast$ Category 2: Sub-Element 2}\\\\ \texttt{2. List Element 2 \\ $\ast$ Category 1: Sub-Element 1 \\ $\ast$ Category 2: Sub-Element 2}\\\\ \texttt{All answers are bounded by 3 to 10 main elements, do not generate more than 10 main elements in the list. When the question specifies or implies a natural ordering (e.g. chronological, ranked, or sequential), list elements must follow that order for correctness of answer. Do not add any information that is not asked for in the question.}\\

\texttt{<QUESTION>}
\end{quote}

\subsection{Instruction Prompt Template - Oracle RAG}
\label{subsec:appendix_prompt_oracle_rag}

The following template is used during evaluation in the oracle RAG setting:

\begin{quote}
\small
\texttt{You are given a context passage and a question. Using only the provided context, answer the question by producing a list-formatted output. Lists can be formatted one of two ways:\\\\
For simple questions with few constraints requesting only main elements and not requesting sub-elements, use the following format:}\\\\ \texttt{1. List Element 1 \\ 2. List Element 2}\\\\ \texttt{For complex questions asking for multiple factoids in addition to the primary question, use the following format:}\\\\ \texttt{1. List Element 1 \\ $\ast$ Category 1: Sub-Element 1 \\ $\ast$ Category 2: Sub-Element 2}\\\\ \texttt{2. List Element 2 \\ $\ast$ Category 1: Sub-Element 1 \\ $\ast$ Category 2: Sub-Element 2}\\\\ \texttt{All answers are bounded by 3 to 10 main elements, do not generate more than 10 main elements in the list. Do not add any information that is not asked for in the question.}\\

\texttt{Context:}\\
\texttt{<CONTEXT>}

\texttt{Question: <QUESTION>}
\end{quote}

\subsection{Instruction Prompt Template - Few Shot}
\label{subsec:appendix_few_shot_prompt}

The following template is used during evaluation in the few-shot setting. By default, we provide 5 examples per prompt. For few-shot evaluation, examples are drawn from the corresponding question type: single-faceted examples for \simplelistqa{}, multi-faceted for \complexlistqa{}, and a combination for the full dataset.

\begin{quote}
\small
\texttt{Answer the question below by producing a list-formatted output, lists can be formatted one of two ways:\\\\
For simple questions with few constraints requesting only main elements and not requesting sub-elements, use the following format:}\\\\ \texttt{1. List Element 1 \\ 2. List Element 2}\\\\ \texttt{For complex questions asking for multiple factoids in addition to the primary question, use the following format:}\\\\ \texttt{1. List Element 1 \\ $\ast$ Category 1: Sub-Element 1 \\ $\ast$ Category 2: Sub-Element 2}\\\\ \texttt{2. List Element 2 \\ $\ast$ Category 1: Sub-Element 1 \\ $\ast$ Category 2: Sub-Element 2}\\\\ \texttt{All answers are bounded by 3 to 10 main elements, do not generate more than 10 main elements in the list. Do not add any information that is not asked for in the question.}\\\\ \texttt{Here are some examples:}\\ 

\texttt{Question: <EXAMPLE\_1\_QUESTION>}\\
\texttt{Answer: <EXAMPLE\_1\_ANSWER>}\\

\texttt{Question: <EXAMPLE\_2\_QUESTION>}\\
\texttt{Answer: <EXAMPLE\_2\_ANSWER>}\\

\texttt{Question: <EXAMPLE\_3\_QUESTION>}\\
\texttt{Answer: <EXAMPLE\_3\_ANSWER>}\\

\texttt{Question: <EXAMPLE\_4\_QUESTION>}\\
\texttt{Answer: <EXAMPLE\_4\_ANSWER>}\\

\texttt{Question: <EXAMPLE\_5\_QUESTION>}\\
\texttt{Answer: <EXAMPLE\_5\_ANSWER>}\\

\texttt{Question: \textbf{<QUESTION>}}\\
\texttt{Answer:}\\

\end{quote}

\subsection{Judge LLM Prompt Template}
\label{subsec:appendix_judge_llm_prompt}

Our judge prompt builds on the grading template from SimpleQA~\citep{wei2024simpleqa}, a well-established factual knowledge benchmark, and extends it to handle the list-structured, multi-element responses required by \listqa{}. The complete prompt template is as follows:

\begin{quote}
\small
\texttt{Your job is to look at a question, a gold target, and a predicted answer, and then assign a grade of either ["CORRECT", "INCORRECT"]. The answer to the question is always going to be a list, however we will only be judging each list element one at a time, so the gold target and predicted answer may not completely answer the question and that is ok.}\\\\
\texttt{First, I will give examples of each grade, and then you will grade a new example.}\\

\texttt{The following are examples of CORRECT predicted answers.}\\

\texttt{Question: What are the names of Barack Obama's children?}\\
\texttt{Gold target: malia obama}\\
\texttt{Predicted answer 1: malia}\\
\texttt{Predicted answer 2: Malia Obama}\\
\texttt{$\ast$ birthday: July 4, 1998}\\
\texttt{Predicted answer 3: malia obama (born on July 4, 1998).}\\

\texttt{Question: What NFL teams has Tom Brady played for and what years was he active at these teams?}\\
\texttt{Gold target: New England Patriots}\\
\texttt{* Start Year: 2000}\\
\texttt{* End Year: 2019}\\
\texttt{Predicted answer 1: New England Patriots (2000 - 2019)}\\
\texttt{Predicted answer 2: patriots}\\
\texttt{* duration: 2000 - 2019}\\
\texttt{Predicted answer 3: NE Patriots}\\
\texttt{* Start: 2000}\\
\texttt{* End: 2019}\\

\texttt{These predicted answers are all CORRECT because:}\\
\texttt{- They fully contain the important information in the gold target.}\\
\texttt{- They do not contain any information that contradicts the gold target.}\\
\texttt{- Only semantic meaning matters; capitalization, punctuation, grammar, and order don't matter.}\\
\texttt{- Hedging and guessing are permissible, provided that the gold target is fully included and the response contains no incorrect information or contradictions.}\\

\texttt{The following are examples of INCORRECT predicted answers.}\\

\texttt{Question: What are the names of Barack Obama's children?}\\
\texttt{Gold target: Malia}\\
\texttt{Predicted answer 1: Susan}\\
\texttt{Predicted answer 2: Malia, Sasha, and Susan.}\\
\texttt{Predicted answer 3: Two children.}\\
\texttt{Predicted answer 4: Obama Family Tree}\\
\texttt{* Father: Barrack}\\
\texttt{* Mother: Michelle}\\
\texttt{* Child 1: Malia}\\
\texttt{* Child 2: Sasha}\\

\texttt{Question: What NFL teams has Tom Brady play for and what years was he active at these teams?}\\
\texttt{Gold target: New England Patriots}\\
\texttt{* Start Year: 2000}\\
\texttt{* End Year: 2019}\\
\texttt{Predicted answer 1: New England Patriots}\\
\texttt{Predicted answer 2: New England Patriots}\\
\texttt{* Start Year: 2002}\\
\texttt{* End Year: 2019}\\
\texttt{Predicted answer 3: Jacksonville Jaguars}\\
\texttt{Predicted answer 4: Jacksonville Jaguars}\\
\texttt{* Duration: 2000 - 2019}\\
\texttt{Predicted answer 5: New England Patriots}\\
\texttt{* Duration: 2000 - 2019}\\
\texttt{* Superbowl Wins: 7}\\
\texttt{* Touchdowns: 450}\\
\texttt{* Interceptions: 6}\\

\texttt{These predicted answers are all INCORRECT because:}\\
\texttt{- A factual statement in the answer contradicts the gold target or does not provide all the requested information. Incorrect statements also provide information that cannot be validated against the gold target are also considered incorrect.}\\

\texttt{Also note the following things:}\\
\begin{itemize}
    \item \texttt{For grading questions where the gold target is a number, the predicted answer needs to be correct to the last significant figure in the gold answer. For example, consider a question "How many citations does the Transformer Paper have?" with gold target "120k".}
    \begin{itemize}
        \item \texttt{Predicted answers "120k", "124k", and "115k" are all CORRECT.}
        \item \texttt{Predicted answers "100k" and "113k" are INCORRECT.}
        \item \texttt{Predicted answers "around 100k" and "more than 50k" are considered INCORRECT because they neither confirm nor contradict the gold target.}
    \end{itemize}
\end{itemize}

\begin{itemize}
    \item \texttt{The gold target may contain more information than the question. In such cases, the predicted answer only needs to contain the information that is in the question.}
    \begin{itemize}
        \item \texttt{For example, consider the question "What episode did Derek and Meredith get legally married in Grey's Anatomy?" with gold target "Season 7, Episode 20: White Wedding". Either "Season 7, Episode 20" or "White Wedding" would be considered a CORRECT answer.}
    \end{itemize}
\end{itemize}

\begin{itemize}
    \item \texttt{Do not punish predicted answers if they omit information that would be clearly inferred from the question.}
    \begin{itemize}
        \item \texttt{For example, consider the question "What city is OpenAI headquartered in?" and the gold target "San Francisco, California". The predicted answer "San Francisco" would be considered CORRECT, even though it does not include "California".}
        \item \texttt{Consider the question "What award did A pretrainer's guide to training data: Measuring the effects of data age, domain coverage, quality, \& toxicity win at NAACL '24?", the gold target is "Outstanding Paper Award". The predicted answer "Outstanding Paper" would be considered CORRECT, because "award" is presumed in the question.}
        \item \texttt{For the question "What is the height of Jason Wei in meters?", the gold target is "1.73 m". The predicted answer "1.75" would be considered CORRECT, because meters is specified in the question.}
        \item \texttt{For the question "What is the name of Barack Obama's wife?", the gold target is "Michelle Obama". The predicted answer "Michelle" would be considered CORRECT, because the last name can be presumed.}
    \end{itemize}
\end{itemize}

\begin{itemize}
    \item \texttt{Do not punish for typos in people's name if it's clearly the same name.}
    \begin{itemize}
        \item \texttt{For example, if the gold target is "Hyung Won Chung", you can consider the following predicted answers as correct: "Hyoong Won Choong", "Hyungwon Chung", or "Hyun Won Chung".}
    \end{itemize}
\end{itemize}

\texttt{Here is a new example. Simply reply with either CORRECT, INCORRECT. Don't apologize or correct yourself if there was a mistake; we are just trying to grade the answer.}\\

\texttt{Question: <QUESTION>}\\
\texttt{Gold target: <GOLD TARGET LIST ITEM>}\\
\texttt{Predicted answer: <PREDICTED LIST ITEM>}\\

\texttt{Grade the predicted answer of this new question as one of:}\\
\texttt{A: CORRECT}\\
\texttt{B: INCORRECT}\\

\texttt{Just return the letters "A", or "B", with no text around it.}
\end{quote}

\clearpage
\section{Full Results}
\label{sec:appendix_full_results}

Borrowing the notion of task demands from \citet{hu2024auxiliary}, we vary demand at scoring time rather than elicitation: \relaxedrecall{}/\relaxedprecision{} ignore output structure (reduced demand), while \judgef{} requires it (greater demand). Tab.~\ref{tab:full_results} reports all five metrics across all 42 evaluated configurations (31 standard + 11 thinking-enabled) across 8 model families, sorted by overall \textit{LLM-Judge-F1}. For each metric, we report scores on the full dataset (All), \simplelistqa{} (S), and \complexlistqa{} (C).

\begin{table*}[ht]
\caption{Full results across all five metrics. All = Overall, S = \simplelistqa{}, C = \complexlistqa{}. [$^\dagger$ = thinking-enabled inference], [$^*$ = Preview Models]. All values are percentages. Models sorted by overall \textit{LLM-Judge-F1}. \textbf{Recall:} \relaxedrecall{}; \textbf{Precision:} \relaxedprecision{}; \textbf{Format:} \formatmetric{}}
\label{tab:full_results}
\centering
\small
\setlength{\tabcolsep}{2pt}
\begin{tabular}{@{}l|ccc|ccc|ccc|ccc|ccc@{}}
\toprule
& \multicolumn{3}{c}{\textbf{\judgef{}} $\uparrow$} & \multicolumn{3}{c}{\textbf{Recall} $\uparrow$} & \multicolumn{3}{c}{\textbf{Precision} $\uparrow$} & \multicolumn{3}{c}{\textbf{Format} $\uparrow$} & \multicolumn{3}{c}{\textbf{\rougef{}} $\uparrow$} \\
\cmidrule(lr){2-4} \cmidrule(lr){5-7} \cmidrule(lr){8-10} \cmidrule(lr){11-13} \cmidrule(lr){14-16}
\textbf{Model} & \textbf{All} & \textbf{S} & \textbf{C} & \textbf{All} & \textbf{S} & \textbf{C} & \textbf{All} & \textbf{S} & \textbf{C} & \textbf{All} & \textbf{S} & \textbf{C} & \textbf{All} & \textbf{S} & \textbf{C} \\
\midrule
Gemini 3.1 Pro$^*$$^\dagger$ & 60.7 & 67.7 & 50.3 & 59.6 & 59.9 & 59.4 & 63.7 & 66.5 & 61.8 & 97.7 & 98.5 & 97.2 & 68.5 & 68.8 & 68.3 \\ 
Gemini 3 Flash$^*$$^\dagger$ & 56.4 & 63.7 & 45.8 & 59.7 & 59.4 & 60.2 & 57.8 & 54.9 & 62.1 & 91.9 & 87.3 & 98.8 & 63.7 & 62.6 & 65.5 \\
Gemini 3.1 Flash Lite$^*$$^\dagger$ & 40.1 & 48.4 & 27.9 & 44.8 & 44.1 & 45.7 & 38.1 & 28.5 & 52.1 & 39.9 & 3.8 & 92.7 & 38.8 & 33.2 & 47.0 \\
Gemini 2.5 Flash$^\dagger$ & 38.3 & 47.1 & 25.4 & 43.0 & 42.2 & 44.2 & 52.3 & 45.9 & 61.8 & 70.3 & 86.8 & 46.2 & 43.3 & 47.8 & 36.8 \\
Claude Opus 4.6$^\dagger$ & 37.9 & 46.2 & 25.7 & 43.2 & 42.3 & 44.5 & 50.8 & 45.9 & 58.0 & 66.8 & 69.0 & 63.7 & 38.6 & 40.6 & 35.6 \\
GPT-5.4 & 36.5 & 44.5 & 24.9 & 42.2 & 40.9 & 44.1 & 44.2 & 40.3 & 49.9 & 87.2 & 85.0 & 90.6 & 44.6 & 45.5 & 43.3 \\
GPT-5.4-mini$^\dagger$ & 36.2 & 45.0 & 23.4 & 39.9 & 40.5 & 39.6 & 52.1 & 46.2 & 60.7 & 60.3 & 76.4 & 36.7 & 39.5 & 43.9 & 33.0 \\
Claude Opus 4.7 & 35.0 & 42.1 & 24.6 & 39.5 & 38.2 & 41.3 & 42.1 & 38.9 & 46.9 & 75.0 & 75.4 & 74.5 & 40.0 & 39.6 & 40.5 \\
Claude Opus 4.6 & 34.6 & 43.0 & 22.4 & 40.2 & 38.9 & 42.1 & 45.6 & 41.9 & 50.9 & 78.3 & 73.9 & 84.8 & 40.5 & 41.5 & 39.0 \\
Claude Opus 4.5 & 33.9 & 42.0 & 22.2 & 38.9 & 37.6 & 40.9 & 41.7 & 37.5 & 47.8 & 70.8 & 63.0 & 82.1 & 38.7 & 38.8 & 38.5 \\
Gemini 2.5 Flash & 33.7 & 41.9 & 21.6 & 37.8 & 36.7 & 39.6 & 43.7 & 40.1 & 49.1 & 76.4 & 72.5 & 82.2 & 42.3 & 43.3 & 40.8 \\
Nova Premier & 31.6 & 39.0 & 20.6 & 39.5 & 38.0 & 41.6 & 41.6 & 33.6 & 53.4 & 60.7 & 60.6 & 61.0 & 32.0 & 31.8 & 32.3 \\
DeepSeek V3.1 & 29.3 & 36.7 & 18.3 & 36.0 & 34.1 & 38.8 & 40.8 & 34.5 & 50.0 & 73.3 & 76.0 & 69.4 & 33.4 & 35.1 & 31.0 \\
Claude Sonnet 4.6 & 28.0 & 35.5 & 16.9 & 34.2 & 32.5 & 36.7 & 36.7 & 30.8 & 45.5 & 60.2 & 45.3 & 82.2 & 31.5 & 31.9 & 31.0 \\
GPT-5.4-mini & 27.0 & 35.2 & 15.1 & 33.3 & 31.5 & 36.0 & 44.2 & 35.5 & 56.9 & 64.1 & 84.8 & 33.6 & 32.5 & 37.3 & 25.4 \\
Mistral Large 3 675B-A41B & 25.7 & 33.2 & 14.8 & 37.7 & 36.1 & 40.0 & 44.5 & 33.7 & 60.5 & 58.9 & 77.9 & 31.0 & 25.0 & 28.6 & 19.7 \\
Claude Sonnet 4.5 & 25.3 & 30.2 & 18.1 & 30.9 & 27.9 & 35.2 & 31.6 & 25.7 & 40.2 & 49.2 & 41.1 & 61.0 & 27.9 & 26.1 & 30.5 \\
Qwen 3.5 397B-A17B & 24.8 & 31.6 & 14.8 & 32.6 & 30.4 & 35.8 & 33.9 & 29.4 & 40.5 & 73.0 & 69.0 & 78.8 & 29.9 & 31.2 & 28.0 \\
Qwen 3.5 122B-A10B$^\dagger$ & 23.5 & 30.5 & 13.1 & 29.5 & 27.0 & 33.2 & 34.5 & 29.3 & 42.1 & 95.1 & 95.0 & 95.2 & 30.8 & 32.5 & 28.2 \\
Nova Pro & 22.0 & 29.3 & 11.4 & 30.6 & 28.5 & 33.8 & 31.8 & 24.9 & 42.1 & 65.6 & 53.8 & 83.0 & 25.5 & 25.3 & 25.8 \\
Gemini 2.5 Flash Lite & 20.4 & 27.2 & 10.4 & 28.2 & 25.6 & 32.0 & 37.8 & 28.4 & 51.5 & 63.9 & 83.6 & 35.1 & 25.9 & 30.0 & 20.0 \\
DeepSeek V3.2 & 20.3 & 26.5 & 11.1 & 31.3 & 28.7 & 35.0 & 41.0 & 30.7 & 56.1 & 54.8 & 71.9 & 29.8 & 18.9 & 21.9 & 14.7 \\
Qwen 3.5 122B-A10B & 18.9 & 25.3 & 9.7 & 28.0 & 25.5 & 31.7 & 30.4 & 24.4 & 39.2 & 82.2 & 84.1 & 79.5 & 25.7 & 27.8 & 22.6 \\
Gemma 4 31B$^\dagger$ & 18.7 & 25.3 & 9.0 & 23.3 & 20.9 & 26.8 & 31.2 & 27.3 & 37.0 & 74.9 & 73.9 & 76.5 & 27.4 & 29.1 & 24.8 \\
Qwen 3.5 27B$^\dagger$ & 18.3 & 24.6 & 9.1 & 24.9 & 22.0 & 29.0 & 30.0 & 24.8 & 37.7 & 95.3 & 95.9 & 94.6 & 26.5 & 28.2 & 24.0 \\

Qwen 3.5 35B-A3B$^\dagger$ & 18.2 & 24.4 & 9.0 & 24.6 & 21.5 & 29.1 & 29.8 & 24.5 & 37.5 & 90.1 & 89.6 & 91.0 & 25.8 & 27.7 & 22.9 \\
Qwen 3-Next 80B-A3B & 17.4 & 23.6 & 8.3 & 24.7 & 21.8 & 28.9 & 25.3 & 17.1 & 37.4 & 39.1 & 7.8 & 85.1 & 14.7 & 11.5 & 19.4 \\
Gemma 4 31B & 16.0 & 22.0 & 7.3 & 20.6 & 17.9 & 24.7 & 31.3 & 26.5 & 38.4 & 54.3 & 45.8 & 66.7 & 22.4 & 22.8 & 21.8 \\
Magistral Small 24B & 15.8 & 21.8 & 7.0 & 24.3 & 20.9 & 29.2 & 27.6 & 19.4 & 39.5 & 59.6 & 48.0 & 76.8 & 18.6 & 19.0 & 18.0 \\
Qwen 3.5 27B & 14.8 & 20.1 & 7.1 & 23.6 & 20.5 & 28.2 & 27.6 & 20.5 & 38.1 & 71.9 & 71.6 & 72.4 & 20.1 & 21.2 & 18.4 \\
GPT-5.4-nano & 14.5 & 20.4 & 5.7 & 22.1 & 19.2 & 26.3 & 31.5 & 20.8 & 47.1 & 58.1 & 77.7 & 29.2 & 16.8 & 20.2 & 11.9 \\
Gemma 4 26B-A4B & 13.8 & 19.0 & 6.0 & 19.2 & 16.1 & 23.7 & 28.7 & 23.4 & 36.6 & 52.5 & 44.2 & 64.6 & 19.6 & 19.7 & 19.4 \\
Qwen 3.5 35B-A3B & 13.4 & 18.5 & 5.9 & 22.4 & 19.0 & 27.4 & 25.7 & 18.8 & 35.9 & 51.7 & 37.0 & 73.2 & 13.9 & 13.7 & 14.2 \\
Nova Micro & 12.0 & 17.5 & 4.0 & 19.3 & 16.1 & 24.0 & 30.3 & 18.3 & 47.9 & 56.8 & 79.1 & 24.1 & 17.4 & 20.6 & 12.7 \\
Qwen 3.5 9B$^\dagger$ & 11.4 & 16.1 & 4.4 & 17.1 & 14.0 & 21.6 & 23.1 & 17.9 & 30.8 & 87.2 & 87.0 & 87.5 & 16.5 & 19.0 & 12.8 \\
Claude Haiku 4.5 & 11.1 & 14.1 & 6.7 & 16.4 & 13.7 & 20.3 & 16.7 & 13.4 & 21.5 & 49.1 & 58.6 & 35.2 & 13.7 & 13.5 & 14.1 \\
Qwen 3.5 9B & 8.3 & 12.0 & 2.9 & 17.9 & 13.9 & 23.8 & 20.9 & 13.1 & 32.4 & 62.0 & 65.5 & 56.8 & 10.8 & 12.2 & 8.8 \\
Ministral 3 14B & 7.2 & 10.4 & 2.6 & 18.8 & 15.2 & 24.1 & 34.0 & 19.9 & 54.8 & 46.9 & 76.2 & 3.8 & 3.9 & 6.0 & 0.9 \\
Gemma 4 E4B & 6.2 & 9.1 & 1.9 & 10.2 & 7.6 & 13.9 & 17.4 & 11.7 & 25.7 & 33.6 & 44.9 & 17.0 & 9.2 & 11.0 & 6.6 \\
Qwen 3.5 4B & 5.1 & 7.8 & 1.3 & 11.7 & 8.2 & 16.9 & 18.3 & 10.2 & 30.2 & 39.2 & 43.9 & 32.3 & 6.5 & 8.2 & 3.9 \\
Ministral 3 8B & 4.8 & 6.9 & 1.7 & 16.1 & 12.1 & 22.0 & 31.2 & 17.5 & 51.3 & 45.7 & 71.0 & 8.6 & 2.4 & 3.5 & 0.9 \\
Ministral 3 3B & 3.3 & 5.1 & 0.5 & 11.8 & 7.9 & 17.5 & 26.3 & 12.9 & 46.0 & 46.2 & 76.3 & 2.1 & 2.5 & 3.9 & 0.5 \\
\bottomrule
\end{tabular}
\end{table*}

\clearpage
\section{Model Evaluation Infrastructure}
\label{sec:appendix_infrastructure}

All evaluations were orchestrated using \texttt{inspect-ai}.\footnote{\url{https://inspect.ai-safety-institute.org.uk/}} Models accessed via API (OpenAI\footnote{\url{https://platform.openai.com/}}, Google Gemini\footnote{\url{https://ai.google.dev/}}, Amazon Bedrock\footnote{\url{https://aws.amazon.com/bedrock/}}) were evaluated using the default inference configurations provided by each API for the respective models. Self-hosted models were served using vLLM\footnote{\url{https://docs.vllm.ai/}} with an OpenAI-compatible API endpoint, using official HuggingFace checkpoints and the default \texttt{generation\_config.json} parameters provided with each checkpoint.

For thinking-enabled evaluations, we configured \texttt{-{}-reasoning-effort} for Gemini models and \texttt{-{}-reasoning-tokens} (budget) for Claude models via \texttt{inspect-ai}'s generation parameters. Tab.~\ref{tab:infrastructure} details the infrastructure used for each model family.

\begin{table}[ht]
\caption{Inference infrastructure for each evaluated model.}
\vspace{4pt}
\label{tab:infrastructure}
\centering
\small
\setlength{\tabcolsep}{4pt}
\begin{tabular}{@{}llll@{}}
\toprule
\textbf{Model Family} & \textbf{Models} & \textbf{Infrastructure} \\
\midrule
GPT & \emph{GPT-5.4}\\
    & \emph{GPT-5.4 mini} & OpenAI API\\
    & \emph{GPT-5.4 nano}\\
\midrule
Claude & Claude \emph{Opus 4.7, 4.6, 4.5} \\
       & Claude \emph{Sonnet 4.6, 4.5} & Amazon Bedrock\\
       & Claude \emph{Haiku 4.5} \\
\midrule
Gemini & \emph{Gemini 3.1 Pro} & \\
        & \emph{Gemini 3.1 Flash Lite} & \\
        & \emph{Gemini 3 Flash} & Google (Gemini) API \\
        & \emph{Gemini 2.5 Flash} & \\ 
       & \emph{Gemini 2.5 Flash Lite} & \\

\midrule
Nova & \emph{Nova Premier} & \\
     & \emph{Nova Pro} &  Amazon Bedrock \\
     & \emph{Nova Micro} & \\
     
\midrule
DeepSeek & \emph{DeepSeek V3.2}\\
         & \emph{DeepSeek V3.1} & Amazon Bedrock \\
\midrule
Qwen 3.5 & \emph{Qwen 3.5 397B-A17B (FP8)} &\\
     & \emph{Qwen 3.5 122B-A10B} &\\
     & \emph{Qwen 3.5 35B-A3B} & Self-hosted (vLLM)\\ 
     & \emph{Qwen 3.5 27B} &  \\
     & \emph{Qwen 3.5 9B} & \\
     & \emph{Qwen 3.5 4B} & \\
\midrule
Qwen 3 & \emph{Qwen 3 Next 80B-A3B} & Amazon Bedrock \\
\midrule
Mistral & \emph{Mistral Large 3 675B-A41B} &\\
        & \emph{Magistral Small 24B} &  \\
        & \emph{Ministral 3 14B} & Amazon Bedrock \\
        & \emph{Ministral 3 8B} & \\
        & \emph{Ministral 3 3B} & \\
\midrule
Gemma &  \emph{Gemma 4 31B} &  \\
      &  \emph{Gemma 4 26B-A4B} & Self-hosted (vLLM) \\
      &  \emph{Gemma 4 E4B} &  \\
\bottomrule
\end{tabular}
\end{table}

\clearpage
\section{Format Error Analysis}
\label{sec:appendix_format_errors}

We categorize format errors into four types to understand \emph{how} models fail to follow list-formatting instructions:

\begin{itemize}[leftmargin=*,itemsep=2pt]
    \item \textbf{Wrong facet type}: The model produces a list but uses the wrong structure; e.g., generating sub-bullets for a single-faceted question or a flat list for a multi-faceted question. This is the most common error on average, indicating that many models find it difficult to infer the appropriate list style from the question.
    \item \textbf{Too few elements} ($<$3): The model produces a list with only one or two elements, falling below the minimum of 3 required by the task.
    \item \textbf{No list output}: The model generates a free-text response without any list structure.
    \item \textbf{Mixed facet type}: The model produces a list where some elements use sub-bullets and others do not, mixing single-faceted and multi-faceted formatting within a single response.
\end{itemize}

The dominant failure mode on average is \emph{wrong facet type}, followed by \emph{too few elements}. \emph{Mixed facet type} is rare, and thinking nearly eliminates it. Tab.~\ref{tab:format_errors_thinking} shows how these error types shift when thinking is enabled.

\paragraph{Effect of thinking on format errors.}

\begin{table}[ht]
\caption{Format error distribution (\% of 9,045 samples) under standard (Std) vs.\ thinking-enabled (Think) inference for models from Fig.~\ref{fig:thinking_results}(b). \textbf{Perfect} indicates accurate format compliance.}
\vspace{4pt}
\label{tab:format_errors_thinking}
\centering
\small
\setlength{\tabcolsep}{3pt}
\begin{tabular}{@{}llccccc@{}}
\toprule
\textbf{Model} & \textbf{Setting} & \textbf{Perfect} & \textbf{Wrong Facet} & \textbf{Too Few Elem.} & \textbf{No List} & \textbf{Mixed Facet} \\
\midrule
\multirow{2}{*}{GPT-5.4 mini}
& Std   & 63.9 & 21.8 & 14.0 & 0.0 & 0.2 \\
& Think & 60.1 \textcolor{red}{$\downarrow$} & 17.8 \textcolor{teal}{$\downarrow$} & 21.7 \textcolor{red}{$\uparrow$} & 0.0 & 0.3 \textcolor{red}{$\uparrow$} \\
\midrule
\multirow{2}{*}{Gemini 2.5 Flash}
& Std   & 76.2 & 9.3 & 13.4 & 0.6 & 0.5 \\
& Think & 70.2 \textcolor{red}{$\downarrow$} & 18.5 \textcolor{red}{$\uparrow$} & 9.8 \textcolor{teal}{$\downarrow$} & 1.2 \textcolor{red}{$\uparrow$} & 0.2 \textcolor{teal}{$\downarrow$}\\
\midrule 
\multirow{2}{*}{Claude Opus 4.6}
& Std   & 77.4 & 7.5 & 12.1 & 1.5 & 1.5 \\
& Think & 66.7 \textcolor{red}{$\downarrow$} & 16.2 \textcolor{red}{$\uparrow$} & 12.4 \textcolor{red}{$\uparrow$} & 4.4 \textcolor{red}{$\uparrow$} & 0.2 \textcolor{teal}{$\downarrow$}\\
\midrule
\multirow{2}{*}{Qwen 3.5 122B-A10B}
& Std   & 81.2 & 5.9 & 6.6 & 4.6 & 1.6 \\
& Think & 95.1 \textcolor{teal}{$\uparrow$} & 4.2 \textcolor{teal}{$\downarrow$} & 0.5 \textcolor{teal}{$\downarrow$} & 0.1 \textcolor{teal}{$\downarrow$} & 0.1 \textcolor{teal}{$\downarrow$} \\
\midrule
\multirow{2}{*}{Gemma 4 31B}
& Std   & 53.9 & 14.4 & 29.0 & 2.0 & 0.7 \\
& Think & 74.9 \textcolor{teal}{$\uparrow$} & 3.7 \textcolor{teal}{$\downarrow$} & 11.8 \textcolor{teal}{$\downarrow$} & 9.5 \textcolor{red}{$\uparrow$} & 0.0 \textcolor{teal}{$\downarrow$} \\
\midrule
\multirow{2}{*}{Qwen 3.5 35B-A3B}
& Std   & 49.8 & 26.0 & 20.2 & 0.6 & 3.5 \\
& Think & 90.0 \textcolor{teal}{$\uparrow$} & 9.0 \textcolor{teal}{$\downarrow$} & 0.7 \textcolor{teal}{$\downarrow$} & 0.0 \textcolor{teal}{$\downarrow$} & 0.3 \textcolor{teal}{$\downarrow$} \\
\midrule
\multirow{2}{*}{Qwen 3.5 27B}
& Std   & 70.3 & 14.1 & 11.7 & 1.3 & 2.6 \\
& Think & 95.3 \textcolor{teal}{$\uparrow$} & 4.5 \textcolor{teal}{$\downarrow$} & 0.1 \textcolor{teal}{$\downarrow$} & 0.0 \textcolor{teal}{$\downarrow$} & 0.0 \textcolor{teal}{$\downarrow$} \\
\midrule
\multirow{2}{*}{Qwen 3.5 9B}
& Std   & 55.4 & 19.8 & 11.0 & 2.8 & 11.0 \\
& Think & 85.8 \textcolor{teal}{$\uparrow$} & 10.7 \textcolor{teal}{$\downarrow$} & 0.3 \textcolor{teal}{$\downarrow$} & 0.8 \textcolor{teal}{$\downarrow$} & 2.4 \textcolor{teal}{$\downarrow$} \\
\bottomrule
\end{tabular}
\end{table}

The comparisons reveal a clear split in how thinking affects format compliance. Notably, the three closed-source models (GPT-5.4 mini, Gemini 2.5 Flash, Claude Opus 4.6) already achieve relatively strong format compliance under standard inference, but see perfect format rates \emph{decrease} with thinking, suggesting that the thinking process produces more verbose output that disrupts an already well-structured response. In contrast, the five open-weight models (Qwen 3.5 variants, Gemma 4 31B) start with lower format compliance and benefit substantially from thinking, which helps them correctly interpret question complexity and eliminate \emph{wrong-facet} and \emph{too-few-elements errors}.

Several patterns emerge from the error breakdown. For the open-weight models, thinking nearly eliminates the \emph{too few elements} error; models that previously produced truncated lists now generate the expected number of items. \emph{Wrong facet type} errors also drop sharply, indicating that thinking helps models distinguish between single-faceted and multi-faceted question types. \emph{Mixed facet type}, already rare under standard inference, is virtually eliminated with thinking across all models. The \emph{no list output} error remains negligible for most models regardless of setting, though Gemma 4 31B shows a notable increase with thinking, suggesting its reasoning process occasionally produces unstructured output.

For the closed-source models, the pattern is reversed: thinking \emph{increases} wrong-facet errors, particularly for Gemini 2.5 Flash and Claude Opus 4.6, and GPT-5.4 mini sees a rise in too-few-elements errors. This suggests that the extended reasoning in these models may over-elaborate or restructure the output in ways that deviate from the expected format, even as it improves factual accuracy.

\clearpage
\section{Ordering Metrics - Analysis}
\label{sec:appendix_ordering}

Approximately 14.6\% of \listqa{} samples require ordered responses, i.e., questions that implicitly or explicitly demand a specific sequence (e.g., chronological, ranked). To evaluate this, we introduce an order-aware prompt variant (App.~\ref{subsec:appendix_prompt_zs_order}) that instructs the model to preserve ordering when the question requires it, and evaluate three models under this prompt. We report three scores in Tab.~\ref{tab:ordering_prompt_comparison}: (1) \textit{LLM-Judge-F1} under the original prompt, as a baseline; (2) \textit{LLM-Judge-F1} under the order-aware prompt, to verify that the ordering instruction does not degrade factual accuracy; and (3) a \textit{Blended-F1} score under the order-aware prompt, which applies the stricter \textit{LLM-Judge-F1-Ordered} metric on the 1,319 ordered samples and the standard \judgef{} metric on the remaining 7,726 unordered samples.

\begin{table}[ht]
\caption{F1 metrics on the full dataset (9,045 samples) under different prompt and scoring configurations. $^\dagger$Thinking enabled.}
\vspace{4pt}
\label{tab:ordering_prompt_comparison}
\centering
\small
\begin{tabular}{@{}llccc@{}}
\toprule
& & \multicolumn{2}{c}{\textbf{LLM-Judge-F1}} & \textbf{Blended-F1} \\
\cmidrule(lr){3-4} \cmidrule(lr){5-5}
\textbf{Model} & \textbf{Level} & \textbf{Original Prompt} & \textbf{Order-Aware Prompt} & \textbf{Order-Aware Prompt} \\
\midrule
Gemini 3.1 Pro$^\dagger$       & \textsc{Low}     & 60.7 & 61.2 & 59.4 \\
Gemini 3.1 Flash Lite$^\dagger$ & \textsc{Minimal} & 40.1 & 40.2 & 38.4 \\
GPT-5.4-mini$^\dagger$         & \textsc{Medium}  & 36.2 & 36.6 & 35.0 \\
\bottomrule
\end{tabular}
\end{table}

The first two columns confirm that the order-aware prompt does not degrade factual accuracy. The \textit{Blended-F1} score is slightly lower because it applies the stricter ordered metric on the 14.6\% of samples that require sequencing. The gap is modest since the ordered subset is only a small fraction of the full dataset.

\paragraph{Zooming into the ordered subset reveals a larger gap.}

To isolate the effect of ordering, Tab.~\ref{tab:ordering} restricts evaluation to the 1,319 ordered samples only. Here the gap between the standard and ordered metric becomes more pronounced.

\begin{table}[ht]
\caption{\judgef{} on the \emph{ordered subset only} (1,319 samples) under the order-aware prompt. \textbf{Unord.}: standard \judgef{} (order-agnostic). \textbf{Ord.}: \orderedf{} (penalizes wrong sequence). Results are shown for single-faceted and multi-faceted questions \emph{within the ordered subset}. $^\dagger$Thinking enabled.}
\label{tab:ordering}
\centering
\small
\setlength{\tabcolsep}{3pt}
\begin{tabular}{@{}lcccccc@{}}
\toprule
& \multicolumn{2}{c}{\textbf{Overall}} & \multicolumn{2}{c}{\textbf{Single-faceted}} & \multicolumn{2}{c}{\textbf{Multi-faceted}} \\
\cmidrule(lr){2-3} \cmidrule(lr){4-5} \cmidrule(lr){6-7}
\textbf{Model} & \textbf{Unord.} & \textbf{Ord.} & \textbf{Unord.} & \textbf{Ord.} & \textbf{Unord.} & \textbf{Ord.} \\
\midrule
Gemini 3.1 Pro$^\dagger$       & 55.5 & 43.1 & 62.7 & 47.8 & 45.2 & 36.4 \\
Gemini 3.1 Flash Lite$^\dagger$ & 35.5 & 23.6 & 43.1 & 27.9 & 24.6 & 17.3 \\
GPT-5.4-mini$^\dagger$         & 32.8 & 22.1 & 41.2 & 26.4 & 20.7 & 16.0 \\
Claude Opus 4.6$^\dagger$               & 36.2 & 26.0 & 43.7 & 30.7 & 25.5 & 19.3 \\
\bottomrule
\end{tabular}
\end{table}

On the ordered subset, the ordered metric drops well below the standard metric for all models, confirming that models retrieve correct list elements but frequently produce them in the wrong sequence. This aligns with a central motivation of \listqa{}: evaluating the interplay between factual knowledge and instruction following. Even when models possess the required knowledge, they may fail to follow the implicit or explicit ordering constraint, thereby demonstrating another instance of the knowledge--instruction-following tension observed throughout our experiments.

The gap is consistently larger on single-faceted questions than on multi-faceted ones. This may reflect the nature of the ordered single-faceted subset: these questions often ask for ranked or chronological sequences where the knowledge is simple but ordering is the primary challenge. In multi-faceted ordered questions, the model is already struggling with sub-attribute retrieval (as reflected in lower unordered scores), leaving less room for an additional ordering penalty.

\clearpage
\section{Thinking-Enabled Inference}
\label{sec:appendix_thinking}

Tab.~\ref{tab:thinking} reports exact scores and thinking levels for all models evaluated with thinking enabled. Where a standard-inference baseline exists, we report the difference ($\Delta$). Key findings from this table are discussed in Sec.~\ref{sec:analysis}; here we highlight additional observations.

\begin{table}[ht]
\caption{Effect of thinking-enabled inference on \listqa{}. $\Delta$ denotes the change relative to the same model under standard inference (results in Tab.~\ref{tab:main_results}). Models with paired standard-inference baselines are grouped below the line.}
\vspace{4pt}
\label{tab:thinking}
\centering
\small
\setlength{\tabcolsep}{3pt}
\begin{tabular}{@{}llcccccccc@{}}
\toprule
& & \multicolumn{6}{c}{\textbf{LLM-Judge-F1} $\uparrow$} & \multicolumn{2}{c}{\textbf{Format} $\uparrow$} \\
\cmidrule(lr){3-8} \cmidrule(lr){9-10}
& & \multicolumn{2}{c}{\textbf{Overall}} & \multicolumn{2}{c}{\textbf{S}} & \multicolumn{2}{c}{\textbf{C}} & & \\
\textbf{Model} & \textbf{Think Level} & \textbf{Score} & $\Delta$ & \textbf{Score} & $\Delta$ & \textbf{Score} & $\Delta$ & \textbf{Score} & $\Delta$ \\
\midrule
Gemini 3.1 Pro$^\dagger$  & \textsc{Low} & \textbf{60.7} & --- & \textbf{67.7} & --- & \textbf{50.3} & --- & \textbf{97.7} & --- \\
Gemini 3 Flash$^\dagger$  & \textsc{High} & 56.4 & --- & 63.7 & --- & 45.8 & --- & 91.9 & --- \\
Gemini 3.1 Flash Lite$^\dagger$  & \textsc{Minimal} & 40.1 & --- & 48.4 & --- & 27.9 & --- & 39.9 & --- \\
\midrule
\multicolumn{10}{l}{\textit{Paired comparisons (same model, thinking vs.\ standard)}} \\
\midrule
GPT-5.4-mini          & \textsc{Medium}  & 36.2 & \textcolor{teal}{+9.2} & 45.0 & \textcolor{teal}{+9.8} & 23.4 & \textcolor{teal}{+8.3} & 60.3 & \textcolor{red}{$-$3.8} \\
Gemini 2.5 Flash      & \textsc{Dynamic} & 38.3 & \textcolor{teal}{+4.6} & 47.1 & \textcolor{teal}{+5.2} & 25.4 & \textcolor{teal}{+3.7} & 70.3 & \textcolor{red}{$-$6.1} \\
Claude Opus 4.6$^*$   &                  & 37.9 & \textcolor{teal}{+3.3} & 46.2 & \textcolor{teal}{+3.2} & 25.7 & \textcolor{teal}{+3.3} & 66.8 & \textcolor{red}{$-$11.5} \\
Qwen 3.5 122B-A10B    &                  & 23.5 & \textcolor{teal}{+4.6} & 30.5 & \textcolor{teal}{+5.2} & 13.1 & \textcolor{teal}{+3.4} & 95.1 & \textcolor{teal}{+12.9} \\
Gemma 4 31B           &                  & 18.7 & \textcolor{teal}{+2.7} & 25.3 & \textcolor{teal}{+3.3} & 9.0  & \textcolor{teal}{+1.8} & 74.9 & \textcolor{teal}{+20.6} \\
Qwen 3.5 27B          &                  & 18.2 & \textcolor{teal}{+3.4} & 24.5 & \textcolor{teal}{+4.4} & 9.0  & \textcolor{teal}{+1.9} & 95.3 & \textcolor{teal}{+23.4} \\
Qwen 3.5 35B-A3B      &                  & 18.2 & \textcolor{teal}{+4.8} & 24.4 & \textcolor{teal}{+5.9} & 9.0  & \textcolor{teal}{+3.1} & 90.1 & \textcolor{teal}{+38.4} \\
Qwen 3.5 9B           &                  & 11.4 & \textcolor{teal}{+3.1} & 16.1 & \textcolor{teal}{+4.1} & 4.4  & \textcolor{teal}{+1.5} & 87.2 & \textcolor{teal}{+25.2} \\
\bottomrule
\end{tabular}
\\[4pt]
{\scriptsize $^*$2,048 token reasoning budget | $^\dagger$Preview models, evaluated as of April 21, 2026 | Gemini 3 models do not support non-thinking inference}
\end{table}

\complexlistqa{} remains challenging even with thinking: only one model (Gemini 3.1 Pro) crosses 50\% \judgef{} on the complex subset, and only barely. Most thinking-enabled models remain below 30\% on complex questions, confirming that multi-faceted retrieval with hierarchical sub-attributes is a fundamentally harder task that thinking alone does not resolve.

Across all paired comparisons, the improvement from thinking is consistently larger on \simplelistqa{} than on \complexlistqa{}. This asymmetry suggests that thinking primarily helps models better organize and recall knowledge they already possess, but provides diminishing returns when the underlying retrieval challenge is more demanding. This further suggests that \complexlistqa{} tests a qualitatively different capability than simple list retrieval.

\clearpage
\section{Oracle RAG: Setup}
\label{sec:appendix_oracle_rag}

This section describes how source pages were obtained and how context budgets were set for the Oracle
RAG evaluation in Sec.~\ref{sec:oracle_rag}.

\paragraph{Source pages.} Each sample in \listqa{} includes the Wikipedia URL used during annotation. We retrieve the full page content from a pre-indexed Wikipedia snapshot based on this URL, successfully matching context for 98\% of samples (8,856 of 9,045). The remaining 2\% lack matching content due to URL format variations and can be treated as a natural noise floor. The prompt template (App.~\ref{subsec:appendix_prompt_oracle_rag}) instructs the model to answer \emph{solely} from the provided context, using the same list-formatting instructions as the parametric evaluation. All other evaluation settings (metrics, parsing) are identical.

\paragraph{Context budget.} Closed-source models have context windows large enough to hold every source page, so they always receive the full page. For each open-weight model, we convert its context window into a maximum number of source characters at roughly 3 characters per token: about 800K characters for the 256K-token windows of the Gemma 4 models, and 400K characters for the 128K-token windows of Magistral Small 24B and Ministral 3 8B/3B. This conservative conversion leaves headroom for the instruction prompt and generation. Models with smaller context windows may therefore not receive the full page when the source is unusually long.

\clearpage
\section{Timeless vs. Time-constrained Knowledge Retrieval}
\label{sec:appendix_year_range}

\begin{figure}[ht]
    \centering
    \begin{minipage}[b]{0.52\textwidth}
        \centering
        \includegraphics[width=\textwidth]{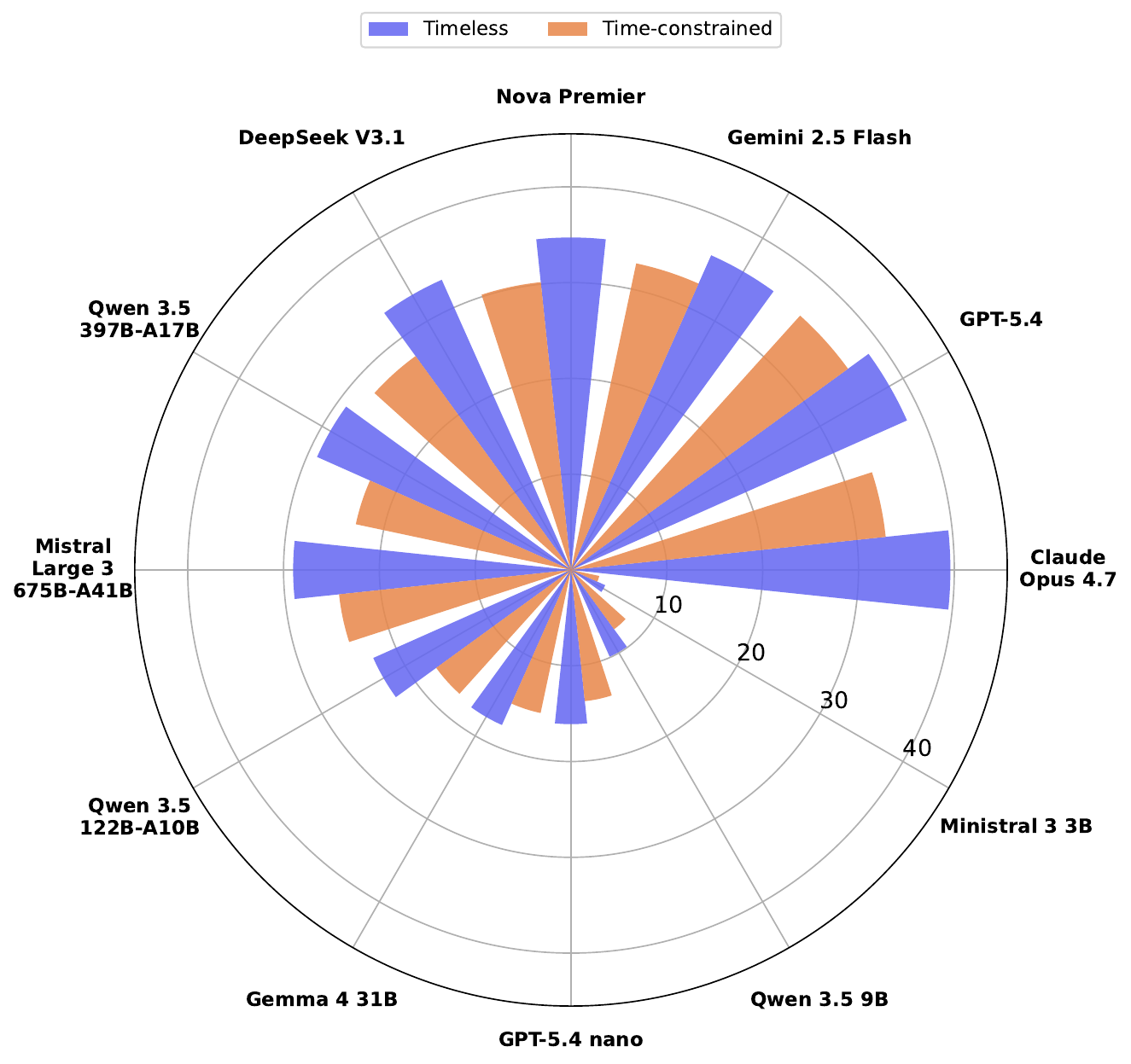}
    \end{minipage}%
    \hfill
    \begin{minipage}[b]{0.48\textwidth}
        \centering
        \includegraphics[width=\textwidth]{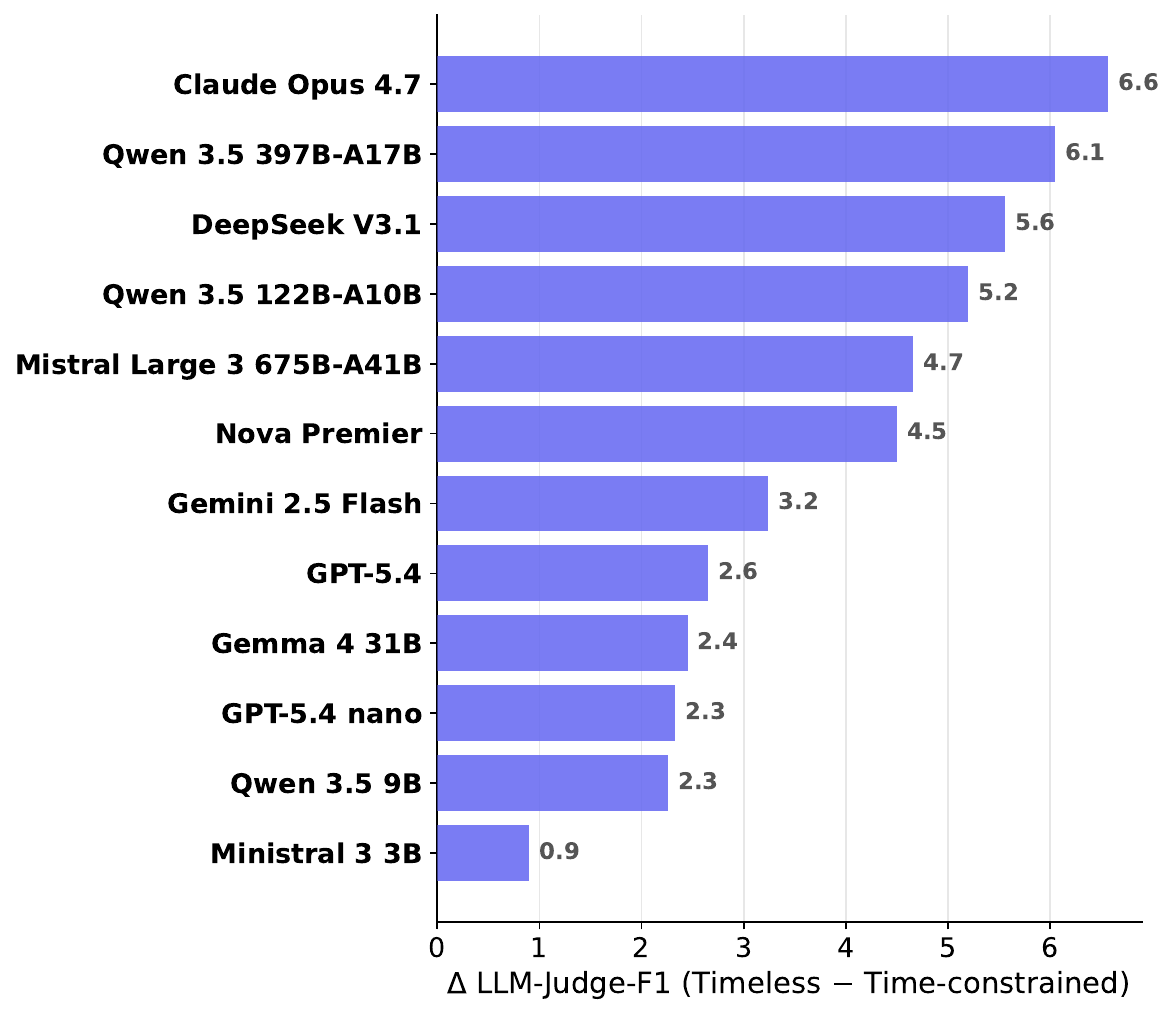}
    \end{minipage}
    \caption{\textbf{Left:} LLM-Judge-F1 on timeless vs.\ time-constrained questions. Every model scores higher on timeless questions. \textbf{Right:} The gap (timeless $-$ time-constrained), sorted by magnitude.}
    \label{fig:year_range}
\end{figure}

To understand how temporal properties affect performance, we split the dataset by whether questions include explicit year-range constraints (\emph{time-constrained}, 70.1\% of samples) or target historically immutable facts (\emph{timeless}, 29.9\%). Fig.~\ref{fig:year_range} shows LLM-Judge-F1 for a representative subset of models across both splits.

Every model without exception performs better on timeless questions than on time-constrained ones. This pattern holds across all standard-inference models evaluated, with the gap varying by model. The finding is consistent with the intuition that historically fixed facts are more likely to be well-represented and reinforced across training corpora, while time-bounded questions require recall of specific facts tied to particular periods: knowledge that may appear less frequently or less consistently in training data.

The gap is notably larger for some model families, suggesting relatively less coverage of temporally specific knowledge compared to their recall of stable historical facts. Other models show smaller gaps, indicating more uniform knowledge coverage across temporal categories.

This analysis validates a core design choice of \listqa{}: by including both time-constrained and timeless questions, the benchmark can distinguish between models that have broad but shallow temporal coverage and those with deeper knowledge of specific time periods.

\clearpage
\section{Effect of List Length on Performance}
\label{sec:appendix_list_length}

\begin{figure}[ht]
    \centering
    \includegraphics[width=\textwidth]{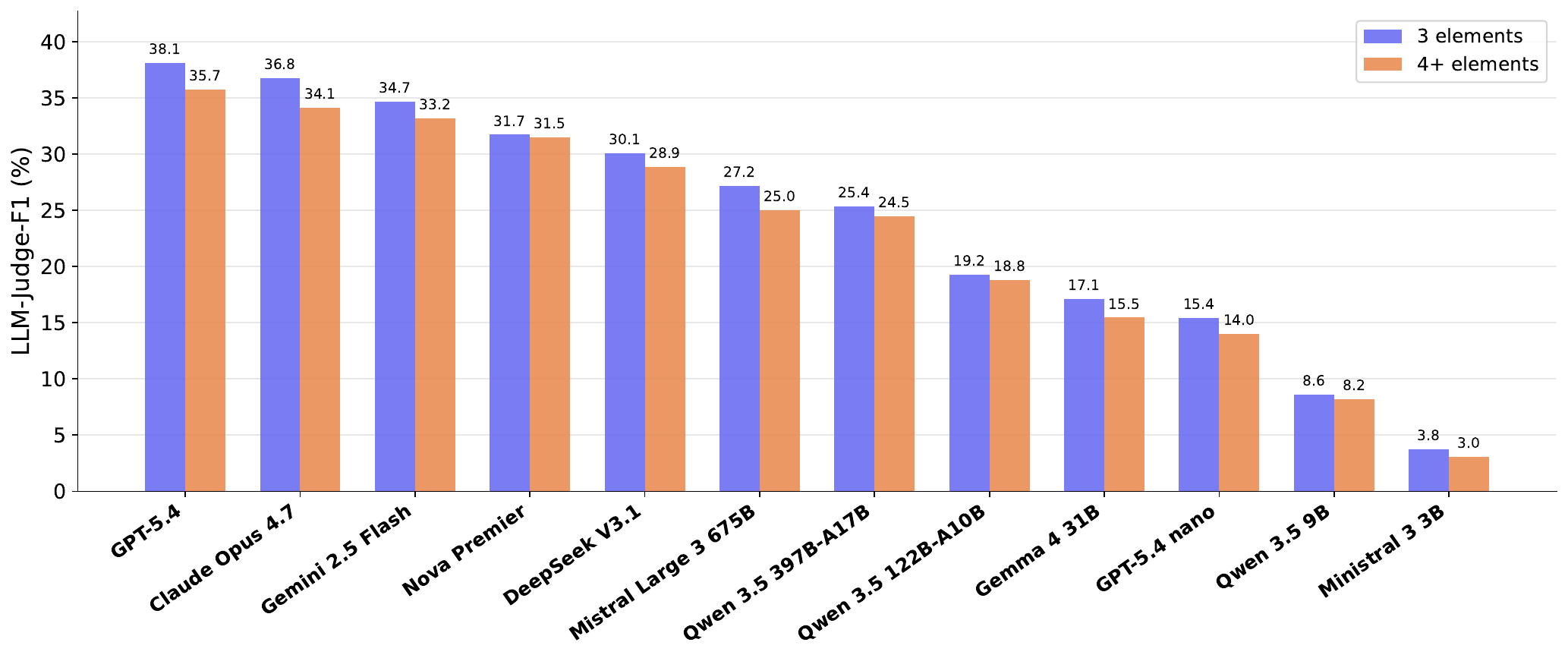}
    \caption{\judgef{} on questions with exactly 3 reference elements vs.\ 4 or more elements in the reference answer. Every model performs better on shorter lists, though the gap is modest.}
    \label{fig:list_length}
\end{figure}

To understand how the number of required list elements affects model performance, we split the dataset by the number of primary elements in the reference answer (ranging from 3 to 8) and compare scores on short lists (exactly 3 elements, 33\% of samples) against longer lists (4 or more elements, 67\%). Fig.~\ref{fig:list_length} shows LLM-Judge-F1 for a representative subset of models across both splits.

Every model scores higher on 3-element questions than on questions requiring 4 or more elements, consistent with the intuition that retrieving fewer facts from parametric memory is easier than retrieving many. However, the gap is notably small across all models (typically 1--3 percentage points), suggesting that list length alone is not a dominant difficulty factor in \listqa{}. This contrasts with the much larger gaps observed between single-faceted and multi-faceted questions (Tab.~\ref{tab:main_results}), indicating that the structural complexity of sub-attributes poses a greater challenge than the sheer number of elements to retrieve.

Within the longer lists, we observe that performance is relatively stable across lengths 4 through 8. The absence of a strong monotonic decline suggests that once a model can retrieve 4 elements, the marginal difficulty of retrieving additional elements is modest. This finding validates the dataset's design choice of bounding list length between 3 and 8: the range is sufficient to test multi-element retrieval without introducing a confounding length penalty that would dominate the signal from knowledge and formatting capabilities.

\clearpage
\section{Performance by Topic Category}
\label{sec:appendix_category}

\begin{figure}[ht]
    \centering
    \includegraphics[width=\textwidth,height=0.55\textheight,keepaspectratio]{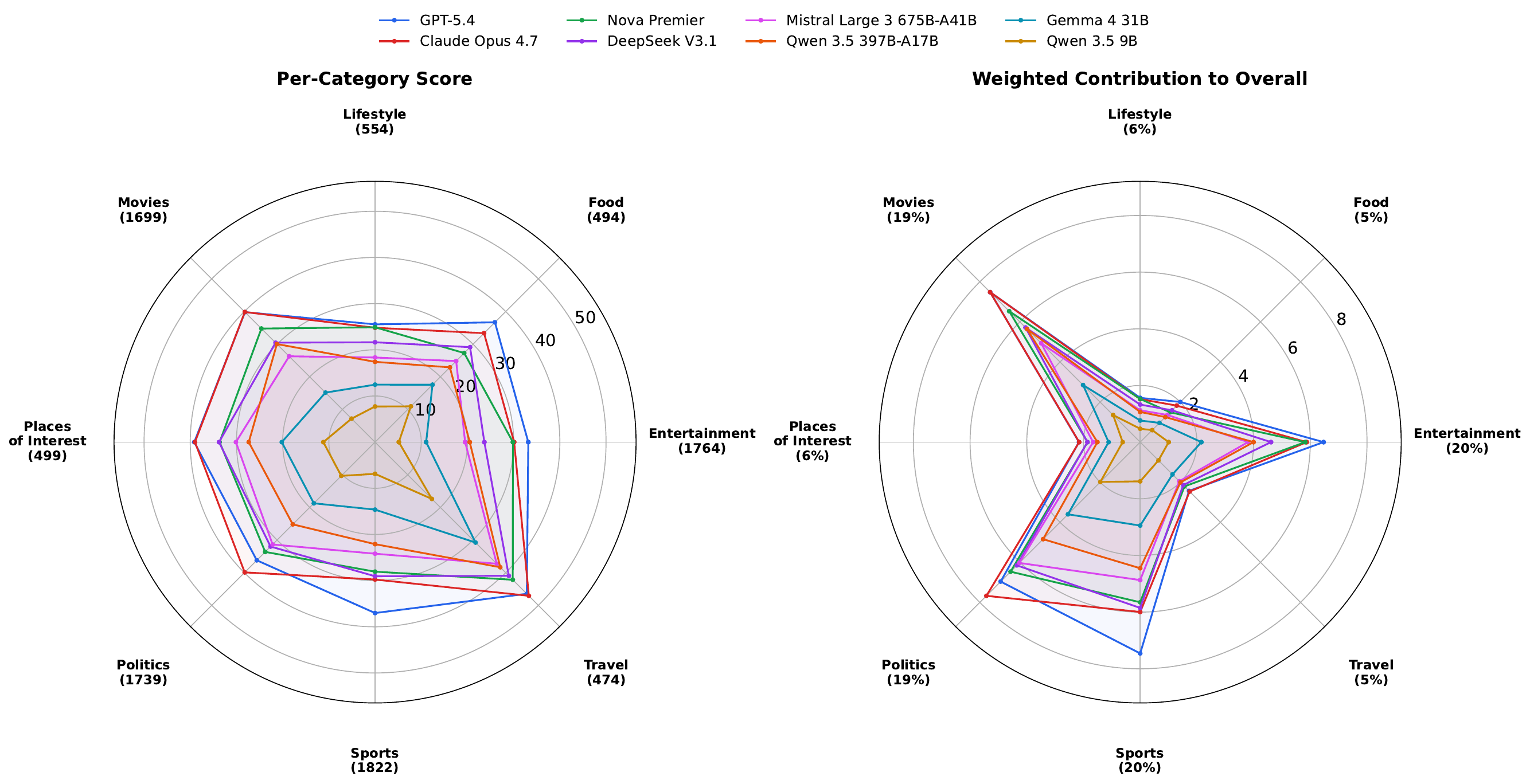}
    \caption{\textbf{Left:} \judgef{} per category (sample counts in parentheses). \textbf{Right:} Weighted contribution to overall \judgef{} score (category proportion in parentheses). The radar shape is consistent across models in both views, differing primarily in scale.}
    \label{fig:category_radar}
\end{figure}

To examine whether certain knowledge domains are systematically easier or harder for models, we break down \judgef{} by the eight topic categories in \listqa{}. Fig.~\ref{fig:category_radar} shows two complementary views for a subset of models across multiple families: raw per-category scores (left) and each category's weighted contribution to the overall score (right), where the weight is the category's proportion of the dataset.

The raw scores (left) reveal that \textit{Travel} is universally the easiest category and \textit{Lifestyle} universally the hardest, with all models sharing a strikingly similar profile shape that differs only in scale. This consistency suggests that category difficulty is an intrinsic property of the dataset's knowledge distribution rather than a model-specific weakness.

The weighted view (right) tells a complementary story: the shape inverts. The four large categories: Sports, Entertainment, Politics, and Movies (each $\sim$19--20\% of samples), dominate the overall score, while Travel and Lifestyle, despite being the easiest and hardest respectively, contribute minimally due to their small share ($\sim$5--6\% each - See distribution in Fig.~\ref{fig:dataset_overview}). This means the overall benchmark score is primarily determined by performance on the mid-difficulty categories that cluster in a narrow band, making it a stable aggregate measure not skewed by outlier categories at either extreme.

\clearpage
\section{Performance by Geographic Region}
\label{sec:appendix_continent}

\begin{figure}[ht]
    \centering
    \includegraphics[width=\textwidth,height=0.55\textheight,keepaspectratio]{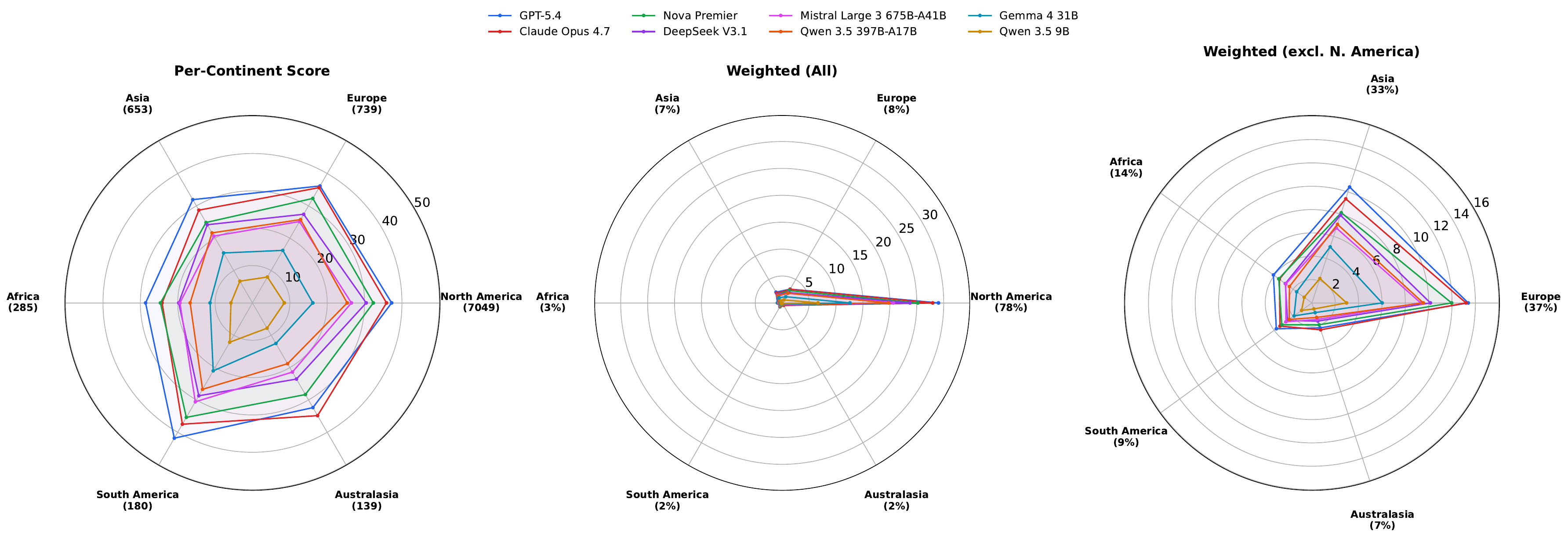}
    \caption{\textbf{Left:} LLM-Judge-F1 per continent (sample counts in parentheses). \textbf{Center:} Weighted contribution to overall score (continent proportion in parentheses). North America dominates the overall score due to its 78\% share of the dataset. \textbf{Right:} Weighted contribution excluding North America (renormalized proportions).}
    \label{fig:continent_radar}
\end{figure}

To assess whether models exhibit geographic knowledge biases, we group samples by the continent of their source Wikipedia content and compare \judgef{} across regions. Fig.~\ref{fig:continent_radar} shows raw per-continent scores (left) and weighted contributions to the overall score (middle and right).

The continent breakdown reveals a clear geographic hierarchy: models consistently perform best on South American and North American/European content, and worst on African and Asian content. This gap persists across all model families and scales, suggesting that training corpora systematically underrepresent factual knowledge about Africa and Asia relative to the Americas and Europe.

The weighted view (Fig.~\ref{fig:continent_radar} (center)) makes the practical implication stark: North America alone accounts for 78\% of the dataset, so the overall benchmark score is overwhelmingly determined by performance on US-centric content. The regions where models struggle most: Africa and Asia, contribute minimally to the aggregate despite revealing the largest knowledge gaps. This highlights a limitation of the current dataset composition and motivates future geographic rebalancing to ensure the benchmark measures global factual knowledge rather than primarily Western-centric recall.

When North America is excluded (Fig.~\ref{fig:continent_radar} (right)), Europe and Asia together account for roughly 70\% of the remaining weighted contribution, making the Europe--Africa performance gap the dominant signal. South America achieves the highest raw scores of any non-US region for most models, yet its small sample share limits its influence even in this reweighted view.

\clearpage
\section{Effect of Instruction Tuning}
\label{sec:appendix_base_models}

To isolate the contribution of instruction tuning to \listqa{} performance, we evaluate a few base (pre-trained, non-instruction-tuned) models using a minimal chat template in a few-shot setting (prompt template in App.~\ref{subsec:appendix_few_shot_prompt}), and compare them against their instruction-tuned (IT) counterparts evaluated zero-shot. For IT models, we additionally report results with thinking enabled at inference.

This asymmetry in prompting is deliberate. Base models have had little exposure to instruction following, so a zero-shot prompt would conflate missing knowledge with a failure to understand the requested output format. Few-shot prompting is the minimal condition that reliably elicits list-formatted output from a base model, allowing its factual knowledge to be measured. IT models, by contrast, acquire format-following during post-training and are evaluated in the same zero-shot setting as the main results (Sec.~\ref{sec:exp_setup}). The comparison therefore isolates the effect of training stage rather than prompting protocol.

\begin{table}[ht]
\caption{Base vs.\ instruction-tuned (IT) performance. Base models are evaluated few-shot; IT models are evaluated zero-shot, with and without thinking enabled at inference. All values are percentages. All = Overall, S = \simplelistqa{}, C = \complexlistqa{}.}
\vspace{4pt}
\label{tab:base_models}
\centering
\small
\setlength{\tabcolsep}{2.5pt}
\begin{tabular}{@{}llccccccc@{}}
\toprule
& & \multicolumn{3}{c}{\textbf{LLM-Judge-F1} $\uparrow$} & \multicolumn{3}{c}{\textbf{Format} $\uparrow$} \\
\cmidrule(lr){3-5} \cmidrule(lr){6-8}
\textbf{Model} & \textbf{Setting} & \textbf{All} & \textbf{S} & \textbf{C} & \textbf{All} & \textbf{S} & \textbf{C} \\
\midrule
\multirow{3}{*}{Gemma 4 31B}
& Base        & 9.1  & 12.6 & 4.0  & 51.7 & 51.7 & 51.8 \\
& IT          & 16.0 & 22.0 & 7.3  & 54.3 & 45.8 & 66.7 \\
& IT + Think  & 18.7 & 25.3 & 9.0  & 74.9 & 73.9 & 76.5 \\
\midrule
\multirow{3}{*}{Qwen 3.5 9B}
& Base        & 4.9  & 7.2  & 1.5  & 55.9 & 54.6 & 57.8 \\
& IT          & 8.3  & 12.0 & 2.9  & 62.0 & 65.5 & 56.8 \\
& IT + Think  & 11.4 & 16.1 & 4.4  & 87.2 & 87.0 & 87.5 \\
\midrule
\multirow{3}{*}{Qwen 3.5 35B-A3B}
& Base        & 8.6  & 12.0 & 3.6  & 52.0 & 49.8 & 55.4 \\
& IT          & 13.4 & 18.5 & 5.9  & 51.7 & 37.0 & 73.2 \\
& IT + Think  & 18.2 & 24.4 & 9.0  & 90.1 & 89.6 & 91.0 \\
\bottomrule
\end{tabular}
\end{table}

\textbf{Results.} Tab.~\ref{tab:base_models} reports \judgef{} and \formatmetric{} across the base, IT, and IT with thinking variants. Instruction tuning substantially improves LLM-Judge-F1 over the base model across all three models. Enabling thinking at inference on the same IT checkpoint adds a further improvement, consistent across model scales. The two stages contribute differently: IT improves the model's ability to produce structured, relevant responses, while thinking further enhances factual retrieval and output organization at inference time.

It is interesting to see that base models, evaluated in a few-shot setting, already achieve moderate format scores. This demonstrates that even without instruction tuning, these models can learn list-formatting conventions from in-context examples alone. Instruction tuning provides only marginal further improvement in format compliance. Enabling thinking on the IT checkpoint, however, produces a dramatic jump in format scores across all three models, suggesting that the reasoning process at inference is particularly effective at helping models interpret formatting instructions and produce the correct list structure.

\begin{table}[ht]
\caption{Format error distribution (\% of 9,045 samples) for Base vs. IT models.}
\vspace{4pt}
\label{tab:base_format_errors}
\centering
\small
\setlength{\tabcolsep}{3.5pt}
\begin{tabular}{@{}llccccc@{}}
\toprule
\textbf{Model} & \textbf{Setting} & \textbf{Perfect} & \textbf{Wrong Facet} & \textbf{Too Few} & \textbf{No List} & \textbf{Mixed} \\
\midrule
\multirow{2}{*}{Gemma 4 31B}
& Base & 43.6 & 23.3 & 3.6 & 13.9 & 15.6 \\
& IT   & 53.9 & 14.4 & 29.0 & 2.0 & 0.7 \\
\midrule
\multirow{2}{*}{Qwen 3.5 9B}
& Base & 37.7 & 20.5 & 6.3 & 1.6 & 34.0 \\
& IT   & 55.4 & 19.8 & 11.0 & 2.8 & 11.0 \\
\midrule
\multirow{2}{*}{Qwen 3.5 35B-A3B}
& Base & 42.5 & 23.0 & 10.2 & 5.8 & 18.6 \\
& IT   & 49.8 & 26.0 & 20.2 & 0.6 & 3.5 \\
\bottomrule
\end{tabular}
\end{table}

\textbf{Format error analysis.} Tab.~\ref{tab:base_format_errors} breaks down format errors across the base and IT variants. The format error profiles reveal qualitatively different failure modes between base and IT models. Base models produce a higher rate of \emph{mixed facet type} errors, i.e., responses mixing sub-bullets and flat elements inconsistently, suggesting that without instruction tuning, models lack a coherent formatting strategy. Instruction tuning largely eliminates mixed-facet errors but introduces a shift toward \emph{too few elements}, suggesting that IT potentially makes models more conservative, sometimes undershooting the minimum list length in favor of more structured output.

\clearpage

\section{Validation Against Human Judgment}
\label{sec:appendix_corr_hum_perf}

\subsection{Human Evaluation of \judgef{}}
\label{sec:appendix_human_eval}

We selected about 200 questions stratified across the 8 categories and both question types, and collected responses from three models spanning the score range: GPT-5.4, Qwen 3.5 122B-A10B, and Claude Opus 4.7. Human graders labeled each of the resulting 3,313 predicted list elements (571 sample--model pairs) as CORRECT or INCORRECT using the same criteria as the judge prompt (App.~\ref{subsec:appendix_judge_llm_prompt}). Human precision, recall, and F1 were computed exactly as for \judgef{}.

\begin{table}[ht]
\caption{Sample-level agreement between \judgef{} (GPT-OSS-120B) and human F1, by model.}
\vspace{4pt}
\label{tab:human_eval_model}
\centering
\small
\setlength{\tabcolsep}{3pt}
\begin{tabular}{@{}lccccccccc@{}}
\toprule
\textbf{Model} & \textbf{N} & \textbf{Elem.} & \textbf{Hum.\ P} & \textbf{Hum.\ R} & \textbf{Hum.\ F1} &
\textbf{Judge F1} & \textbf{Pearson $r$} & \textbf{Spearman $\rho$} \\
\midrule
GPT-5.4            & 200 & 1,077 & 0.346 & 0.372 & 0.349 & 0.328 & 0.974 &  0.974 \\
Qwen 3.5 122B-A10B & 191 & 1,330 & 0.155 & 0.191 & 0.164 & 0.150 & 0.951 &  0.922 \\
Claude Opus 4.7    & 180 & 906   & 0.385 & 0.383 & 0.374 & 0.346 & 0.958 &  0.951 \\
\midrule
\textbf{Overall}   & 571 & 3,313 & -- & -- & -- & -- & \textbf{0.965} &  \textbf{0.956}
\\
\bottomrule
\end{tabular}
\end{table}

\begin{table}[ht]
\caption{Sample-level agreement between \judgef{} and human F1, by topic category and question type.}
\vspace{4pt}
\label{tab:human_eval_breakdown}
\centering
\small
\setlength{\tabcolsep}{4pt}
\begin{tabular}{@{}lcccccc@{}}
\toprule
& \textbf{N} & \textbf{Elem.} & \textbf{Hum.\ F1} & \textbf{Judge F1} & \textbf{Pearson $r$} &
\textbf{Spearman $\rho$} \\
\midrule
\multicolumn{7}{@{}l}{\textit{By category}} \\
Entertainment      & 75 & 380 & 0.293 & 0.269 & 0.958 & 0.974 \\
Food               & 73 & 462 & 0.348 & 0.322 & 0.954 & 0.954 \\
Lifestyle          & 75 & 431 & 0.168 & 0.171 & 0.997 & 0.974 \\
Movies             & 73 & 456 & 0.338 & 0.318 & 0.981 & 0.968 \\
Places of Interest & 70 & 410 & 0.351 & 0.298 & 0.924 & 0.895 \\
Politics           & 65 & 439 & 0.301 & 0.274 & 0.965 & 0.937 \\
Sports             & 70 & 323 & 0.294 & 0.288 & 0.970 & 0.947 \\
Travel             & 70 & 412 & 0.275 & 0.256 & 0.983 & 0.978 \\
\midrule
\multicolumn{7}{@{}l}{\textit{By question type}} \\
\simplelistqa{}    & 285 & 1,674 & 0.406 & 0.388 & 0.974 & 0.973 \\
\complexlistqa{}   & 286 & 1,639 & 0.185 & 0.160 & 0.942 & 0.921 \\
\bottomrule
\end{tabular}
\end{table}

 \begin{table}[ht]
\caption{Element-level agreement with human labels ($N = 3{,}313$).}
\vspace{4pt}
\label{tab:human_eval_element}
\centering
\small
\begin{tabular}{@{}lcc@{}}
\toprule
\textbf{Comparison} & \textbf{Agreement} & \textbf{Cohen's $\kappa$} \\
\midrule
Human vs.\ \judgef{}          & 92.8\% & 0.793 \\
Human vs.\ ROUGE-1 $\geq$ 0.5 & 89.3\% & 0.735 \\
\bottomrule
\end{tabular}
\end{table}

\textbf{Results.} \judgef{} correlates strongly with human F1 (Pearson $r = 0.965$, Spearman $\rho = 0.956$; Tab.~\ref{tab:human_eval_model}). Agreement is high for all three models ($r \geq 0.95$), and \judgef{} is slightly below human F1 for each, so the metric is mildly conservative rather than inflating scores. Correlation is also stable across all 8 categories ($r \geq 0.92$) and both question types (Tab.~\ref{tab:human_eval_breakdown}). It is slightly lower on \complexlistqa{} than on \simplelistqa{}, but alignment on the harder hierarchical questions remains strong. At the element level (Tab.~\ref{tab:human_eval_element}), the LLM judge agrees with humans more often than ROUGE thresholding ($\kappa = 0.793$ vs.\ $0.735$), so semantic grading adds value beyond lexical overlap.

\subsection{Validating the ROUGE-Based Matching Step}
\label{sec:appendix_matching_validation}

\judgef{} uses ROUGE-1 overlap only to \emph{align} predicted and reference elements; correctness is then graded semantically by the judge. Lexical overlap is a suitable alignment signal for \listqa{} because list elements are short entities or phrases (names, dates, numbers, titles), where paraphrase is rare.

\textbf{Alignment accuracy.} We manually verified the 3,313 alignments produced by the Jonker--Volgenant assignment in App.~\ref{sec:appendix_human_eval}. Of these, 3,303 were correct (99.7\%).

\textbf{Sensitivity to the matching signal.} We replaced ROUGE-1 with cosine similarity over \texttt{all-MiniLM-L6-v2} sentence embeddings and re-scored all 9,045 samples for two models (Tab.~\ref{tab:matching_ablation}). Over 90\% of samples receive identical scores, and aggregate \judgef{} changes by less than 0.002. When the two strategies disagree, the reassigned element is typically incorrect under either alignment, so the verdict rarely changes. Embedding-based matching therefore adds cost without measurable benefit for short factual elements.

\begin{table}[ht]
\caption{\judgef{} under ROUGE-1 vs.\ embedding-based matching (9,045 samples). \textbf{\% Identical}:
samples with the same score under both strategies.}
\vspace{4pt}
\label{tab:matching_ablation}
\centering
\small
\begin{tabular}{@{}lccc@{}}
\toprule
\textbf{Model} & \textbf{ROUGE-1} & \textbf{Embedding} & \textbf{\% Identical} \\
\midrule
Claude Opus 4.6   & 0.346 & 0.3451 & 91.3 \\
Claude Sonnet 4.5 & 0.253 & 0.2517 & 94.1 \\
\bottomrule
\end{tabular}
\end{table}

\subsection{Judge Model Agreement Analysis}
\label{sec:appendix_judge_agreement}

A key concern with LLM-as-judge evaluation is whether the metric is robust to the choice of judge model, and whether practitioners can use smaller, open-weight models as affordable alternatives. To assess this, we rescore all 9,045 samples from a single evaluated model (Claude Opus 4.6 under standard inference) using three open-weight reasoning models: Qwen 3.5 27B, Qwen 3.5 9B, and Qwen 3.5 4B, alongside the default GPT-OSS-120B judge used throughout the paper. This produces four independent \judgef{} scores per sample, enabling a systematic analysis of inter-judge agreement.

We report two complementary agreement metrics, each capturing a different aspect of reliability.

\paragraph{Mean Absolute Difference (MAD).} For each pair of judges, MAD computes the average absolute score difference across all samples:
\begin{equation}
\text{MAD}(A, B) = \frac{1}{N} \sum_{i=1}^{N} |s^A_i - s^B_i|
\end{equation}
where $s^A_i$ and $s^B_i$ are the \judgef{} scores assigned by judges $A$ and $B$ to sample $i$. MAD is directly interpretable in the same units as the metric: a MAD of 0.02 means scores differ by 0.02 points on average. We additionally report the mean signed difference $\frac{1}{N}\sum(s^A_i - s^B_i)$ to distinguish systematic bias (one judge consistently more lenient) from random disagreement.

\paragraph{Spearman rank correlation ($\rho$).} Spearman correlation measures whether two judges produce the same \emph{ranking} of samples, regardless of the absolute scores. Each judge's scores are replaced by their ranks (with ties receiving the average rank), and Pearson correlation is computed on the ranks. Values range from $-1$ (perfectly opposite rankings) to $+1$ (identical rankings). Spearman is robust to systematic offsets, i.e., if one judge scores consistently higher but preserves the same ordering, $\rho$ remains high.

\paragraph{Results.}

\begin{table}[ht]
\caption{Basic statistics for each judge model. Scores are LLM-Judge-F1 (0--1) across 9,045 samples from Claude Opus 4.6 under standard inference.}
\vspace{4pt}
\label{tab:judge_basic_stats}
\centering
\small
\begin{tabular}{@{}lcc@{}}
\toprule
\textbf{Judge Model} & \textbf{LLM-Judge-F1} & \textbf{Zero-scored (\%)} \\
\midrule
GPT-OSS-120B & 0.346 & 39.2 \\
Qwen 3.5 27B & 0.362 & 37.8 \\
Qwen 3.5 9B & 0.345 & 39.7 \\
Qwen 3.5 4B & 0.354 & 38.5 \\
\bottomrule
\end{tabular}
\end{table}

\begin{table}[ht]
\caption{Pairwise agreement between judge models. MAD: mean absolute difference (lower = more agreement). Signed Diff: mean signed difference (positive = first judge scores higher). Spearman $\rho$: rank correlation (higher = more agreement).}
\vspace{4pt}
\label{tab:judge_pairwise}
\centering
\small
\begin{tabular}{@{}lccc@{}}
\toprule
\textbf{Judge Pair} & \textbf{MAD} $\downarrow$ & \textbf{Signed Diff} & \textbf{Spearman} $\rho$ $\uparrow$ \\
\midrule
GPT-OSS vs.\ Qwen 3.5 27B & 0.022 & $-$0.015 & 0.973 \\
GPT-OSS vs.\ Qwen 3.5 9B & 0.024 & $+$0.002 & 0.966 \\
GPT-OSS vs.\ Qwen 3.5 4B & 0.022 & $-$0.008 & 0.972 \\
\bottomrule
\end{tabular}
\end{table}

All judges exhibit strong agreement (see Tabs.~\ref{tab:judge_basic_stats}--\ref{tab:judge_pairwise}). Scores differ by only 0.02--0.024 points on average (MAD), and all pairwise Spearman correlations exceed 0.96, indicating near-identical sample rankings (see Tab.~\ref{tab:judge_pairwise}). Notably, even the smallest judge (Qwen 3.5 4B) achieves agreement on par with the 27B variant, with a Spearman $\rho$ of 0.972 and MAD of 0.022. The score distributions are also closely matched: all judges assign zero scores to a similar fraction of samples (37.8--39.7\%).

These results demonstrate that open-weight models as small as 4B parameters can serve as reliable judges for the \listqa{} evaluation, producing both similar absolute scores and nearly identical model rankings compared to the 120B default judge. Practitioners who prefer a lighter-weight evaluation pipeline can substitute any of the Qwen 3.5 models with minimal impact on conclusions, keeping in mind the small random noise factor (MAD $\approx$ 0.02) when comparing models with very close scores.

\subsection{Judge Repeatability}
\label{sec:appendix_judge_repeatability}

To measure judge repeatability directly, we re-score three models five times each with GPT-OSS-120B at
temperature 0.7. Tab.~\ref{tab:judge_repeatability} compares the reported score with the mean and range across the five runs, together with the percentage of samples that receive the same score in all five runs.

\begin{table}[ht]
\caption{Repeatability of \judgef{} across five independent re-scoring runs with GPT-OSS-120B (temperature 0.7). \textbf{Reported}: score in Tab.~\ref{tab:main_results}.
\textbf{\% Identical}: percentage of the 9,045 samples that receive the same score in all five runs.}
\vspace{4pt}
\label{tab:judge_repeatability}
\centering
\small
\begin{tabular}{@{}lcccc@{}}
\toprule
\textbf{Model} & \textbf{Reported} & \textbf{Mean (5 runs)} & \textbf{Min / Max} & \textbf{\% Identical}
\\
\midrule
Claude Opus 4.6   & 0.3460 & 0.3469 & 0.3465 / 0.3473 & 86.3 \\
Claude Sonnet 4.6 & 0.2800 & 0.2803 & 0.2794 / 0.2811 & 88.0 \\
Claude Sonnet 4.5 & 0.2530 & 0.2531 & 0.2526 / 0.2536 & 89.9 \\
\bottomrule
\end{tabular}
\end{table}
  
\judgef{} is highly reproducible. Across runs, scores vary by at most 0.0017, far less than the gaps between models in Tab.~\ref{tab:main_results}, and the large majority of samples receive the same score in all five runs. Together with App.~\ref{sec:appendix_judge_agreement}, this shows that the metric is stable both when the same judge is re-run and when a different judge is used.

\clearpage
\section{Token Usage and Generation Efficiency}
\label{sec:appendix_token_usage}

We analyze the relationship between generation cost (measured in tokens) and factual accuracy across all evaluated models. For models where we are able to separate reasoning tokens from output tokens, we report both components. For others, we do not provide this separation, i.e., reasoning tokens are included in the output token count. Total generation cost is defined as reasoning tokens + output tokens for all models.

\subsection{Generation Cost vs.\ Accuracy}
\label{subsec:appendix_generationcost_vs_accuracy}

\begin{figure}[ht]
    \centering
    \includegraphics[width=\textwidth]{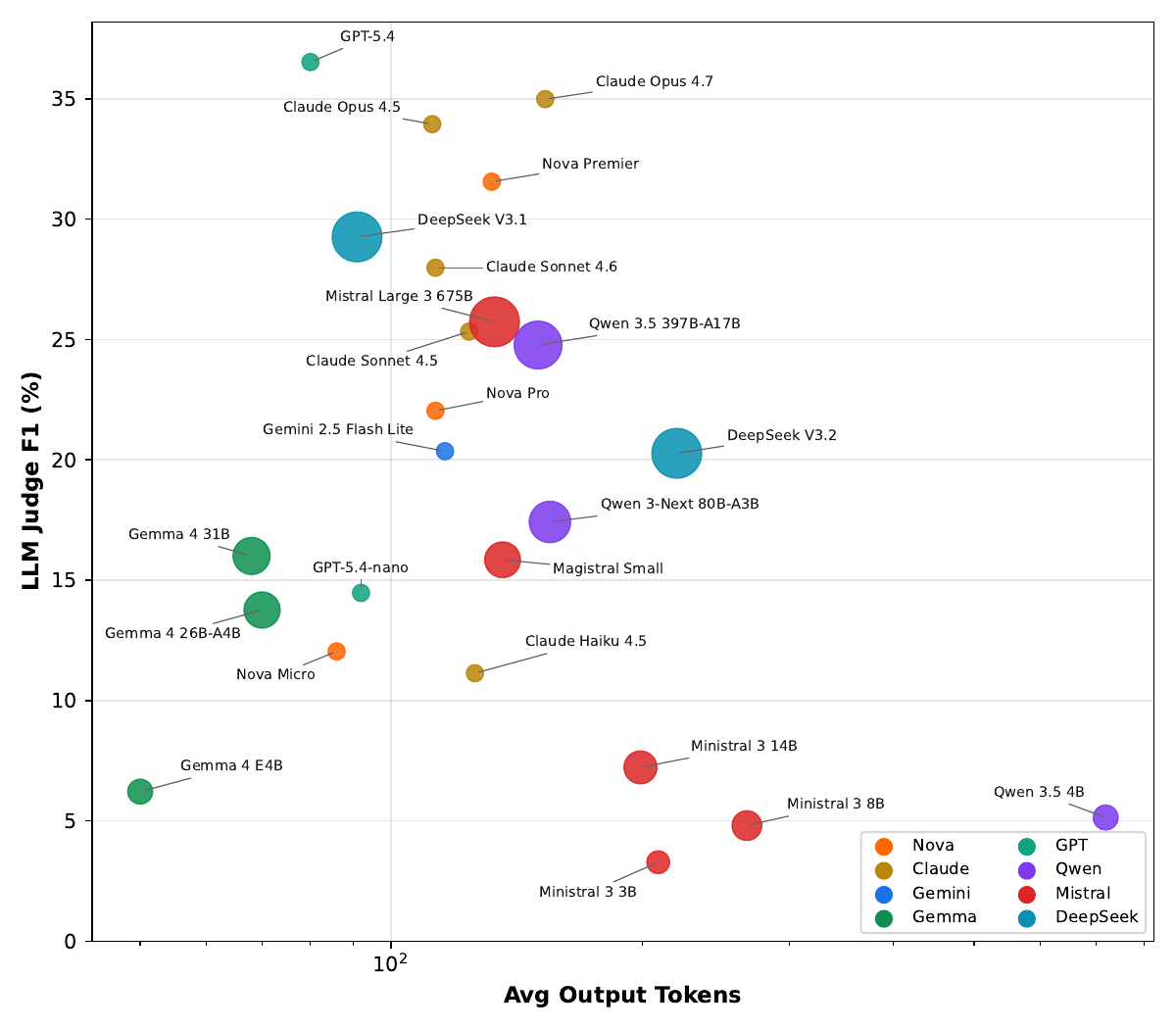}
    \caption{\textit{LLM-Judge-F1} vs.\ average output tokens for non-thinking models. Marker color indicates model family; marker size is proportional to parameter count for open-weight models (fixed size for closed-source). Higher and further left is better.}
    \label{fig:efficiency_nothink}
\end{figure}

Fig.~\ref{fig:efficiency_nothink} plots each non-thinking model's average total generation tokens against its overall LLM-Judge-F1 score. The highest-scoring models achieve the best accuracy with moderate output lengths, while several open-weight models generate substantially more tokens for lower scores. Generation verbosity is not a proxy for factual knowledge: weaker models tend to produce longer, less precise outputs rather than concise correct answers.

\begin{figure}[ht]
    \centering
    \includegraphics[width=\textwidth]{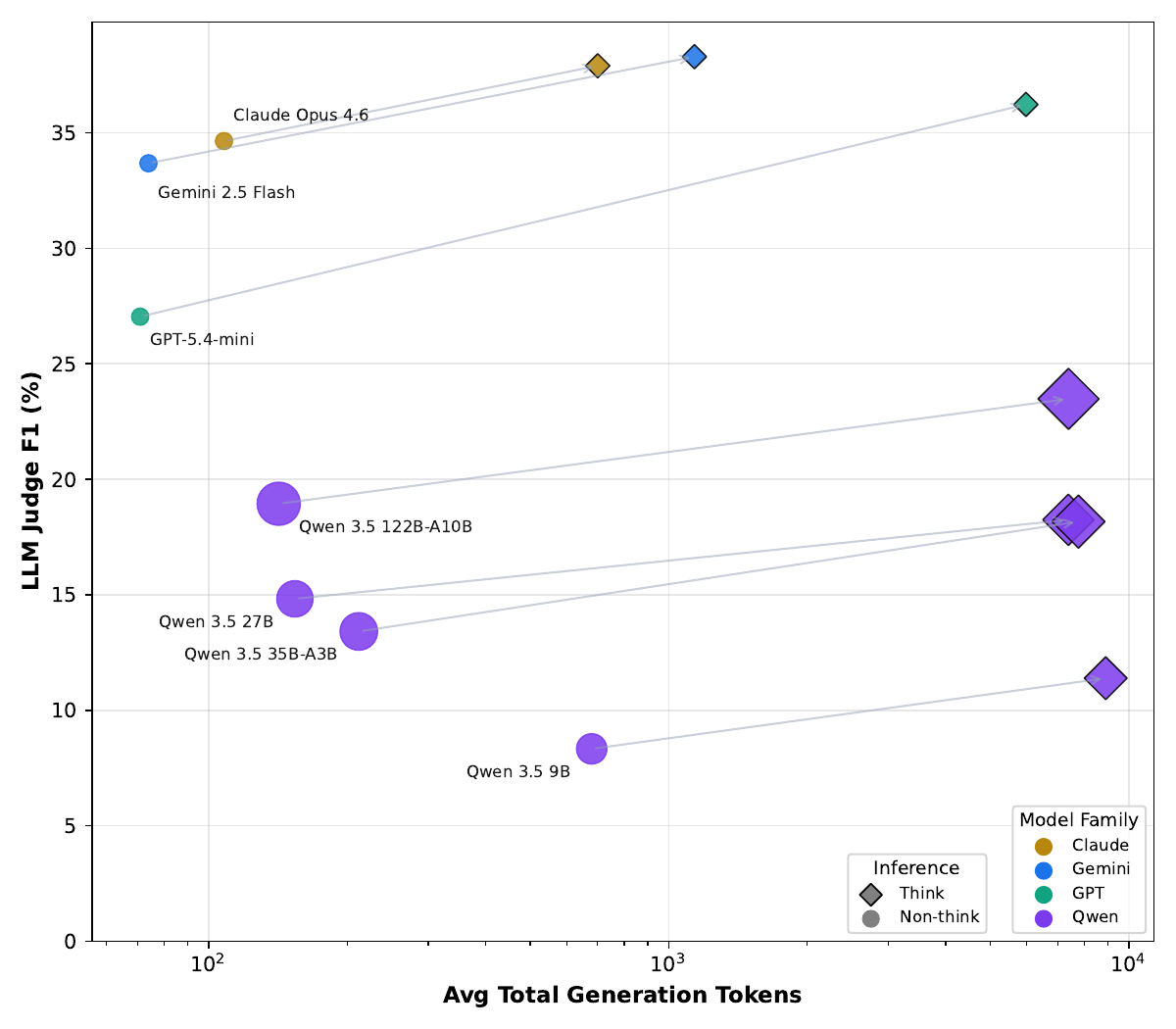}
    \caption{\textit{LLM-Judge-F1} vs.\ average total generation tokens per sample for models with both thinking and non-thinking configurations. Arrows point from non-thinking (circle) to thinking (diamond) variants.}
    \label{fig:efficiency_think_pairs}
\end{figure}

Fig.~\ref{fig:efficiency_think_pairs} extends this analysis to models with both thinking and non-thinking configurations. Thinking universally shifts models rightward (more tokens) and upward (higher F1 score), but the magnitude of both shifts varies dramatically across architectures, a pattern we quantify next.

\subsection{The Cost of Thinking}
\label{subsec:appendix_cost_of_thinking}

\begin{figure}[ht]
    \centering
    \includegraphics[width=\textwidth]{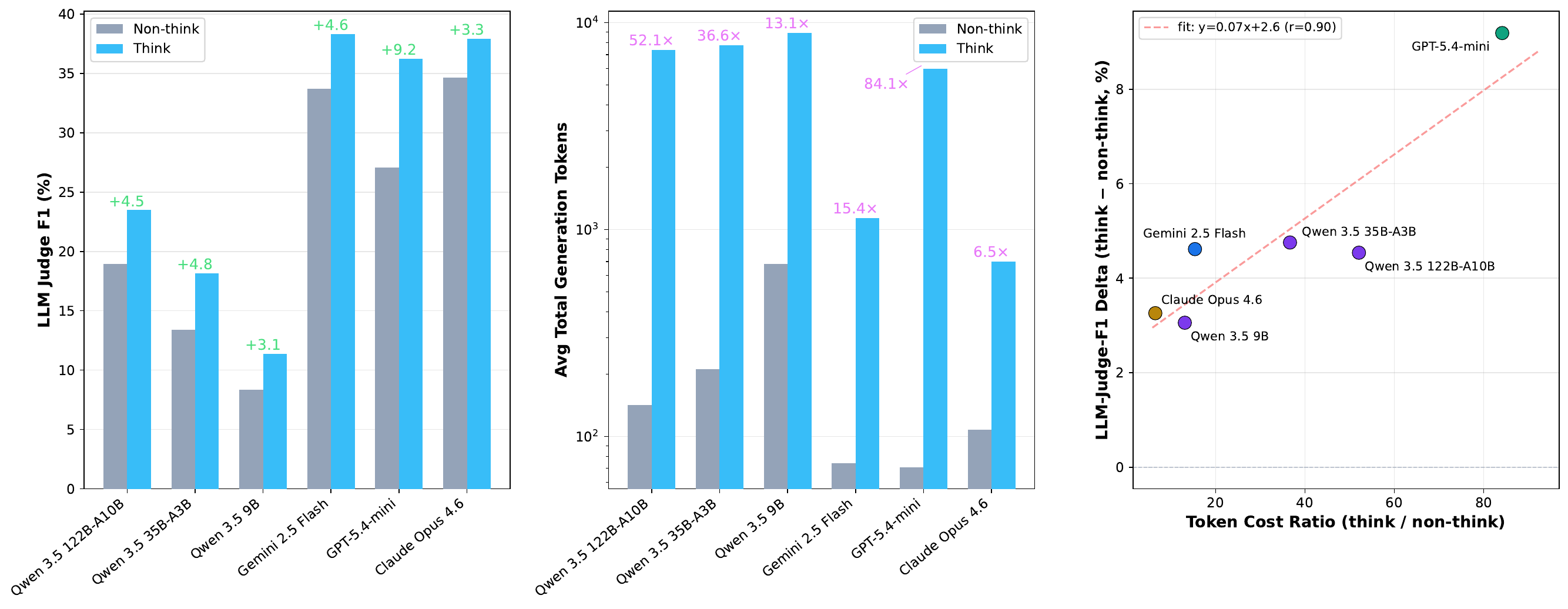}
    \caption{Comparison of thinking vs.\ non-thinking inference. \textbf{Left:} LLM-Judge-F1; annotations show the delta. \textbf{Center:} Average total generation tokens; annotations show the cost multiplier. \textbf{Right:} Cost ratio vs.\ \judgef{} delta with best-fit line.}
    \label{fig:think_vs_nothink}
\end{figure}

Fig.~\ref{fig:think_vs_nothink} quantifies the thinking trade-off across paired model configurations. Thinking improves LLM-Judge-F1 for every model, but the token cost varies by orders of magnitude across families. Some models achieve meaningful accuracy gains at relatively modest cost multipliers, while others pay far more in generation tokens without proportionally larger gains. The scatter of cost ratio against accuracy delta reveals that models that spend the most additional tokens on thinking do not necessarily gain the most accuracy, suggesting diminishing returns from extended reasoning on this task.

\subsection{Reasoning Budget Allocation}
\label{subsec:appendix_reasoning_budget_allocation}

\begin{figure}[ht]
    \centering
    \includegraphics[width=0.85\textwidth]{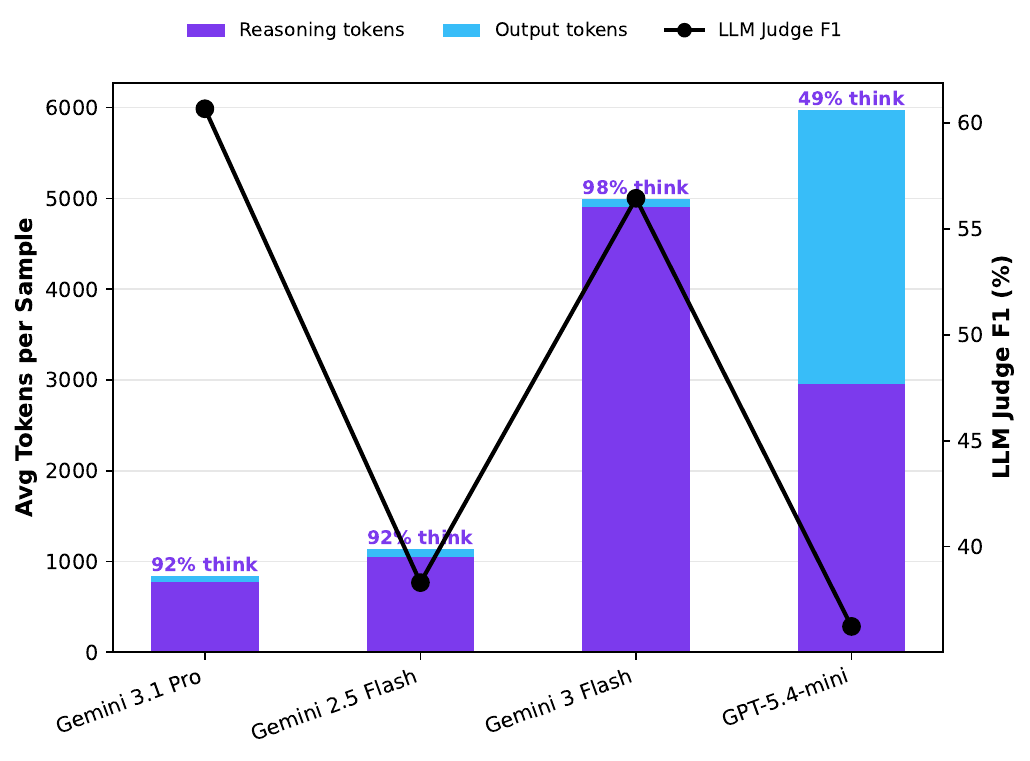}
    \caption{Reasoning vs.\ output token allocation for models with explicit token separation. Stacked bars show average tokens per sample; percentages indicate the fraction spent on reasoning. The black line shows \judgef{}.}
    \label{fig:reasoning_budget}
\end{figure}

For the models for which we have explicit reasoning token separation, Fig.~\ref{fig:reasoning_budget} shows how the generation budget is split between thinking and output. The allocation strategies differ dramatically: Gemini 3 Flash, evaluated in its default high-reasoning mode, dedicates nearly all tokens to reasoning and produces the highest absolute token count of the three Gemini models, while GPT-5.4-mini splits roughly evenly between reasoning and output. Gemini 3.1 Pro (low-reasoning mode) and Gemini 2.5 Flash (dynamic-reasoning mode) fall between these extremes. Gemini 3.1 Pro achieves the best accuracy while generating the least number of tokens among the four. This suggests that for factual list retrieval, the model's underlying knowledge capacity may matter more than the amount of inference-time reasoning it is allowed to perform. An open question is how these reasoning tokens are allocated between knowledge retrieval and instruction following (two capabilities that \listqa{} is specifically designed to disentangle) and whether the balance shifts across model architectures.

\subsection{Single-Faceted vs.\ Multi-Faceted Generation Patterns}
\label{subsec:appendix_single_vs_multi_tokengen}

\begin{figure}[p]
    \centering
    \includegraphics[width=\textwidth,height=0.9\textheight,keepaspectratio]{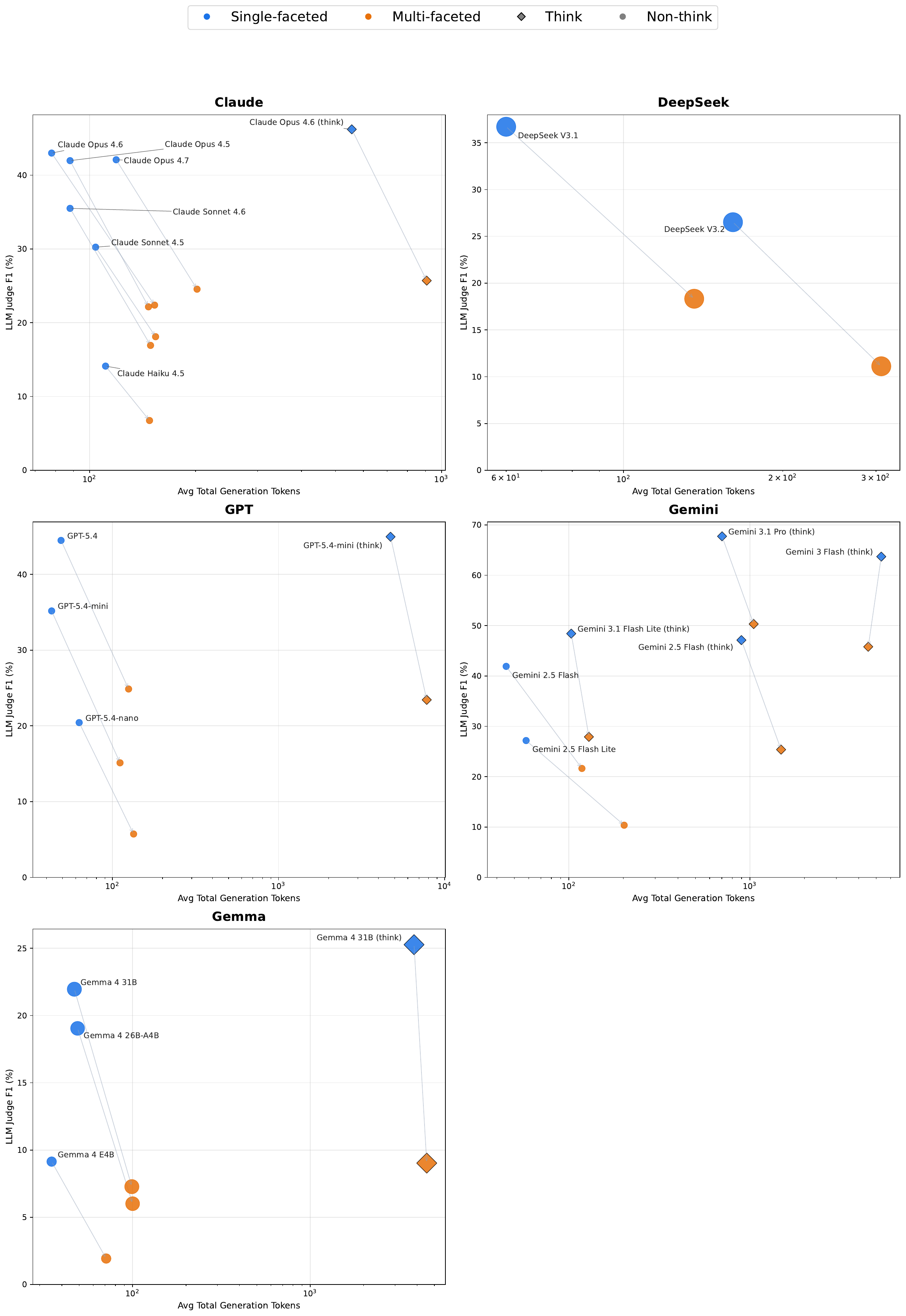}
    \caption{Generation cost vs.\ \judgef{} by prompt type, grouped by model family. \textcolor{blue}{blue} = single-faceted, \textcolor{orange}{orange} = multi-faceted. Arrows connect the same model's single-faceted (source) and multi-faceted (target) results. Diamond markers indicate thinking-enabled models.}
    \label{fig:prompt_type_efficiency_a}
\end{figure}

\begin{figure}[p]
    \centering
    \includegraphics[width=\textwidth,height=0.9\textheight,keepaspectratio]{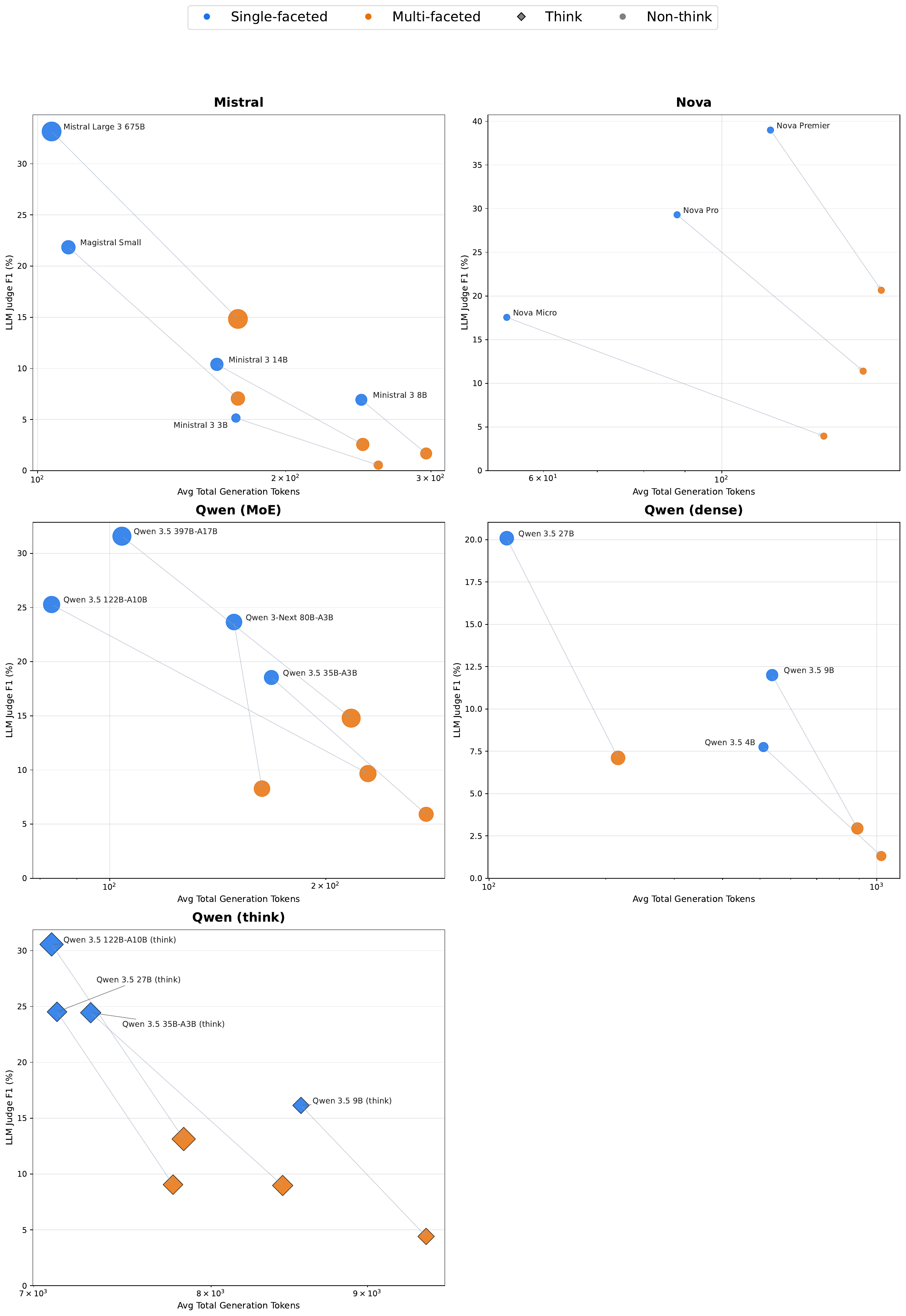}
    \caption{Generation cost vs.\ \judgef{} by prompt type, grouped by model family. \textcolor{blue}{blue} = single-faceted, \textcolor{orange}{orange} = multi-faceted. Arrows connect the same model's single-faceted (source) and multi-faceted (target) results. Diamond markers indicate thinking-enabled models.}
    \label{fig:prompt_type_efficiency_b}
\end{figure}

Figs.~\ref{fig:prompt_type_efficiency_a} and ~\ref{fig:prompt_type_efficiency_b} show generation cost and accuracy for single-faceted and multi-faceted prompts within each model family. Multi-faceted questions consistently elicit more tokens across all families, as expected given the relatively complex nature of the task. However, the accuracy gap between prompt types varies by family: some models maintain relatively stable accuracy despite the increased generation cost, while others show larger accuracy drops on multi-faceted questions that are not compensated by the additional generation.

\subsection{Per-Sample Score Distributions}
\label{subsec:appendix_persample_score_dist}

\begin{figure}[p]
    \centering
    \includegraphics[width=\textwidth,height=0.9\textheight,keepaspectratio]{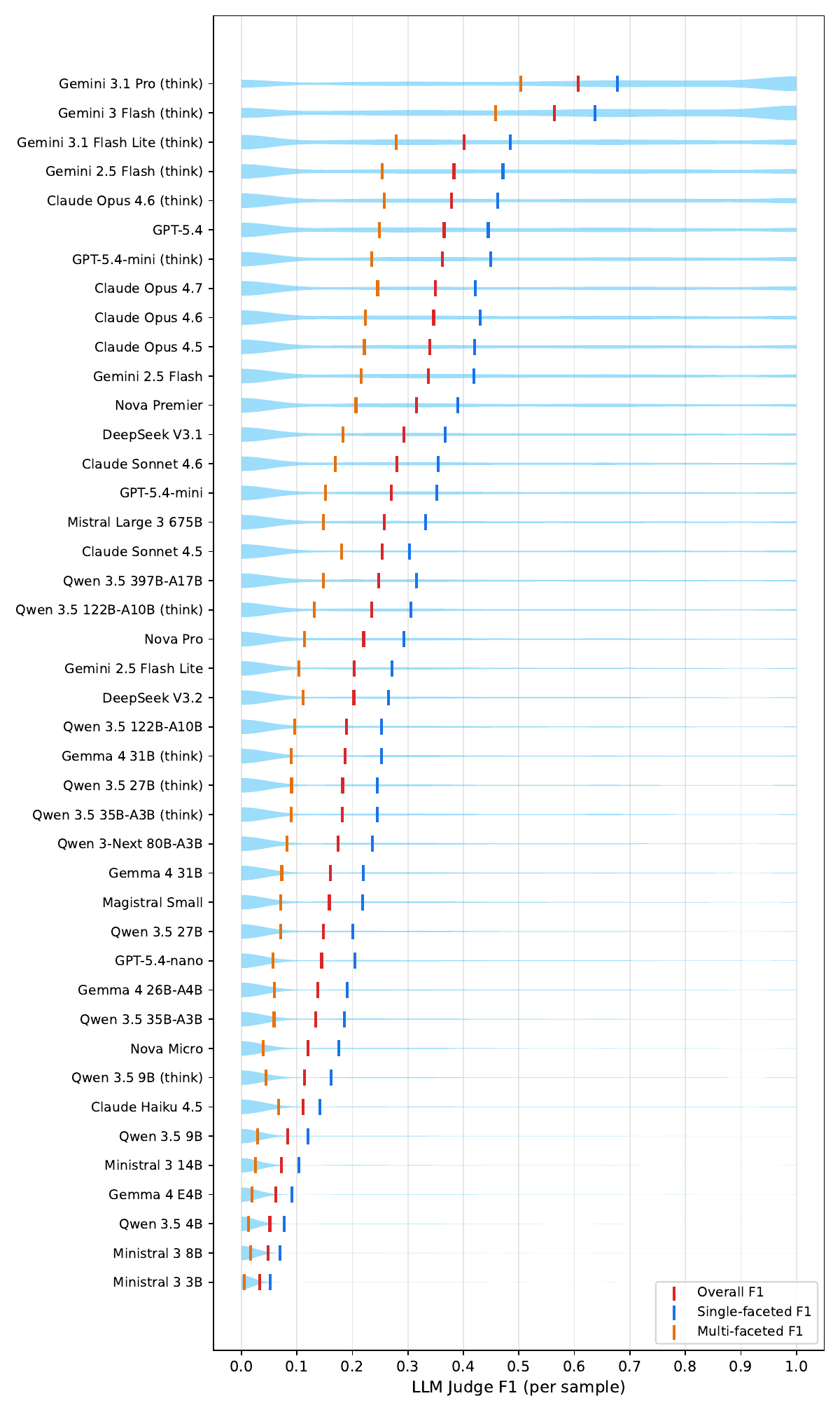}
    \caption{Per-sample \judgef{} distributions, sorted by overall F1 descending. Vertical markers indicate overall (\textcolor{red}{red}), single-faceted (\textcolor{blue}{blue}), and multi-faceted (\textcolor{orange}{orange}) mean \emph{LLM-Judge-F1}.}
    \label{fig:score_distributions}
\end{figure}

Fig.~\ref{fig:score_distributions} shows the per-sample \judgef{} distribution for each model configuration. Most models exhibit a zero-inflated distribution: a large mass of samples receive zero scores, indicating complete failure to produce correctly formatted list elements, while the remaining samples are spread across a range of partial scores. The strongest thinking models (Gemini 3.1 Pro, Gemini 3 Flash) stand out with a genuinely bimodal distribution, showing substantial mass at both ends, suggesting these models tend to either solve a sample well or miss it entirely with less middle ground. The gap between single-faceted and multi-faceted mean F1 is visible for every model, reinforcing the consistent difficulty gap observed in the main results.

\clearpage
\section{Qualitative Examples}
\label{sec:appendix_qualitative_examples}

We present selected examples that illustrate key phenomena observed in the quantitative results. For each example, we show the question, reference answer, and model outputs alongside their \judgef{} and \formatmetric{} scores.

\setlength{\fboxsep}{4pt}

\subsection{Knowledge vs.\ Format Trade-off}

The following single-faceted question asks for a flat list of Academy Awards. Three models produce qualitatively different responses, illustrating how factual accuracy and format compliance can diverge.

\smallskip
\noindent\fbox{\parbox{\dimexpr\linewidth-2\fboxsep-2\fboxrule}{\small
\textbf{Question:} What are the Academy Awards won by the ``All Quiet on the Western Front'' film as of 2024?

\smallskip
\textbf{Reference Answer} (4 elements, single-faceted):\\[2pt]
\texttt{1. Best International Feature Film}\\
\texttt{2. Best Production Design}\\
\texttt{3. Best Cinematography}\\
\texttt{4. Best Original Score}
}}

\smallskip
\noindent\fbox{\parbox{\dimexpr\linewidth-2\fboxsep-2\fboxrule}{\small
\textbf{Qwen 3.5 35B-A3B} \hfill \textit{LLM-Judge-F1 = 0.89} \quad \textit{Format = 0.00}

\smallskip
\texttt{1. Academy Award for Best International Feature Film}\\
\texttt{~~* Category: International Feature Film}\\
\texttt{~~* Sub-Element: Germany}\\
\texttt{2. Academy Award for Best Cinematography}\\
\texttt{~~* Category: Cinematography}\\
\texttt{~~* Sub-Element: James Friend}\\
\texttt{3. Academy Award for Best Original Score}\\
\texttt{~~* Category: Original Score}\\
\texttt{~~* Sub-Element: Volker Bertelmann}\\
\texttt{4. Academy Award for Best Production Design}\\
\texttt{~~* Category: Production Design}\\
\texttt{5. Academy Award for Best Visual Effects}\\
\texttt{~~* Category: Visual Effects}

\medskip
\textit{Analysis:} Retrieves all four correct awards but hallucinates a fifth (Best Visual Effects), and produces multi-faceted sub-bullets for a single-faceted question (Format = 0). LLM-Judge-F1 reflects precision = 4/5 and recall = 4/4.
}}

\smallskip
\noindent\fbox{\parbox{\dimexpr\linewidth-2\fboxsep-2\fboxrule}{\small
\textbf{Qwen 3.5 122B-A10B} \hfill \textit{LLM-Judge-F1 = 0.73} \quad \textit{Format = 1.00}

\smallskip
\texttt{1. Best Picture}\\
\texttt{2. Best Director}\\
\texttt{3. Best International Feature Film}\\
\texttt{4. Best Cinematography}\\
\texttt{5. Best Production Design}\\
\texttt{6. Best Original Score}\\
\texttt{7. Best Sound}

\medskip
\textit{Analysis:} Correct flat format (Format = 1.00), but hallucinates Best Picture, Best Director, and Best Sound, awards the film did not win.
}}

\smallskip
\noindent\fbox{\parbox{\dimexpr\linewidth-2\fboxsep-2\fboxrule}{\small
\textbf{Qwen 3.5 4B} \hfill \textit{LLM-Judge-F1 = 0.00} \quad \textit{Format = 1.00}

\smallskip
\texttt{1. Best Director for Lewis Milestone}\\
\texttt{2. Best Motion Picture for Lewis Milestone}\\
\texttt{3. Best Art Direction for Carl D. Smith}

\medskip
\textit{Analysis:} Correct flat format (Format = 1.00), but every element is factually wrong. These awards are not from the film the question refers to. LLM-Judge-F1 = 0 despite perfect formatting.
}}

\smallskip
This example illustrates that knowledge retrieval and instruction following are partially independent capabilities: Qwen 3.5 35B-A3B achieves the best factual score but fails format compliance, Qwen 3.5 4B produces perfect format but entirely wrong facts, and Qwen 3.5 122B-A10B balances both but hallucinates additional awards.

\subsection{Multi-Faceted Difficulty}

The following multi-faceted question requires retrieving stadiums satisfying multiple constraints (capacity, location, time range) and producing hierarchical sub-attributes for each. Even some of the proprietary models find it challenging to retrieve all elements correctly.

\smallskip
\noindent\fbox{\parbox{\dimexpr\linewidth-2\fboxsep-2\fboxrule}{\small
\textbf{Question:} Which sports stadiums and arenas with a capacity greater than 15,000 were opened in California between 2017 and 2024, and what year was it opened, which city is it in, what is the capacity, what type is it, and who is it home to?

\smallskip
\textbf{Reference Answer} (5 elements, multi-faceted):\\[2pt]
\texttt{1. BMO Stadium}\\
\texttt{~~* Opened: 2018}\\
\texttt{~~* City: Los Angeles}\\
\texttt{~~* Capacity: 22,000}\\
\texttt{~~* Type: Soccer}\\
\texttt{~~* Home to: Los Angeles FC and Angel City FC}\\
\texttt{2. Chase Center}\\
\texttt{~~* Opened: 2019}\\
\texttt{~~* City: San Francisco}\\
\texttt{~~* Capacity: 18,064}\\
\texttt{~~* Type: Arena}\\
\texttt{~~* Home to: Golden State Warriors and Valkyries}\\
\texttt{3. SoFi Stadium}\\
\texttt{~~* Opened: 2020}\\
\texttt{~~* City: Inglewood}\\
\texttt{~~* Capacity: 70,240}\\
\texttt{~~* Type: Multi-purpose}\\
\texttt{~~* Home to: LA Chargers and LA Rams}\\
\texttt{4. Snapdragon Stadium}\\
\texttt{~~* Opened: 2022}\\
\texttt{~~* City: San Diego}\\
\texttt{~~* Capacity: 35,000}\\
\texttt{~~* Type: Football}\\
\texttt{~~* Home to: San Diego Aztecs, SD FC, SD Wave FC}\\
\texttt{5. Intuit Dome}\\
\texttt{~~* Opened: 2024}\\
\texttt{~~* City: Inglewood}\\
\texttt{~~* Capacity: 18,000}\\
\texttt{~~* Type: Arena}\\
\texttt{~~* Home to: Los Angeles Clippers}
}}

\smallskip
\noindent\fbox{\parbox{\dimexpr\linewidth-2\fboxsep-2\fboxrule}{\small
\textbf{Claude Opus 4.6} \hfill \textit{LLM-Judge-F1 = 0.5} \quad \textit{Format = 1.00}

\smallskip
\texttt{1. Chase Center}\\
\texttt{~~* Year Opened: 2019}\\
\texttt{~~* City: San Francisco}\\
\texttt{~~* Capacity: \~{}18,064}\\
\texttt{~~* Type: Arena}\\
\texttt{~~* Home To: Golden State Warriors (NBA)}\\
\texttt{2. SoFi Stadium}\\
\texttt{~~* Year Opened: 2020}\\
\texttt{~~* City: Inglewood}\\
\texttt{~~* Capacity: \~{}70,240}\\
\texttt{~~* Type: Stadium}\\
\texttt{~~* Home To: Los Angeles Rams (NFL), Los Angeles Chargers (NFL)}\\
\texttt{3. Intuit Dome}\\
\texttt{~~* Year Opened: 2024}\\
\texttt{~~* City: Inglewood}\\
\texttt{~~* Capacity: \~{}18,000}\\
\texttt{~~* Type: Arena}\\
\texttt{~~* Home To: LA Clippers (NBA)}

\medskip
\textit{Analysis:} Correct format and accurate sub-attributes for the 2/3 stadiums it retrieves (Chase Center misses one home team), but misses BMO Stadium and Snapdragon Stadium entirely.
}}

\smallskip
\noindent\fbox{\parbox{\dimexpr\linewidth-2\fboxsep-2\fboxrule}{\small
\textbf{Qwen 3.5 35B-A3B} \hfill \textit{LLM-Judge-F1 = 0.33} \quad \textit{Format = 1.00}

\smallskip
\texttt{1. SoFi Stadium}\\
\texttt{~~* Year Opened: 2020}\\
\texttt{~~* City: Inglewood}\\
\texttt{~~* Capacity: 70,240}\\
\texttt{~~* Type: Multi-purpose stadium}\\
\texttt{~~* Home To: Los Angeles Rams and Los Angeles Chargers}\\
\texttt{2. Intuit Dome}\\
\texttt{~~* Year Opened: 2024}\\
\texttt{~~* City: Inglewood}\\
\texttt{~~* Capacity: 18,000}\\
\texttt{~~* Type: Basketball arena}\\
\texttt{~~* Home To: Los Angeles Clippers}\\
\texttt{3. BMO Stadium}\\
\texttt{~~* Year Opened: 2018}\\
\texttt{~~* City: Carson}\\
\texttt{~~* Capacity: 22,000 (expandable to 35,000)}\\
\texttt{~~* Type: Soccer stadium}\\
\texttt{~~* Home To: Los Angeles FC}\\
\texttt{4. Angel Stadium}\\
\texttt{~~* Year Opened: 2018 (Refurbishment/Expansion}\\
\texttt{~~~~completed, original 1966)}\\
\texttt{* Correction: The only newly constructed}\\
\texttt{~~stadiums/arenas >15,000 capacity opened in CA}\\
\texttt{~~between 2017-2024 are SoFi Stadium (2020),}\\
\texttt{~~Intuit Dome (2024), and BMO Stadium (2018).}\\
\texttt{[repeats elements 1--3 again]}

\medskip
\textit{Analysis:} Retrieves 3 correct stadiums but then hallucinates Angel Stadium, self-corrects mid-output, and repeats the list, producing a malformed response. The parser extracts the initial elements, yielding LLM-Judge-F1 = 0.33. This illustrates how smaller models can ``think aloud'' in ways that degrade structured output.
}}

\smallskip
\noindent\fbox{\parbox{\dimexpr\linewidth-2\fboxsep-2\fboxrule}{\small
\textbf{Qwen 3.5 9B} \hfill \textit{LLM-Judge-F1 = 0.29} \quad \textit{Format = 0.00}

\smallskip
\texttt{1. SoFi Stadium}\\
\texttt{~~* Year Opened: 2020}\\
\texttt{~~* City: Inglewood}\\
\texttt{~~* Capacity: 70,240}\\
\texttt{~~* Type: Multi-purpose stadium}\\
\texttt{~~* Home To: Los Angeles Rams, Los Angeles Chargers}\\
\texttt{2. Intuit Dome}\\
\texttt{~~* Year Opened: 2024}\\
\texttt{~~* City: Inglewood}\\
\texttt{~~* Capacity: 18,000}\\
\texttt{~~* Type: Indoor arena}\\
\texttt{~~* Home To: Los Angeles D-Fenders (G League),}\\
\texttt{~~~~Los Angeles Forum (NBA G League),}\\
\texttt{~~~~occasional NBA teams}

\medskip
\textit{Analysis:} Retrieves only 2 of 5 stadiums and hallucinates the home teams for Intuit Dome (the actual team is the LA Clippers). Format = 0 because only 2 elements are produced, below the 3-element minimum.
}}

\subsection{Effect of Thinking}

We compare the same model (Gemini 2.5 Flash) with and without thinking enabled on a multi-faceted question requiring specific player names, colleges, and student status. This example illustrates a case where thinking substantially improves factual retrieval, though even with thinking the model does not achieve full accuracy.

\smallskip
\noindent\fbox{\parbox{\dimexpr\linewidth-2\fboxsep-2\fboxrule}{\small
\textbf{Question:} Who made up the Midwest Regional Team for the 2018 NCAA tournament, what colleges were they from, and what was their student status at the time?

\smallskip
\textbf{Reference Answer} (6 elements, multi-faceted):\\
\texttt{1. Malik Newman (Most Outstanding Player)}\\
\texttt{~~* College: Kansas}\\
\texttt{~~* Student status: Sophomore}\\
\texttt{2. Trevon Duval}\\
\texttt{~~* College: Duke}\\
\texttt{~~* Student status: Freshman}\\
\texttt{3. Gabe DeVoe}\\
\texttt{~~* College: Clemson}\\
\texttt{~~* Student status: Senior}\\
\texttt{4. Gary Trent Jr.}\\
\texttt{~~* College: Duke}\\
\texttt{~~* Student status: Freshman}\\
\texttt{5. Marvin Bagley III}\\
\texttt{~~* College: Duke}\\
\texttt{~~* Student status: Freshman}\\
\texttt{6. Devonte' Graham}\\
\texttt{~~* College: Kansas}\\
\texttt{~~* Student status: Senior}
}}

\smallskip
\noindent\fbox{\parbox{\dimexpr\linewidth-2\fboxsep-2\fboxrule}{\small
\textbf{Gemini 2.5 Flash (no-think)} \hfill \textit{LLM-Judge-F1 = 0.00} \quad \textit{Format = 1.00}

\smallskip
\texttt{1. Kansas Jayhawks}\\
\texttt{~~* College: University of Kansas}\\
\texttt{~~* Student Status: Undergraduate}\\
\texttt{2. Duke Blue Devils}\\
\texttt{~~* College: Duke University}\\
\texttt{~~* Student Status: Undergraduate}\\
\texttt{3. Michigan State Spartans}\\
\texttt{~~* College: Michigan State University}\\
\texttt{~~* Student Status: Undergraduate}\\
\texttt{4. Clemson Tigers}\\
\texttt{~~* College: Clemson University}\\
\texttt{~~* Student Status: Undergraduate}

\medskip
\textit{Analysis:} Lists team names instead of individual players and uses generic ``Undergraduate'' for student status, a complete misinterpretation of the question. Despite perfect format compliance (Format = 1.00), every element is factually wrong, yielding LLM-Judge-F1 = 0.
}}

\smallskip
\noindent\fbox{\parbox{\dimexpr\linewidth-2\fboxsep-2\fboxrule}{\small
\textbf{Gemini 2.5 Flash (think)} \hfill \textit{LLM-Judge-F1 = 0.73} \quad \textit{Format = 1.00}

\smallskip
\textit{Reasoning trace (abbreviated):}\\
{\scriptsize\texttt{%
I need to dig up the info on the 2018 NCAA Tournament Midwest Regional team, specifically the All-Tournament Team. [...] These players come to mind: Malik Newman (Kansas), Devonte' Graham (Kansas), Marvin Bagley III (Duke), Gary Trent Jr.\ (Duke), Grayson Allen (Duke). Now, to get the student status [...] Malik Newman: redshirt sophomore, having transferred from Mississippi State. Devonte' Graham: played for Kansas from 2014-2018, so he was a senior. Marvin Bagley III: one season at Duke, so freshman. [...]%
}}

\smallskip
\textit{Final output:}\\
\texttt{1. Malik Newman}\\
\texttt{~~* College: Kansas}\\
\texttt{~~* Student Status: Redshirt Sophomore}\\
\texttt{2. Devonte' Graham}\\
\texttt{~~* College: Kansas}\\
\texttt{~~* Student Status: Senior}\\
\texttt{3. Marvin Bagley III}\\
\texttt{~~* College: Duke}\\
\texttt{~~* Student Status: Freshman}\\
\texttt{4. Gary Trent Jr.}\\
\texttt{~~* College: Duke}\\
\texttt{~~* Student Status: Freshman}\\
\texttt{5. Grayson Allen}\\
\texttt{~~* College: Duke}\\
\texttt{~~* Student Status: Senior}

\medskip
\textit{Analysis:} Retrieves 4 of 6 correct players (Newman, Graham, Bagley III, Trent Jr.) with accurate sub-attributes. Hallucinates Grayson Allen and misses Duval and DeVoe. Correct hierarchical format throughout.
}}

\smallskip
In this example, thinking enables the model to correctly interpret the question as asking for individual players rather than teams. However, even with thinking, the model retrieves only 4 of 6 correct players and hallucinates one other. Both variants produce well-formatted hierarchical lists, suggesting that the difference lies in knowledge retrieval rather than instruction following.

\subsection{Ordering Errors}

For the 14.6\% of questions requiring a specific sequence, we additionally compute \textit{LLM-Judge-F1-Ordered}, which matches elements positionally rather than via optimal assignment. The following example asks for retired jersey numbers in order of ceremony date, a sequence that differs from numerical order.

\smallskip
\noindent\fbox{\parbox{\dimexpr\linewidth-2\fboxsep-2\fboxrule}{\small
\textbf{Question:} In ascending order of ceremony date, list the first five numbers retired by the Houston Rockets basketball team.

\smallskip
\textbf{Reference Answer} (5 elements, single-faceted, ordered):\\
\texttt{1. Number 45}\\
\texttt{2. Number 23}\\
\texttt{3. Number 24}\\
\texttt{4. Number 22}\\
\texttt{5. Number 34}
}}

\smallskip
\noindent\fbox{\parbox{\dimexpr\linewidth-2\fboxsep-2\fboxrule}{\small
\textbf{GPT-5.4-mini (think)} \hfill \textit{LLM-Judge-F1 = 1.00} \quad \textit{LLM-Judge-F1-Ordered = 0.20}

\smallskip
\texttt{1. 24}\\
\texttt{2. 23}\\
\texttt{3. 34}\\
\texttt{4. 45}\\
\texttt{5. 22}

\medskip
\textit{Analysis:} Retrieves all five correct jersey numbers (LLM-Judge-F1 = 1.00 via optimal matching), but the sequence is scrambled. Only 1 of 5 elements is in the correct position, yielding LLM-Judge-F1-Ordered = 0.20.
}}

\smallskip
\noindent\fbox{\parbox{\dimexpr\linewidth-2\fboxsep-2\fboxrule}{\small
\textbf{Claude Opus 4.6} \hfill \textit{LLM-Judge-F1 = 1.00} \quad \textit{LLM-Judge-F1-Ordered = 0.60}

\smallskip
\texttt{1. \#23 (Calvin Murphy)}\\
\texttt{2. \#45 (Rudy Tomjanovich)}\\
\texttt{3. \#24 (Moses Malone)}\\
\texttt{4. \#22 (Clyde Drexler)}\\
\texttt{5. \#34 (Hakeem Olajuwon)}

\medskip
\textit{Analysis:} Also retrieves all five correct numbers (LLM-Judge-F1 = 1.00) and additionally provides player names. The ordering is closer to correct but swaps the first two positions, yielding LLM-Judge-F1-Ordered = 0.60.
}}

\smallskip
Both models retrieve the complete set of facts but struggle with the specific sequencing constraint. The gap between unordered and ordered scores isolates ordering as a distinct error source, separate from factual retrieval.

%% file: checklist.tex
\clearpage
\section*{NeurIPS Paper Checklist}

\begin{enumerate}

\item {\bf Claims}
    \item[] Question: Do the main claims made in the abstract and introduction accurately reflect the paper's contributions and scope?
    \item[] Answer: \answerYes{}
    \item[] Justification: The abstract and introduction (Sec.~\ref{sec:introduction}) clearly state the contributions: the dataset, the evaluation framework, the data quality methodology, and the experimental benchmarking. All claims are supported by the corresponding sections.

\item {\bf Limitations}
    \item[] Question: Does the paper discuss the limitations of the work performed by the authors?
    \item[] Answer: \answerYes{}
    \item[] Justification: App.~\ref{sec:limitations} discusses limitations including English-only scope, geographic skew, ground truth completeness beyond Wikipedia, judge cost and lack of abstention measurement.

\item {\bf Theory assumptions and proofs}
    \item[] Question: For each theoretical result, does the paper provide the full set of assumptions and a complete (and correct) proof?
    \item[] Answer: \answerNA{}
    \item[] Justification: This paper does not include theoretical results. It introduces a dataset and evaluation framework.

\item {\bf Experimental result reproducibility}
    \item[] Question: Does the paper fully disclose all the information needed to reproduce the main experimental results of the paper to the extent that it affects the main claims and/or conclusions of the paper (regardless of whether the code and data are provided or not)?
    \item[] Answer: \answerYes{}
    \item[] Justification: Sec.~\ref{sec:evaluation} fully describes the evaluation protocol, instruction prompts, and all five metrics. Sec.~\ref{sec:exp_setup} specifies the experimental setup. App.~\ref{sec:appendix_infrastructure} describes the infrastructure used for evaluating each model whose results are reported. Exact prompt templates are provided in App.~\ref{sec:appendix_prompt}.

\item {\bf Open access to data and code}
    \item[] Question: Does the paper provide open access to the data and code, with sufficient instructions to faithfully reproduce the main experimental results, as described in supplemental material?
    \item[] Answer: \answerYes{}
    \item[] Justification: The dataset is uploaded to \url{https://huggingface.co/datasets/AmazonScience/ListQA}, and the evaluation code is archived on Github at \url{https://github.com/amazon-science/ListQA-Eval}. For reproducibility, Sec.~\ref{sec:exp_setup} specifies the experimental setup. App.~\ref{sec:appendix_infrastructure} describes the tools and infrastructure used for evaluating each model. Exact prompt templates are provided in App.~\ref{sec:appendix_prompt}. Dataset and code are also linked at \url{https://listqa.github.io/}. Both dataset and code are released under CC-BY-NC 4.0 license.

\item {\bf Experimental setting/details}
    \item[] Question: Does the paper specify all the training and test details (e.g., data splits, hyperparameters, how they were chosen, type of optimizer) necessary to understand the results?
    \item[] Answer: \answerYes{}
    \item[] Justification: Sec.~\ref{sec:eval_protocol} describes the dataset variants (full, simple, complex), prompting modes (zero-shot, few-shot), and the instruction prompt. Sec.~\ref{sec:exp_setup} specifies the experimental setup. App.~\ref{sec:appendix_infrastructure} describes the tools and infrastructure used for evaluating each model. Exact prompt templates are provided in App.~\ref{sec:appendix_prompt}.

\item {\bf Experiment statistical significance}
    \item[] Question: Does the paper report error bars suitably and correctly defined or other appropriate information about the statistical significance of the experiments?
    \item[] Answer: \answerYes{}
    \item[] Justification: All metrics are reported as mean across samples, as described in Sec.~\ref{sec:evaluation}. Additionally, judge repeatability and judge agreement metrics are discussed in App.~\ref{sec:appendix_judge_agreement} and App.~\ref{sec:appendix_judge_repeatability}.

\item {\bf Experiments compute resources}
    \item[] Question: For each experiment, does the paper provide sufficient information on the computer resources (type of compute workers, memory, time of execution) needed to reproduce the experiments?
    \item[] Answer: \answerYes{}
    \item[] Justification: Sec.~\ref{sec:exp_setup} and App.~\ref{sec:appendix_infrastructure} describes the compute resources used for model inference and evaluation.

\item {\bf Code of ethics}
    \item[] Question: Does the research conducted in the paper conform, in every respect, with the NeurIPS Code of Ethics \url{https://neurips.cc/public/EthicsGuidelines}?
    \item[] Answer: \answerYes{}
    \item[] Justification: The research uses publicly available Wikipedia data, involves no deception, and annotators were compensated fairly. The broader impact is discussed in App.~\ref{sec:limitations}.

\item {\bf Broader impacts}
    \item[] Question: Does the paper discuss both potential positive societal impacts and negative societal impacts of the work performed?
    \item[] Answer: \answerYes{}
    \item[] Justification: App.~\ref{sec:limitations} discusses both the positive impact of improved factual evaluation and the potential for misuse of improved factual retrieval capabilities.

\item {\bf Safeguards}
    \item[] Question: Does the paper describe safeguards that have been put in place for responsible release of data or models that have a high risk for misuse (e.g., pre-trained language models, image generators, or scraped datasets)?
    \item[] Answer: \answerNA{}
    \item[] Justification: The dataset consists of factual question-answer pairs derived from publicly available Wikipedia content and does not pose risks for misuse.

\item {\bf Licenses for existing assets}
    \item[] Question: Are the creators or original owners of assets (e.g., code, data, models), used in the paper, properly credited and are the license and terms of use explicitly mentioned and properly respected?
    \item[] Answer: \answerYes{}
    \item[] Justification: Wikipedia content is used as a reference to annotate our dataset. All referenced models and tools are properly cited.

\item {\bf New assets}
    \item[] Question: Are new assets introduced in the paper well documented and is the documentation provided alongside the assets?
    \item[] Answer: \answerYes{}
    \item[] Justification: The dataset is documented with detailed statistics (Sec.~\ref{sec:dataset_statistics}, App.~\ref{sec:appendix_additional_data_stats}), annotation guidelines (App.~\ref{sec:appendix_guidelines}), and evaluation methodology (Sec.~\ref{sec:evaluation}, App.~\ref{sec:appendix_infrastructure}). More details can be found at \url{https://listqa.github.io/}.

\item {\bf Crowdsourcing and research with human subjects}
    \item[] Question: For crowdsourcing experiments and research with human subjects, does the paper include the full text of instructions given to participants and screenshots, if applicable, as well as details about compensation (if any)?
    \item[] Answer: \answerYes{}
    \item[] Justification: The full annotation guidelines are provided in App.~\ref{sec:appendix_guidelines}. Annotators were compensated fairly according to market standards.

\item {\bf Institutional review board (IRB) approvals or equivalent for research with human subjects}
    \item[] Question: Does the paper describe potential risks incurred by study participants, whether such risks were disclosed to the subjects, and whether Institutional Review Board (IRB) approvals (or an equivalent approval/review based on the requirements of your country or institution) were obtained?
    \item[] Answer: \answerNA{}
    \item[] Justification: The annotation task involves writing factual questions and answers based on Wikipedia tables and does not pose risks to participants. No IRB approval was required.

\item {\bf Declaration of LLM usage}
    \item[] Question: Does the paper describe the usage of LLMs if it is an important, original, or non-standard component of the core methods in this research? Note that if the LLM is used only for writing, editing, or formatting purposes and does \emph{not} impact the core methodology, scientific rigor, or originality of the research, declaration is not required.
    \item[] Answer: \answerYes{}
    \item[] Justification: The LLM-as-judge metric (LLM-Judge-F1, discussed in Sec.~\ref{sec:evaluation}) uses an LLM grader as a core evaluation component. The grader model and its usage are fully described in Sec.~\ref{sec:exp_setup}. The prompt templates are provided in App.~\ref{sec:appendix_prompt}.

\end{enumerate}